\documentclass[fleqn,10pt]{wlscirep}
\usepackage[utf8]{inputenc}
\usepackage[T1]{fontenc}
\usepackage{caption}
\usepackage{subcaption}
\usepackage{soul}
\usepackage{algorithm}
\usepackage{algpseudocode}

\usepackage{booktabs,threeparttable,pifont}
\usepackage{multirow}
 
\newcommand{\xmark}{\ding{55}}

\title{CarBench: A Comprehensive Benchmark for Neural Surrogates on High-Fidelity 3D Car Aerodynamics}

\author[1,2,*]{Mohamed Elrefaie}
\author[3]{Dule Shu}
\author[3]{Matt Klenk}
\author[1,2]{Faez Ahmed}
\affil[1]{Department of Mechanical Engineering, Massachusetts Institute of Technology, Cambridge, MA, USA}
\affil[2]{Schwarzman College of Computing, Massachusetts Institute of Technology, Cambridge, MA, USA}
\affil[3]{Future Product Innovation, Toyota Research Institute, Los Altos, CA, USA}

\affil[*]{Corresponding Author: mohamed.elrefaie@mit.edu}

\begin{abstract}
Benchmarking has been the cornerstone of progress in computer vision, natural language processing, and the broader deep learning domain, driving algorithmic innovation through standardized datasets and reproducible evaluation protocols. The growing availability of large-scale Computational Fluid Dynamics (CFD) datasets has opened new opportunities for applying machine learning to aerodynamic and engineering design. Yet, despite this progress, large-scale standardized benchmarks remain limited for high-resolution three-dimensional automotive aerodynamics, making it difficult to evaluate and compare models systematically.  In this work, we introduce CarBench, a large-scale standardized benchmark dedicated to high-resolution 3D automotive aerodynamics, performing a large-scale evaluation of state-of-the-art models on DrivAerNet++, the largest public dataset for automotive aerodynamics, containing over 8,000 high-fidelity car simulations. We assess eleven representative model configurations spanning multiple modeling paradigms: classical neural operator methods (e.g., Fourier Neural Operator), geometric deep learning approaches (PointNet, RegDGCNN, PointMAE, PointTransformer), transformer-based neural solvers (Transolver, Transolver++, AB-UPT), and implicit field networks leveraging triplane representations (TripNet). Beyond standard interpolation tasks, we perform cross-category experiments in which representative transformer-based solvers trained on a single car archetype are evaluated on unseen categories to probe generalization. Our analysis covers predictive accuracy, pressure-field fidelity relative to the CFD reference, computational efficiency, and statistical uncertainty in evaluation metrics, providing a unified comparison across modeling families and quantifying the robustness of performance estimates under shifts in test geometry distributions. To accelerate progress in data-driven engineering, we open-source the complete benchmark framework, including training pipelines, metric-level uncertainty estimation routines based on bootstrap resampling, and pretrained model weights, establishing a reproducible foundation for large-scale learning from high-fidelity external automotive aerodynamics simulations, available at \url{https://github.com/Mohamedelrefaie/CarBench}.
\end{abstract}

\begin{document}

\flushbottom
\maketitle

\begin{figure}[h!]
    \centering
    \includegraphics[width=\linewidth]{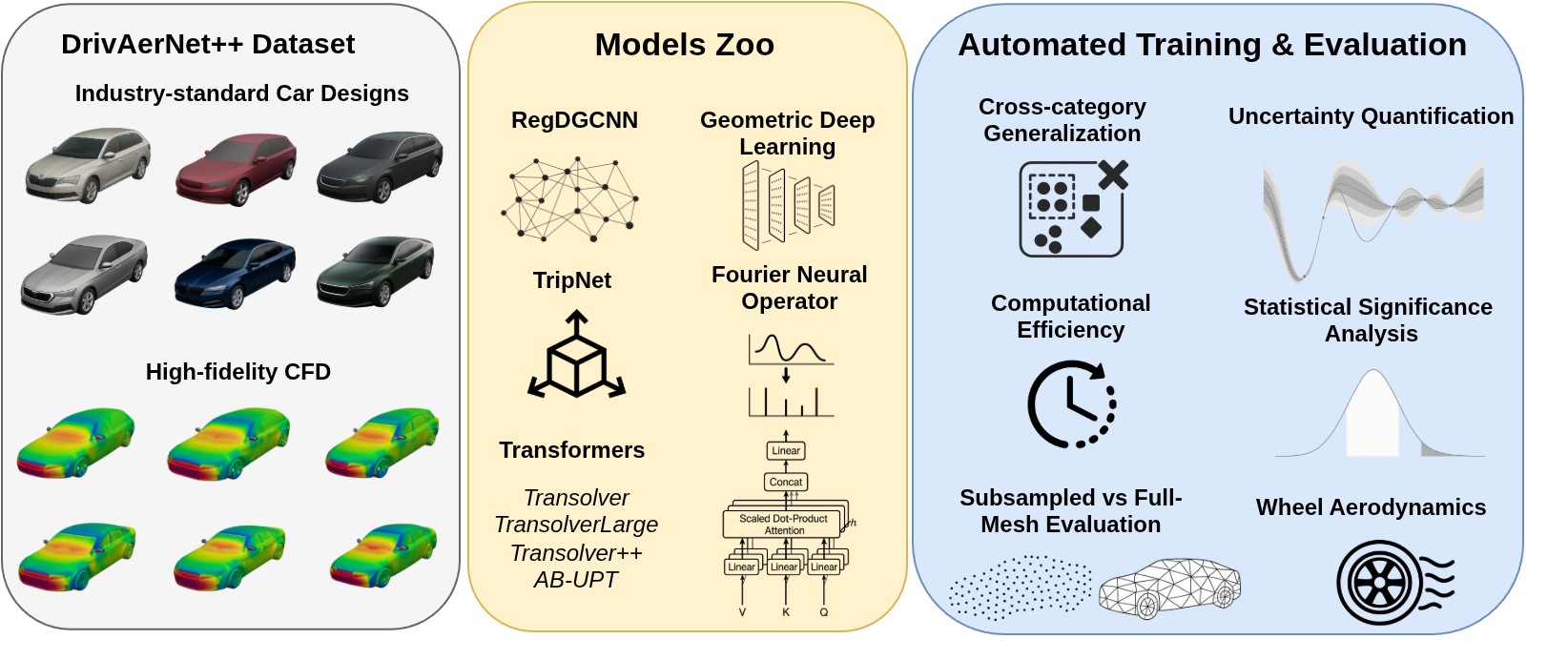}
\caption{Overview of CarBench. The benchmark provides a unified framework for evaluating machine learning models on high-fidelity automotive aerodynamics. It integrates diverse car geometries, CFD-derived surface pressure fields, and automated training pipelines with physics-based and ML metrics. Architectures from multiple model families (GNNs, Geometric Deep Learning, Neural Operators, Transformers, and Triplane networks) are evaluated under a standardized protocol. The framework includes cross-category generalization, uncertainty quantification, computational efficiency analysis, statistical significance evaluation, subsampled vs.\ full-mesh performance comparison, and wheel aerodynamics prediction. Together, these components enable a comprehensive and domain-specific assessment of aerodynamic surrogate models.
}
    \label{fig:DrivAerNet-Benchmark_Overview}
\end{figure}

\section*{Introduction}

Standardized benchmarks have played a pivotal role in advancing computer vision, natural language processing, and deep learning at large, enabling fair comparison, reproducibility, and accelerated innovation across tasks and model families.  In computer vision, the ImageNet benchmark~\cite{deng2009imagenet} enabled systematic evaluation at scale and catalyzed key breakthroughs such as AlexNet~\cite{krizhevsky2012imagenet}, a convolutional neural network that significantly outperformed previous methods and popularized deep learning for image classification. Similarly, in natural language processing, benchmarks like GLUE~\cite{wang2018glue}, SuperGLUE~\cite{wang2019superglue}, and MMLU~\cite{hendrycks2020measuring} supported the development of increasingly capable language models, including BERT~\cite{devlin2019bert} and GPT-3~\cite{brown2020language}, which established state-of-the-art performance across a range of understanding and reasoning tasks. These shared benchmarks help align evaluation practices and support incremental progress across diverse model families and applications.

In contrast, machine learning for computational design and numerical simulations remains fragmented and lacks a unified framework for quantitative comparison. While deep neural networks have shown great promise in accelerating computational fluid dynamics (CFD) simulations and enabling data-driven aerodynamic design~\cite{elrefaie2024surrogate, sung2025blendednet, elrefaie2024drivaernet, elrefaie2024drivaernet++, wu2024transolver, luo2025transolverPlus, alkin2025UPT}, the field still lacks a reproducible benchmark that can evaluate models systematically across architectures, geometries, and flow regimes. Most existing studies rely on small proprietary datasets~\cite{jacob2021deep,jacob2025benchmarking} or task-specific preprocessing pipelines, often reporting results that are not directly comparable. Moreover, inconsistencies in evaluation, such as mixing physics-based coefficients ($C_D$, $C_L$) with machine-learning-centric losses (MAE, MSE) across incompatible scales, have made it challenging to measure genuine algorithmic progress or generalization to unseen geometries. Recent open datasets such as DrivAerML~\cite{ashton2024drivaerml}, AhmedML~\cite{ashton2024ahmedml}, and SHIFT-SUV~\cite{luminarycloud_shift_suv_2025} represent important steps toward democratizing data-driven CFD-based research. However, their limited geometric diversity and lack of standardized train, validation, and test splits make it difficult to compare models fairly or assess true generalization. Even recent works that utilize subsets of DrivAerNet++~\cite{luo2025transolverPlus, liu2025aerogto} have focused on narrow tasks or specific car archetypes, leaving the broader question of benchmarking at scale unresolved. Consequently, despite rapid progress in learning-based CFD, there remains no established benchmark comparable to ImageNet or GLUE that unifies data, metrics, and evaluation protocols for the aerodynamic design community. While the benchmark proposed by Tangsali et al.~\cite{tangsali2025benchmarking} represents an important early step toward standardized AI evaluation in automotive aerodynamics, its scope reflects the constraints of an initial, narrowly focused effort. The study evaluates only three models on a relatively small subset of the DrivAerML dataset, with no assessment of cross-category generalization, uncertainty quantification, or statistical significance. In addition, the test set contains only 48 samples, which makes it difficult to draw statistically meaningful conclusions or assess true generalization performance across diverse geometries. The benchmark also relies on manually defined data splits (90/10) and does not address out-of-distribution testing or full-surface and volumetric flow extrapolation, which limits its usefulness for robust model comparison and real-world deployment. Furthermore, several modern state-of-the-art architectures (e.g., TripNet, Transolver, Transolver++, AB-UPT) are not included, even though Alkin et al.~\cite{alkin2025UPT} have shown that the models evaluated by Tangsali et al. lag behind contemporary methods by more than an order of magnitude in predictive accuracy.

To fill this gap, we introduce \textbf{CarBench}, the first large-scale standardized benchmark for machine learning in aerodynamic design. Built upon the high-fidelity DrivAerNet++ dataset~\cite{elrefaie2024drivaernet++}, which encompasses 8,150 steady-state CFD simulations of realistic car geometries, our benchmark provides a rigorous and reproducible platform for evaluating surrogate models under consistent physical and numerical settings. The benchmark defines common data splits, standardized evaluation metrics, and unified training configurations, enabling direct comparison across architectures and learning paradigms. Accurate aerodynamic modeling plays a pivotal role in car design, as small improvements in drag can translate directly into substantial gains in fuel efficiency and electric vehicle range. 
Among the various components of drag, pressure-induced effects dominate in most passenger cars due to their bluff geometries and large separated wakes. 
While friction drag can be significant for streamlined configurations, the contribution of pressure drag often reaches up to 80--90\% of the total aerodynamic resistance~\cite{schuetz2015aerodynamics, sudin2014review}. 
Consequently, learning to predict high-fidelity surface pressure fields is not only critical for estimating global aerodynamic coefficients such as $C_D$ and $C_L$, but also for understanding and optimizing flow separation, wake behavior, and overall car performance. 
In this benchmark, we therefore focus on the task of learning surface-level aerodynamic quantities from geometry, establishing a foundation for scalable, data-driven aerodynamic design.

We comprehensively evaluate eleven state-of-the-art (SOTA) models spanning multiple inductive biases, including neural operator methods such as the Fourier Neural Operator~\cite{li2020fourier}, geometric deep learning architectures such as PointNet~\cite{qi2017pointnet}, RegDGCNN~\cite{elrefaie2025drivaernet}, PointTransformer~\cite{zhao2021point}, and PointMAE~\cite{pang2023masked}, transformer-based neural solvers such as Transolver~\cite{wu2024transolver}, Transolver++~\cite{luo2025transolverPlus}, and AB-UPT~\cite{alkin2025UPT}, and implicit triplane representations such as TripNet~\cite{chen2025tripnet}. Our evaluation covers both interpolation and cross-category generalization across distinct car archetypes, including Fastbacks, Notchbacks, and Estatebacks, quantifying predictive accuracy, pressure-field fidelity relative to the CFD reference, and computational efficiency on high-resolution aerodynamic surfaces. To promote transparency and community participation, all training pipelines, evaluation scripts, and pretrained model checkpoints are released as open-source resources, which will be made publicly available upon publication. An overview of the benchmark framework is shown in Figure~\ref{fig:DrivAerNet-Benchmark_Overview}. 

CarBench also bears directly on generative design. The DrivAerNet++ geometries are themselves generated parametrically from eight base archetypes, and surrogates of the kind evaluated here are what generative and inverse design loops query in place of a solver: TripOptimizer~\cite{vatani2025tripoptimizer} performs differentiable shape optimization on this dataset, and the multi-agent Design Agents system~\cite{elrefaie2025ai} couples aerodynamic and aesthetic objectives. A generated design is only as reliable as the surrogate that scored it, so the accuracy and the uncertainty measured here indicate how reliable the surrogate scoring such a pipeline is.
Figure~\ref{fig:DrivAerNet-Benchmark_Overview} illustrates the full evaluation pipeline, including standardized metrics, multiple architectural families, model-scaling studies, and domain-specific tasks such as wheel--flow prediction. 
A central component of the framework is the dual-resolution evaluation protocol, which compares model predictions on subsampled surface representations with their corresponding full-resolution CFD surface meshes, enabling a more complete assessment of spatial accuracy. 
By establishing a reproducible, physics-grounded benchmark for large-scale numerical simulations, CarBench provides a unified foundation for measuring progress in data-driven aerodynamics. 
It enables fair comparison across models, promotes standardized evaluation practices, and supports future work on generalizable and physically consistent surrogate modeling for engineering design.

In this first release of CarBench, we focus exclusively on steady-state RANS simulations of external automotive aerodynamics and on learning surface pressure fields at a fixed freestream condition from geometry alone. We do not evaluate global aerodynamic coefficients such as drag and lift, volumetric flow fields, or unsteady regimes; these aspects are left as explicit targets for future extensions of the benchmark.

\section*{Results}
\subsection*{Learning Large-scale High-fidelity 3D Car
Aerodynamics}

\begin{table}[h!]
\scriptsize
\centering
\caption{Quantitative comparison on the unseen test set of 1{,}154 samples across machine learning models, ranked from worst to best by Relative L2 (lower is better). 
All models are evaluated with batch size 1 on an NVIDIA A100-SXM4-80GB GPU under identical inference settings. 
Per ISO GUM (Guide to the expression of uncertainty in measurement), uncertainties are rounded to two significant digits and central values are rounded to the same decimal place; precision is standardized per column using the largest uncertainty in that column (MSE (Mean Squared Error): 10, MAE (Mean Absolute Error): 0.01, RMSE (Root Mean Squared Error): 0.1, $R^2_{\text{test}}$: $10^{-4}$, Rel L2 (Relative L2 Error): $10^{-4}$). Here NeuralOperator denotes the voxelized Fourier Neural Operator baseline, i.e.\ FNO (voxelized), in which the geometry is mapped to a fixed $32^3$ grid rather than through a learned deformation; see the Baseline Models section of the Supplementary Information for the distinction from Geo-FNO and GINO. An asterisk (\textsuperscript{*}) denotes a model evaluated from the checkpoint provided by its original authors rather than trained in this work; the corresponding models trained from random initialization on the DrivAerNet++ training split alone are reported in Supplementary Table~6.}
\label{tab:quant-compare}
\setlength{\tabcolsep}{3pt}
\begin{tabular}{lccccccccc}
\hline
\\[-2ex]
\textbf{Model} & \textbf{\shortstack{Parameters\\(M)}} & \textbf{\shortstack{Peak\\Memory (GB)}} & \textbf{\shortstack{Mean\\Latency (ms)}} & \textbf{\shortstack{Throughput\\(samples/s)}}
& \textbf{\shortstack{MSE\\(m$^4$/s$^4$)}} & \textbf{\shortstack{MAE\\(m$^2$/s$^2$)}} & \textbf{\shortstack{RMSE\\(m$^2$/s$^2$)}}
& $R^2_{\text{test}}$ & \textbf{\shortstack{Rel L2}} \\
\hline
\\[-2ex]
PointNet~\cite{qi2017pointnet}        & 1.67 & 0.29 & 1.54  & 648.85 & 3350 \,$\pm$\, 50  & 31.12 \,$\pm$\, 0.170 & 57.9 \,$\pm$\, 0.44 & 0.7639 \,$\pm$\, 0.0027 & 0.3803 \,$\pm$\, 0.0020 \\
NeuralOperator~\cite{li2020fourier}   & 2.10 & 0.04 & 2.14  & 466.36 & 2130 \,$\pm$\, 47  & 25.02 \,$\pm$\, 0.140 & 46.2 \,$\pm$\, 0.51 & 0.8503 \,$\pm$\, 0.0026 & 0.3016 \,$\pm$\, 0.0019 \\
PointMAE~\cite{pang2023masked}        & 1.67 & 0.39 & 3.11  & 321.68 & 1716 \,$\pm$\, 39  & 22.58 \,$\pm$\, 0.110 & 41.4 \,$\pm$\, 0.47 & 0.8791 \,$\pm$\, 0.0023 & 0.2713 \,$\pm$\, 0.0016 \\
PointNetLarge~\cite{qi2017pointnet}   & 32.58& 1.50 & 8.29  & 120.65 & 1384 \,$\pm$\, 37  & 20.51 \,$\pm$\, 0.095 & 37.2 \,$\pm$\, 0.50 & 0.9025 \,$\pm$\, 0.0022 & 0.2436 \,$\pm$\, 0.0013 \\
RegDGCNN~\cite{elrefaie2025drivaernet} & 1.44 & 27.11& 231.98& 4.31   & 957 \,$\pm$\, 50   & 17.41 \,$\pm$\, 0.090 & 30.9 \,$\pm$\, 0.81 & 0.9327 \,$\pm$\, 0.0033 & 0.2006 \,$\pm$\, 0.0016 \\
PointTransformer~\cite{zhao2021point} & 3.05 & 6.65 & 95.68 & 10.50  & 917 \,$\pm$\, 82   & 17.05 \,$\pm$\, 0.085 & 30.3 \,$\pm$\, 1.3  & 0.9359 \,$\pm$\, 0.0054 & 0.1909 \,$\pm$\, 0.0024 \\
TripNet~\cite{chen2025tripnet}  & 7.92 & 0.51 & 41.32 & 24.20 
& 661 \,$\pm$\, 70   & 13.53 \,$\pm$\, 0.10  & 25.7 \,$\pm$\, 1.3 & 0.9537 \,$\pm$\, 0.0046   & 0.1623 \,$\pm$\, 0.0022 \\
Transolver++~\cite{luo2025transolverPlus} & 1.81 & 1.30 & 28.47 & 35.12 & 653 \,$\pm$\, 99   & 13.65 \,$\pm$\, 0.085 & 25.6 \,$\pm$\, 1.9  & 0.9543 \,$\pm$\, 0.0066 & 0.1573 \,$\pm$\, 0.0023 \\
Transolver~\cite{wu2024transolver}    & 2.47 & 1.51 & 29.84 & 33.51  & 605 \,$\pm$\, 99   & 12.75 \,$\pm$\, 0.090 & 24.6 \,$\pm$\, 2.0  & 0.9577 \,$\pm$\, 0.0067 & 0.1503 \,$\pm$\, 0.0024 \\
TransolverLarge~\cite{wu2024transolver} & 7.58& 1.68 & 28.41 & 35.20  & 579 \,$\pm$\, 100  & 12.18 \,$\pm$\, 0.090 & 24.1 \,$\pm$\, 2.1  & 0.9595 \,$\pm$\, 0.0069 & 0.1457 \,$\pm$\, 0.0025 \\
AB-UPT\textsuperscript{*}~\cite{alkin2025UPT}   & 6.01 & 0.27 & 30.65 & 32.63  & 559 \,$\pm$\, 160  & 10.81 \,$\pm$\, 0.100 & 23.6 \,$\pm$\, 3.2  & 0.9607 \,$\pm$\, 0.0101 & 0.1358 \,$\pm$\, 0.0024 \\
\hline
\end{tabular}
\end{table}

Table~\ref{tab:quant-compare} provides a comprehensive quantitative comparison of eleven SOTA machine learning models evaluated on the kinematic surface pressure fields of the DrivAerNet++ dataset. 
The results reveal a clear performance stratification across model families. 
Classical point-based networks such as PointNet~\cite{qi2017pointnet} and PointMAE~\cite{pang2023masked} achieve limited predictive accuracy ($R^2_{\text{test}}$ < 0.88, Rel L2 > 0.27), suggesting that these architectures may be insufficient for capturing the complex spatial patterns present in aerodynamic flows. 
Their limitations stem from simplified aggregation mechanisms, lower model capacity, and the lack of surface continuity modeling. 
Geometric and graph-based methods perform significantly better: RegDGCNN~\cite{elrefaie2025drivaernet} achieves a Rel L2 of 0.2006 and $R^2_{\text{test}} = 0.933$, leveraging localized message passing to improve surface-pressure prediction, albeit with substantial computational overhead (27~GB memory, 232~ms inference latency). 
PointTransformer~\cite{zhao2021point} further reduces the error to Rel L2 = 0.1909 and achieves $R^2_{\text{test}} = 0.936$, demonstrating improved spatial representation at moderate computational cost (6.6~GB).

\begin{figure}[h!]
    \centering
    \includegraphics[width=\linewidth]{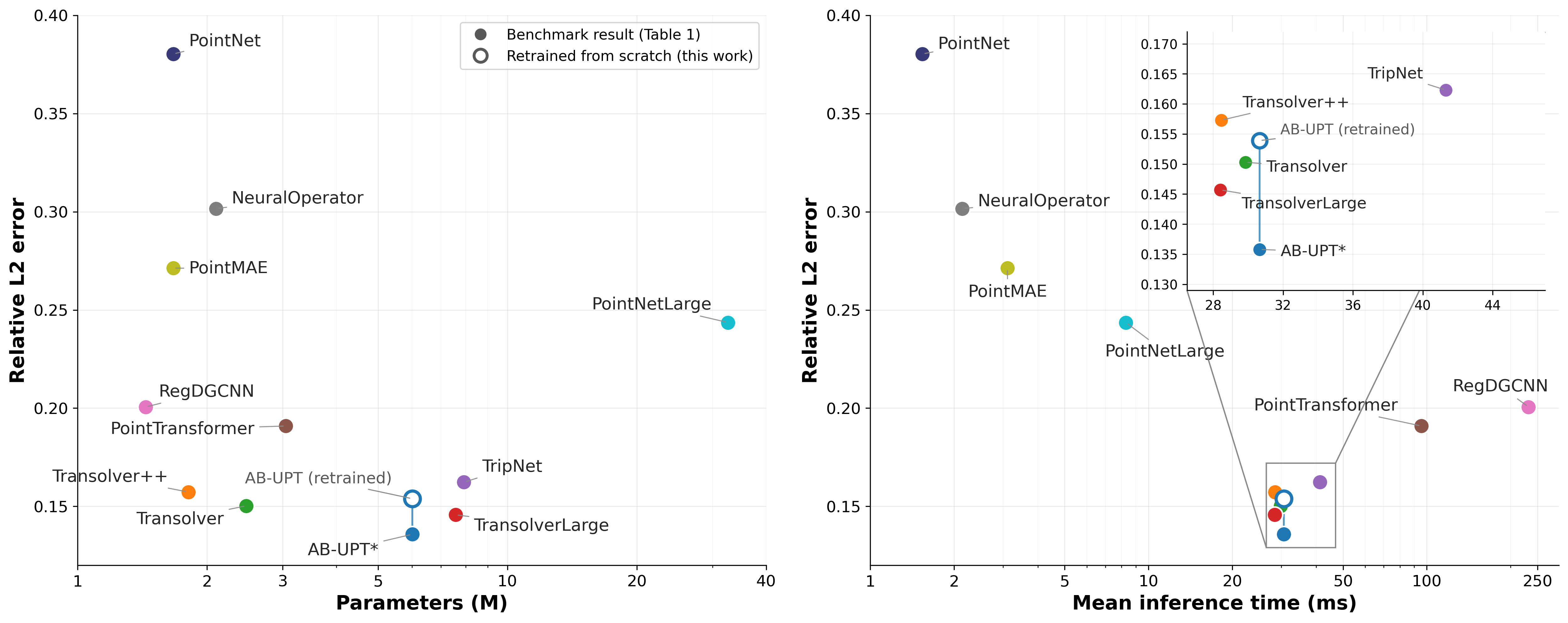}
    \caption{
Performance--efficiency trade-offs across model families on CarBench. 
Left: Relative L2 error versus model size (in millions of parameters, log scale). 
Right: Relative L2 error versus mean inference time (ms) on an NVIDIA A100 GPU. 
Transformer-based models together with the implicit-field model (TripNet) achieve the best accuracy--efficiency balance, combining low error with compact parameter counts and fast inference. 
Point-based baselines suffer from higher errors despite lower complexity, while graph-based networks incur large memory and latency overheads. 
These plots highlight the accuracy--efficiency trade-off across aerodynamic surrogate models, revealing models that are both accurate and scalable. The filled marker denotes the model evaluated from the checkpoint provided by its original authors (AB-UPT\textsuperscript{*}); the hollow marker denotes the same architecture retrained from random initialization in this work on the DrivAerNet++ training split alone, joined to its counterpart by a grey line. TripNet appears only as retrained in this work. Because parameter count and inference latency are determined by the architecture rather than by the weights, the two AB-UPT variants differ only along the accuracy axis: retraining shifts AB-UPT from a Relative L2 of $0.1358$ to $0.1539$.
}
    \label{fig:pareto-efficiency}
\end{figure}

\begin{figure}[h!]
    \centering
    \includegraphics[width=\linewidth]{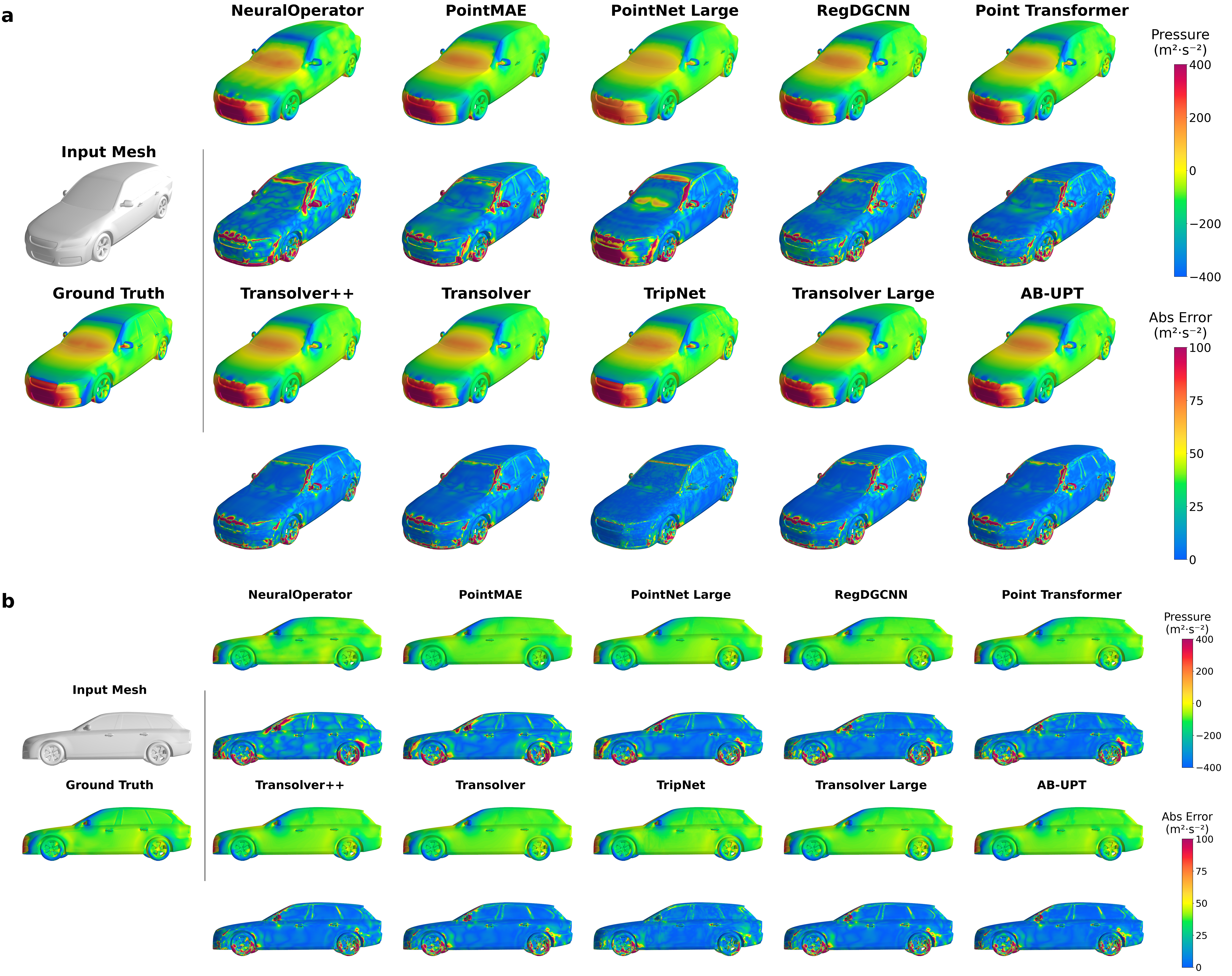}
\caption{Qualitative comparison of surface pressure predictions for design \texttt{E\_S\_WW\_WM\_648} from the unseen test set of the DrivAerNet++ dataset. \textbf{a}, Isometric view. \textbf{b}, Side view. In both panels the top row shows the predicted pressure field for each model and the bottom row the corresponding absolute error against the CFD ground truth. While all models capture the general pressure distribution, transformer-based architectures produce smoother predictions that match the CFD reference more closely around the stagnation region and in areas of high curvature, and TripNet gives the smoothest error map, which follows from its triplane-based volumetric sampling. TransolverLarge and AB-UPT\textsuperscript{*}, together with TripNet, form the top-performing group in both views, maintaining spatial coherence and capturing high-pressure regions more accurately than the other baselines. The panels labelled AB-UPT show the author-provided checkpoint, denoted AB-UPT\textsuperscript{*} in the tables.}
    \label{fig:composite-both}
\end{figure}

Transformer-based neural solvers offer the best trade-off between accuracy and efficiency. Both Transolver~\cite{wu2024transolver} and Transolver++~\cite{luo2025transolverPlus} achieve strong predictive performance ($R^2_{\text{test}} \approx 0.95$) with compact model sizes (2.5M and 1.8M parameters, respectively) and low inference latency of around 30~ms. The triplane-based TripNet~\cite{chen2025tripnet} performs comparably ($R^2_{\text{test}} = 0.954$) at a compact 7.9M parameters and low memory cost (0.5~GB), though with a higher inference latency of $41$~ms. Scaling the transformer architecture improves accuracy further---TransolverLarge reaches a relative error of 0.1457 and an $R^2_{\text{test}}$ of 0.9595, demonstrating the benefits of increased model capacity. The highest accuracy is achieved by AB-UPT\textsuperscript{*}~\cite{alkin2025UPT}, which sets a new benchmark with $R^2_{\text{test}} = 0.9607$ and Rel L2 = 0.1358, along with the lowest mean absolute error (10.8~m$^2$/s$^2$) and RMSE (23.6~m$^2$/s$^2$). This improvement is achieved with a moderate model size: AB-UPT\textsuperscript{*} contains 6.01M parameters, making it smaller than TransolverLarge (7.6M) and approximately 2.4$\times$ larger than the base Transolver. Despite this intermediate size, AB-UPT\textsuperscript{*} remains considerably more parameter-efficient than heavier architectures such as PointNetLarge (32.6M).  Transolver++ achieves Rel L2 = 0.1573 with the smallest parameter count among the high-performing models. Together, these results confirm that transformer-based and tokenization-aware architectures consistently outperform point-based and graph-based networks in both accuracy and scalability, establishing them as the state-of-the-art for learning high-resolution 3D aerodynamic fields. As illustrated in Figure~\ref{fig:pareto-efficiency}, transformer-based models occupy the most favourable region of the accuracy--efficiency plane, offering the best balance between predictive accuracy and computational efficiency. 
Transolver, Transolver++, and AB-UPT\textsuperscript{*} all achieve low relative errors (Rel L2 < 0.16) while maintaining compact model sizes under 7M parameters and sub-35~ms inference times. 
In contrast, networks such as PointNet and PointMAE trade accuracy for efficiency, running fastest while showing significantly higher errors. 
TripNet is accurate at modest parameter and memory cost, but its triplane representation must be fitted once per geometry before it can be queried, a cost not reflected in its query latency, while graph-based models like RegDGCNN suffer from long inference times despite strong performance. 
Overall, transformer-based solvers consistently occupy favorable regions of the accuracy--efficiency landscape, underscoring their suitability for scalable, high-fidelity surrogate modeling in aerodynamic applications.

Figure~\ref{fig:composite-both} provides a qualitative comparison of surface pressure predictions for the representative design \texttt{E\_S\_WW\_WM\_648} from the unseen test set, corresponding to an Estateback configuration with a smooth underbody and open wheels. This geometry was chosen as a typical rather than a best-case example: its error sits close to the median over the test split (Rel L2 $0.1499$ against a test-split median of $0.1456$ for TransolverLarge). Transformer-based solvers (Transolver, Transolver++, TransolverLarge, and AB-UPT\textsuperscript{*}), together with the implicit-field model (TripNet), produce smooth, physically consistent surface pressure fields that correctly capture high-pressure stagnation zones, surface curvature effects, and pressure recovery along the roof and rear surfaces. In contrast, point-based models introduce localized artifacts, especially near sharp geometric transitions such as the A-pillar, C-pillar, mirrors, and wheel housings. The corresponding error maps further show that AB-UPT\textsuperscript{*}, TransolverLarge, and TripNet deliver the most uniform and lowest-magnitude errors across the entire surface geometry, closely matching the CFD ground-truth distribution and demonstrating good robustness to complex surface features. The error distributions for TransolverLarge and AB-UPT\textsuperscript{*} are similar, likely a result of their shared reliance on point-cloud input representations with identical sampling strategies. In contrast, TripNet operates on triplane-encoded volumetric fields and discretizes the surface mesh differently, which may explain the distinct structure observed in its error map despite achieving similarly low overall error.

\subsection*{Wheel Aerodynamics}

\begin{figure}[h!]
    \centering
    \includegraphics[width=\linewidth]{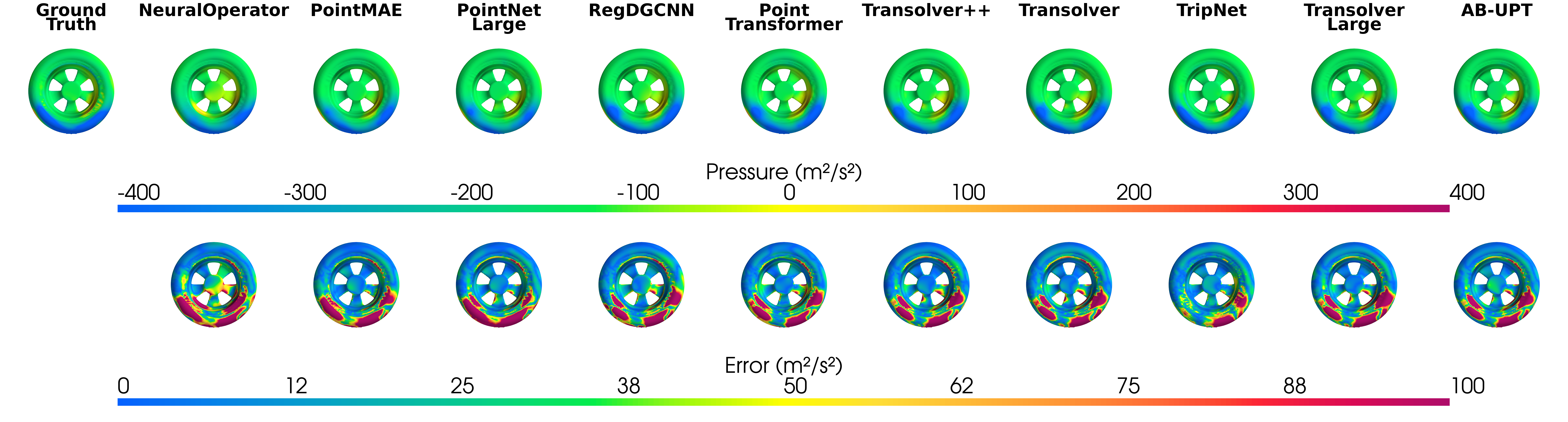}
\caption{Pressure prediction on the front-left wheel compared to the CFD ground truth. 
Models are shown in approximate order of Relative L2 error (worst to best, left to right); PointNet is omitted for space. 
Wheels can contribute up to 25\% of a car's total aerodynamic drag, making accurate pressure modeling around rotating components critical. 
Compared to prior open-source datasets, DrivAerNet++ uniquely provides detailed and smooth wheel and tire geometries, enabling evaluation across both open and closed wheel configurations.}

    \label{fig:front_left_wheel_comparison}
\end{figure}

We further evaluate model performance on the rotating wheel components
(Figure~\ref{fig:front_left_wheel_comparison}). In DrivAerNet++, wheels are
simulated using a rotating-wall boundary condition that induces realistic
surface-pressure gradients arising from tire rotation, rim geometry, and
near-ground interactions. The ground is modeled with a moving-wall condition
to match the car-relative flow field and ensure physically correct shear
behavior in the ground contact region. Wheels can contribute up to 25\% of a
car's total aerodynamic drag~\cite{brandt2019effects,pfadenhauer1996influence},
making them a critical region for assessing local prediction accuracy and
capturing high-frequency pressure variations. Unlike prior datasets,
DrivAerNet++ includes both open and closed wheel designs as well as smooth and
detailed tire geometries,
enabling a systematic evaluation of ML-based aerodynamic
prediction across diverse wheel geometries simulated using rotating-wall
boundary conditions.
These components introduce additional geometric complexity and strong
three-dimensional pressure gradients, driven by spoke geometry, rim cavities,
and tire curvature. Even under steady-state RANS conditions, these features
produce highly localized flow acceleration, separation tendencies, and
recirculation pockets around the wheel housing. Such effects create a
high-variance prediction regime where surrogate models often struggle to
maintain accuracy. By benchmarking model performance in this challenging
region, CarBench supports a closer assessment of local prediction quality.
Figure~\ref{fig:front_left_wheel_comparison} is a qualitative component-level
comparison; wheel-specific error metrics are not reported. In the illustrated wheel example, TripNet,  TransolverLarge, and AB-UPT\textsuperscript{*}
visually reproduce the CFD pressure pattern more closely than the other displayed predictions,
capturing the circumferential pressure distribution and steep gradients around the rim. In
contrast, point-based models exhibit localized artifacts and higher sensitivity
to small geometric features, highlighting the importance of global attention
mechanisms and sufficiently expressive representations for wheel aerodynamics.

\subsection*{Cross-category Generalization Analysis}

\begin{table}[h!]
\centering
\setlength{\tabcolsep}{3pt}
\caption{
Cross-category generalization performance of Transolver and Transolver++ across combinations of car design categories. 
Each row corresponds to a training/test category split using the DrivAerNet++ dataset. 
Models are evaluated in a zero-shot setting, with no fine-tuning or transfer learning on the test category. 
We report the Relative L2 error (lower is better) and $R^2_{\text{test}}$ score (higher is better), together with the training and test set sizes and the corresponding train-to-test ratio. In every experiment the designs of the training categories are split $80/20$ into training and validation sets, while all designs of the held-out categories are used for testing. The same category therefore appears with different counts depending on its role: the $5{,}332$ Fastback designs contribute $4{,}265$ training designs when trained on, and all $5{,}332$ are evaluated when held out. 
These results highlight the models' ability to extrapolate aerodynamic behavior across distinct vehicle classes under varying data imbalance conditions.}

\label{tab:crosscat-combined}
\begin{tabular}{llccccccc}
\hline
\\[-2ex]
\textbf{Train categories} & \textbf{Test categories} 
& \multicolumn{2}{c}{\textbf{Rel L2}} 
& \multicolumn{2}{c}{$\boldsymbol{R^2_{\text{test}}}$} 
& \textbf{Train} & \textbf{Test} & \textbf{Ratio} \\
\cline{3-6}
 & & \footnotesize Transolver & \footnotesize Transolver++ 
   & \footnotesize Transolver & \footnotesize Transolver++ 
   & \textbf{size} & \textbf{size} &  \\
\hline
\\[-2ex]
Fastbacks + Notchbacks              & Estatebacks                         & 0.2391 & 0.2182 & 0.9088 & 0.9241 & 5{,}388 & 1{,}386 & 3.89 \\
Estatebacks + Fastbacks             & Notchbacks                          & 0.1725 & 0.1729 & 0.9542 & 0.9540 & 5{,}374 & 1{,}403 & 3.83 \\
Estatebacks + Notchbacks            & Fastbacks                           & 0.4894 & 0.4988 & 0.6083 & 0.5933 & 2{,}231 & 5{,}332 & 0.42 \\
\hline
\\[-2ex]
Estatebacks                         & Fastbacks + Notchbacks              & 0.3980 & 0.4192 & 0.7446 & 0.7167 & 1{,}108 & 6{,}735 & 0.16 \\
Notchbacks                          & Estatebacks + Fastbacks             & 0.4295 & 0.4483 & 0.7002 & 0.6731 & 1{,}122 & 6{,}718 & 0.17 \\
Fastbacks                           & Estatebacks + Notchbacks            & 0.2539 & 0.2453 & 0.8990 & 0.9057 & 4{,}265 & 2{,}789 & 1.53 \\
\hline
\end{tabular}
\end{table}

Generalization to out-of-distribution (OOD) geometries is essential for real-world aerodynamic design, where predictive models must handle unseen vehicle configurations during early-stage development. All conventional passenger vehicles can be categorized into one of three aerodynamic archetypes: Estateback, Notchback, or Fastback, following the classification proposed by Hucho~\cite{hucho2013aerodynamik}. We conduct cross-category experiments in which models trained on one or more car archetypes are tested on entirely disjoint shape categories. This setup probes whether models learn transferable aerodynamic priors rather than memorizing category-specific geometric cues. To explore generalization under systematically varied training distributions, we focus on Transolver and Transolver++ as representative architectures. Both models offer strong accuracy--efficiency trade-offs, modest parameter counts, and fully open-source training pipelines, making them suitable for repeated retraining across many data splits. Accordingly, cross-category results reported here should be interpreted as characterizing these two representative architectures rather than the entire model zoo.
In every experiment the training categories are split $80/20$ into training and validation designs, while the held-out categories are tested in full. The same category therefore carries a smaller count when it is trained on than when it is tested --- the $5{,}332$ Fastbacks, for example, contribute $4{,}265$ training designs but are all evaluated when held out. As shown in Table~\ref{tab:crosscat-combined}, several consistent trends emerge:

\paragraph{1. Training sets containing Fastbacks show the strongest transfer.}
Fastback-containing training sets range from $4{,}265$ to $5{,}388$ designs depending on the split, and both models transfer well to Estateback and Notchback categories. For instance, training on Fastbacks alone and testing on Estatebacks + Notchbacks (last row of Table~\ref{tab:crosscat-combined}) yields
Rel L2 = 0.2539 (Transolver) and 0.2453 (Transolver++), among the best zero-shot results in the table. Fastback-containing training sets are also the largest in the table, so this trend reflects the combined effect of training-set size and geometric coverage.

\paragraph{2. Training on small categories leads to poor generalization across all directions.}
The Estateback-only and Notchback-only training sets contain $1{,}108$ and $1{,}122$ designs, respectively. When used alone for training, they generalize poorly to the other two categories, with Rel L2 in the range 0.3980--0.4483 and $R^2_{\text{test}}$ often below 0.72. These training sets are simultaneously the smallest and the least geometrically varied of the splits, and the design does not separate the two effects.

\paragraph{3. Fastback-containing two-category training sets show the strongest transfer.}
Training on (Fastbacks + Notchbacks) or (Estatebacks + Fastbacks) shows markedly stronger transfer. For example:
\begin{itemize}
    \item Fastbacks + Notchbacks $\rightarrow$ Estatebacks: Rel L2 = 0.2391 (Transolver)
    \item Estatebacks + Fastbacks $\rightarrow$ Notchbacks: Rel L2 $\approx$ 0.1725--0.1729, the best cross-category result in the table
\end{itemize}
The two-category configurations are associated with stronger transfer in the reported rows; however, the current splits do not isolate the effect of adding a category, because the target sets and training sizes differ between rows.

\paragraph{4. Predicting Fastbacks without Fastback training data is the hardest direction.}
Training on Estatebacks + Notchbacks ($2{,}231$ training designs) and testing on the full set of $5{,}332$ Fastbacks (third row of Table~\ref{tab:crosscat-combined}) results in the worst performance overall, with Rel~L2 values around $0.49$--$0.50$. Fastbacks feature curvature patterns, roofline transitions, and wake characteristics that are less prominent in the Estateback and Notchback geometries, placing them furthest from this training distribution. This asymmetry tracks the training-set sizes: the Fastback-only training set ($4{,}265$ designs) is nearly twice the combined Estateback + Notchback training set ($2{,}231$ designs), and models trained on it generalize well to the other categories, whereas the smaller combined set does not support zero-shot prediction of Fastbacks.

\paragraph{5. Training-set size is strongly associated with cross-category transfer.}
In the two asymmetric transfer directions, the larger training set generalizes substantially better than the smaller but more diverse one. Training solely on Fastbacks ($4{,}265$ training designs) yields strong generalization to Estatebacks and Notchbacks, achieving Rel~L2 values of $0.2539$ and $0.2453$. In contrast, training on the more diverse but much smaller Estateback + Notchback set ($2{,}231$ training designs) results in the poorest overall performance when predicting Fastbacks, with Rel~L2 near $0.49$--$0.50$. We report this as an observed association rather than a causal ordering. The two transfer directions differ simultaneously in training-set size, training composition, target archetype and test-set size, and no two rows of Table~\ref{tab:crosscat-combined} share a test set, so the present design cannot separate the contribution of dataset size from that of geometric diversity. Establishing which dominates would require a size-matched control, for example retraining on size-matched Fastback-only and Estateback\,+\,Notchback sets and evaluating both on the same held-out Notchback split.

Overall, transfer in these experiments is strongest when the training distribution spans a sufficient range of rear-end shapes and the distinct wake topologies and separation patterns they induce --- in these splits, when Fastbacks are part of the training set --- and weakest when the training set omits Fastbacks entirely, whether it is a small single category or the combined Estateback\,+\,Notchback set. As the preceding paragraphs note, the present splits do not separate the contribution of training-set size from that of geometric diversity. We note that this experiment varies the \emph{geometry} of the training distribution while the flow conditions are held fixed; the resulting measure of generalization is therefore one of geometric extrapolation, and should not be read as generalization across flow regimes. We return to this distinction in the Limitations and Future Work section.

\subsection*{Computational Efficiency and Performance Trade-offs}
The efficiency analysis reveals clear trade-offs between accuracy, runtime, and memory cost (Table~\ref{tab:quant-compare}). PointNet achieves the highest throughput ($649$~samples/s) and NeuralOperator the lowest memory footprint ($0.04$~GB) at $466$~samples/s, but in both cases this comes at the expense of predictive accuracy ($R^2_{\text{test}} = 0.76$ and $0.85$ respectively). In contrast, graph-based models such as RegDGCNN exhibit substantial latency ($\sim$232~ms) and extremely high memory usage (27~GB), reflecting the overhead of dynamic $k$-NN graph construction and message passing. Transformer-based architectures strike a far more favorable balance. Both Transolver and Transolver++ operate at 28--30~ms latency with modest memory requirements (1.3--1.5~GB), while TransolverLarge remains similarly efficient (28~ms, 1.68~GB). Implicit field models such as TripNet run at $41$~ms with low memory usage ($0.51$~GB); this figure is pressure-query latency from a fitted triplane and excludes the one-time geometry-specific fitting stage. AB-UPT\textsuperscript{*} offers competitive runtime characteristics (31~ms, 0.27~GB) while achieving the best overall accuracy, demonstrating that high predictive fidelity can be attained without incurring excessive computational overhead.

\subsection*{Subsampled vs.\ Full-Resolution Mesh Evaluation} 

\begin{table*}[h!]
\centering
\small
\caption{Subsampled versus full-resolution evaluation on a stratified subset of $61$ geometries drawn
from the held-out test split, sampled to preserve the archetype proportions of the full split, with all three protocols applied to the same geometries. \emph{Subsampled}
scores a $10{,}000$-point prediction on those points; \emph{Interpolated} upscales that same
prediction to every node of the full CFD surface mesh ($3.7$--$5.5\times10^{5}$ nodes, mean
$455{,}000$) with the radial-basis interpolation;
\emph{Patchwise} instead predicts every node directly using $\lceil N/10^4 \rceil$ disjoint
$10{,}000$-point passes ($37$--$55$ per geometry), and therefore involves no interpolation.
Patchwise evaluation recovers almost all of the apparent degradation, showing that the loss attributed to resolution is dominated by interpolation error.
$^{\dagger}$AB-UPT\textsuperscript{*} and TripNet decode arbitrary query points directly, so their
patchwise column evaluates every node from a single geometry encoding, with queries chunked only to
fit memory, rather than $\lceil N/10^4 \rceil$ independent passes. They are held to the identical
interpolated protocol as every other model, with the prediction restricted to the same $10{,}000$
points and upscaled in the same way, so that the column compares evaluation protocols rather than
architectures. AB-UPT\textsuperscript{*} is evaluated from the author-provided checkpoint, as in Table~\ref{tab:quant-compare}.}
\label{tab:surface_fullmesh_comparison}
\setlength{\tabcolsep}{6pt}
\renewcommand{\arraystretch}{1.1}
\begin{tabular}{lccccccc}
\toprule
& \multicolumn{3}{c}{\textbf{Relative L2}} & \multicolumn{3}{c}{$\mathbf{R^2}$} \\
\cmidrule(lr){2-4}\cmidrule(lr){5-7}
\textbf{Model} & \textbf{Subsampled} & \textbf{Interpolated} & \textbf{Patchwise}
               & \textbf{Subsampled} & \textbf{Interpolated} & \textbf{Patchwise} \\
\midrule
PointNet & 0.3763 & 0.3937 & 0.3791 & 0.7700 & 0.7485 & 0.7664 \\
NeuralOperator & 0.2962 & 0.3259 & 0.2987 & 0.8575 & 0.8277 & 0.8549 \\
PointMAE & 0.2619 & 0.3137 & 0.2670 & 0.8888 & 0.8403 & 0.8839 \\
PointNetLarge & 0.2390 & 0.3203 & 0.2423 & 0.9075 & 0.8337 & 0.9044 \\
RegDGCNN & 0.1954 & 0.2803 & 0.1994 & 0.9378 & 0.8724 & 0.9346 \\
PointTransformer & 0.1851 & 0.2798 & 0.1892 & 0.9442 & 0.8728 & 0.9410 \\
TripNet$^{\dagger}$ & 0.1569 & 0.2763 & 0.1610 & 0.9596 & 0.8759 & 0.9568 \\
Transolver++ & 0.1448 & 0.2717 & 0.1492 & 0.9656 & 0.8800 & 0.9625 \\
Transolver & 0.1436 & 0.2714 & 0.1482 & 0.9661 & 0.8802 & 0.9629 \\
TransolverLarge & 0.1391 & 0.2701 & 0.1437 & 0.9680 & 0.8813 & 0.9650 \\
AB-UPT\textsuperscript{*}$^{\dagger}$ & 0.1311 & 0.2688 & 0.1351 & 0.9716 & 0.8825 & 0.9688 \\
\bottomrule
\end{tabular}
\end{table*}

A common limitation in prior aerodynamic learning studies is that model accuracy is often reported only on subsampled point clouds or reduced mesh representations~\cite{li2020fourier, wu2024transolver}, leaving open whether the reported accuracy survives at the resolution of the underlying simulation. We therefore evaluate every model at full mesh resolution under three protocols applied to the same held-out geometries. \emph{Subsampled} scores a $10{,}000$-point prediction on those points, the regime in which the models were trained. \emph{Interpolated} upscales that same prediction to every node of the full CFD surface mesh ($3.7$--$5.5\times10^{5}$ nodes) using the radial-basis interpolant. \emph{Patchwise} instead covers the mesh with $\lceil N/10^4 \rceil$ disjoint $10{,}000$-point subsets ($37$--$55$ per geometry), passes each through the model independently and stitches the results, so that every node is predicted directly at the point density the model was trained on and no interpolation is performed. The final, smaller subset is padded back to $10{,}000$ points so that the token-set size seen by the network is identical in every pass. The subsampled and interpolated protocols are applied identically to all eleven models; the patchwise protocol differs for the two query-based decoders only in that a single geometry encoding serves every query (see the note to Table~\ref{tab:surface_fullmesh_comparison}). In particular, the two models with query-based decoders, which can evaluate every node directly and therefore never require interpolation in practice, are nonetheless held to the same interpolated protocol, with their prediction restricted to the same $10{,}000$ points before upscaling, so that the comparison isolates the evaluation protocol rather than confounding it with the architecture.

Table~\ref{tab:surface_fullmesh_comparison} shows that the two full-resolution protocols give very different answers, and that the choice of protocol dominates the apparent effect of resolution. Under interpolation all eleven models fall within a narrow band ($0.269$--$0.394$ Relative L2), and the differences between models are largely erased: the most accurate models roughly double their error, while the weakest are barely affected. Under patchwise evaluation the same models retain almost exactly their subsampled accuracy: TransolverLarge moves from $0.1391$ to $0.1437$, Transolver from $0.1436$ to $0.1482$, RegDGCNN from $0.1954$ to $0.1994$, and PointNet from $0.3763$ to $0.3791$. Across all eleven models the patchwise penalty is between $0.002$ and $0.005$ Relative L2, and the ranking is preserved.

The interpolation penalty is largest for the most accurate models ($+0.138$ Relative L2 for AB-UPT\textsuperscript{*} and $+0.131$ for TransolverLarge, against $+0.017$ for PointNet). More accurate predictions contain more high-frequency spatial detail, which is what is lost when a field is reconstructed from $10{,}000$ scattered samples, so the interpolation error grows with the sharpness of the field being interpolated. A protocol that interpolates therefore penalizes accurate models more than inaccurate ones. That this penalty is a property of the protocol and not of any particular architecture is shown by the two query-based models: AB-UPT\textsuperscript{*} and TripNet decode arbitrary query points directly and so never need to interpolate in practice, yet when their prediction is restricted to the same $10{,}000$ points and upscaled in the same way they lose just as much as the others ($+0.138$ and $+0.119$ Relative L2). Their advantage lies in avoiding the interpolation step rather than in reduced sensitivity to interpolation error.

Patchwise evaluation is not free, since it replaces one forward pass per geometry with $37$--$55$. Measured over $20$ geometries on an otherwise idle A100, the cost per geometry ranges from $0.05$~s for TripNet, whose triplane is fitted once and then queried, to $4.8$~s for PointTransformer, whose attention cost grows quickly with the number of passes; for the Transolver family it is $1.2$--$1.9$~s. Evaluating all eleven models over the $1{,}154$ test geometries patchwise costs approximately $4.4$ GPU-hours in total, against $3.3$ GPU-hours for the nearest-neighbor interpolation protocol, the additional forward passes being largely offset by the cost of interpolating each prediction onto half a million nodes. By contrast, radial-basis interpolation requires a dense solve per geometry and costs $66.3$~s per geometry on its own, or roughly $234$ hours for the same study. Accuracy at full resolution is thus attainable at a cost comparable to, and in this case far below, the interpolation it replaces. We therefore recommend patchwise evaluation for reported full-resolution results, treat interpolation as acceptable only for rapid iteration, and note that query-based decoders avoid the trade-off entirely, since a single geometry encoding already yields a prediction at every node.

\subsection*{Statistical Uncertainty of Evaluation Metrics}
In this section, the term `uncertainty' refers to statistical uncertainty in aggregate evaluation metrics, rather than predictive uncertainty on individual samples or spatially varying epistemic/aleatoric uncertainty.
To rigorously assess the uncertainty in our performance metrics, we apply stratified paired bootstrap resampling~\cite{efron1992bootstrap,efron1994introduction} with $B=2000$ replicates. Entire car geometries are resampled with replacement, ensuring that the spatial correlation structure of each surface is preserved. Stratification by car archetype (Estateback, Notchback, Fastback) maintains the original category proportions in every bootstrap replicate, which leads to confidence intervals that accurately reflect the geometric and aerodynamic diversity of the test set. Across models, the bootstrap uncertainty in Relative L2 error remains small compared to the mean values, with all confidence intervals lying between $ \pm 0.0013 $ and $ \pm 0.0025 $, while the mean Relative L2 spans from $0.1358$ to $0.3803$. This narrow range means that the ranking by Relative L2 is statistically stable; for example, the gap between AB-UPT\textsuperscript{*} $(0.1358 \pm 0.0024)$ and TransolverLarge $(0.1457 \pm 0.0025)$ is several times larger than their respective uncertainties. Lower performing models such as PointNet and NeuralOperator also show larger spread in absolute metrics, with MSE uncertainties of $ \pm 50 $ and $ \pm 47 $, respectively, which is consistent with their higher average error levels. For mid-range models such as RegDGCNN and PointTransformer, RMSE uncertainties of $ \pm 0.81 $ and $ \pm 1.3 $ indicate moderate variability across the 1{,}154 test geometries. 

In contrast, the strongest models in terms of Relative L2, including TripNet, Transolver, Transolver++, TransolverLarge, and AB-UPT\textsuperscript{*}, have similar Relative L2 uncertainties in the band $ \pm 0.0022 $ to $ \pm 0.0025 $, but their MSE and RMSE intervals are not uniformly small. AB-UPT\textsuperscript{*}, for instance, achieves the lowest overall error with Relative L2 $0.1358 \pm 0.0024$ and RMSE $23.6 \pm 3.2$, where the RMSE uncertainty is larger than for some weaker models in absolute terms. This pattern reflects the fact that even high-performing models can exhibit noticeable variation on the most challenging geometries, especially those that dominate the tails of the error distribution. There is no clear monotonic relationship between parameter count and uncertainty. TripNet, with 7.9 million parameters, has RMSE uncertainty of $ \pm 1.3 $, tighter than the smaller Transolver and Transolver++ models, which lie in the $ \pm 1.9 $ to $ \pm 2.0 $ range, while AB-UPT\textsuperscript{*} with 6.01 million parameters shows the largest RMSE uncertainty of $ \pm 3.2 $. This suggests that architectural design and training procedure influence robustness more than parameter count alone. Overall, the bootstrap analysis supports that the ranking of models by Relative L2 is statistically reliable, while also revealing that even the best models retain non-negligible variability on complex three-dimensional flows in DrivAerNet++.

\subsection*{Statistical Significance Analysis}
Table~\ref{tab:quant-compare} reports performance metrics with 95\% bootstrap confidence intervals, enabling a robust evaluation of model ranking beyond point estimates. AB-UPT\textsuperscript{*} achieves the lowest relative L2 error ($0.1358 \pm 0.0024$) and the highest coefficient of determination. Because the intervals in Table~\ref{tab:quant-compare} are marginal, their overlap or separation is not by itself a test of a difference between two models. All models are evaluated on the same $1{,}154$ geometries, so we compare them pairwise instead, resampling geometries and differencing the two models on each ($B=10{,}000$, stratified by archetype). On this basis AB-UPT\textsuperscript{*} is ahead of TransolverLarge by $0.0099$ Relative L2 ($95\%$ CI $[0.0082, 0.0115]$), and is the more accurate model on $88.6\%$ of individual geometries; it leads TripNet by $0.0265$ ($[0.0245, 0.0284]$, better on $96.7\%$), and TransolverLarge leads TripNet by $0.0166$ ($[0.0141, 0.0190]$, better on $91.2\%$). The paired comparison therefore resolves the ordering of the leading models, which the overlap of marginal $R^2$ intervals alone does not. These intervals reflect resampling of the test geometries for a single training run and do not capture seed-to-seed variability. The next-best model, TransolverLarge, achieves $\text{Rel L2}\, = 0.1457 \pm 0.0025$ and $R^2_{\text{test}} = 0.9595 \pm 0.0069$, followed by Transolver ($0.1503 \pm 0.0024$) and Transolver++ ($0.1573 \pm 0.0023$). TransolverLarge and Transolver have marginal intervals that overlap near $0.148$; distinguishing them would require the same paired comparison used above rather than an inspection of those intervals. In contrast, earlier architectures such as PointTransformer ($0.1909 \pm 0.0024$) and RegDGCNN ($0.2006 \pm 0.0016$) fall well outside the top-performing range, and have intervals that do not intersect with those of the best-performing models. These results confirm that transformer-based models, particularly unified architectures like AB-UPT\textsuperscript{*}, offer statistically robust improvements in predictive fidelity over prior methods.

\subsection*{Statistical Error Distributions and Failure-Case Analysis} Systematically analyzing the worst-performing cases for each model is crucial to identify where the predictions break down and which flow regions, such as high-curvature surfaces, remain most challenging. 
Such analysis provides valuable insights into model limitations and helps guide future improvements in network architecture and data representation. To provide a detailed assessment of model performance beyond average errors, we report both relative and absolute percentile-based error statistics, which together capture accuracy across typical and extreme flow regions. 
The \textit{median relative error} measures the scale-invariant deviation between predicted and true surface pressures, offering a normalized indicator of global model accuracy that is independent of absolute pressure magnitude. 
In contrast, the percentile-based metrics (\textit{P50}, \textit{P90}, \textit{P95}, and \textit{P99}) quantify absolute deviations in physical units (m$^2$/s$^2$), providing an interpretable measure of how far predictions deviate from ground truth across the surface. 
The \textit{P50 error} represents the median deviation per surface point and reflects the model's typical predictive fidelity in low- and moderate-gradient regions. 
The \textit{P90} and \textit{P95} percentiles capture the upper tail of the error distribution, corresponding to more complex aerodynamic regions such as sharp edges, high-curvature regions, and stagnation zones. Finally, the \textit{P99 error} measures the rare but large outliers, typically occurring near high-curvature features or sharp discontinuities in the pressure field.

\begin{table}[h!]
\centering
\small
\setlength{\tabcolsep}{5pt}
\caption{
Statistical error distribution of surface pressure prediction across models, sorted from worst to best according to Median Relative Error. 
Median relative error is dimensionless, while percentile errors are reported in absolute pressure units (m$^2$/s$^2$). 
Lower values indicate higher predictive accuracy. 
Percentiles describe the cumulative distribution of absolute errors across all surface points, capturing both typical and extreme deviations.
As in Table~\ref{tab:quant-compare}, AB-UPT\textsuperscript{*} denotes the checkpoint provided by the original authors, while TripNet refers to our own reimplementation.
}
\label{tab:percentile_errors}
\begin{tabular}{lccccc}
\toprule
\textbf{Model} & 
\textbf{Median Rel. Error} & 
\textbf{P50 Error (m$^2$/s$^2$)} & 
\textbf{P90 Error (m$^2$/s$^2$)} & 
\textbf{P95 Error (m$^2$/s$^2$)} & 
\textbf{P99 Error (m$^2$/s$^2$)} \\
\midrule
PointNet           & 0.19 & 17.57 & 66.04 & 101.32 & 249.43 \\
NeuralOperator                & 0.15 & 14.36 & 53.35 & 82.93  & 192.86 \\
PointMAE           & 0.13 & 13.17 & 48.44 & 72.70  & 167.14 \\
PointNetLarge     & 0.12 & 11.10 & 46.61 & 72.06  & 156.91 \\
RegDGCNN           & 0.10 & 9.91  & 37.03 & 54.15  & 111.75 \\
PointTransformer  & 0.10 & 9.50  & 34.77 & 50.40  & 103.05 \\
Transolver++       & 0.09 & 7.99  & 30.99 & 44.73  & 89.25  \\
TripNet         & 0.08 & 7.55 & 30.77 & 45.18 & 94.09 \\
Transolver         & 0.07 & 6.89  & 27.96 & 40.75  & 83.04  \\
TransolverLarge   & 0.07 & 6.84  & 27.88 & 40.67  & 83.17  \\
AB-UPT\textsuperscript{*}           & 0.06 & 6.20 & 25.40 & 37.90 & 78.30 \\

\bottomrule
\end{tabular}
\end{table}

\noindent
Table~\ref{tab:percentile_errors} shows a clear progression in accuracy from early point-based models to modern operator-, transformer-, and triplane-based architectures. Classical baselines such as PointNet and NeuralOperator exhibit the highest median relative errors ($\geq 0.15$) and large tail errors (P99 $>190$~m$^2$/s$^2$). Geometric deep networks (RegDGCNN, PointTransformer) substantially reduce both median and high-percentile errors, reflecting improved local feature aggregation. Transformer-style solvers and implicit models achieve the strongest overall performance. Transolver, TransolverLarge, and Transolver++ maintain low median errors (0.07--0.09) with P50 errors below $8$~m$^2$/s$^2$. TripNet matches their median error (0.08) but carries a heavier tail (P99 = $94.09$~m$^2$/s$^2$), while AB-UPT\textsuperscript{*} attains the best results overall, achieving the lowest median error (0.06) and smallest extreme deviations (P99 = $78.30$~m$^2$/s$^2$). These results highlight the benefits of global receptive fields and rich feature parameterizations for accurate and robust surface-pressure prediction.

\subsection*{Convergence Behavior of Neural Surrogate Models} 
\begin{figure}[h!]
    \centering
    \includegraphics[width=\linewidth]{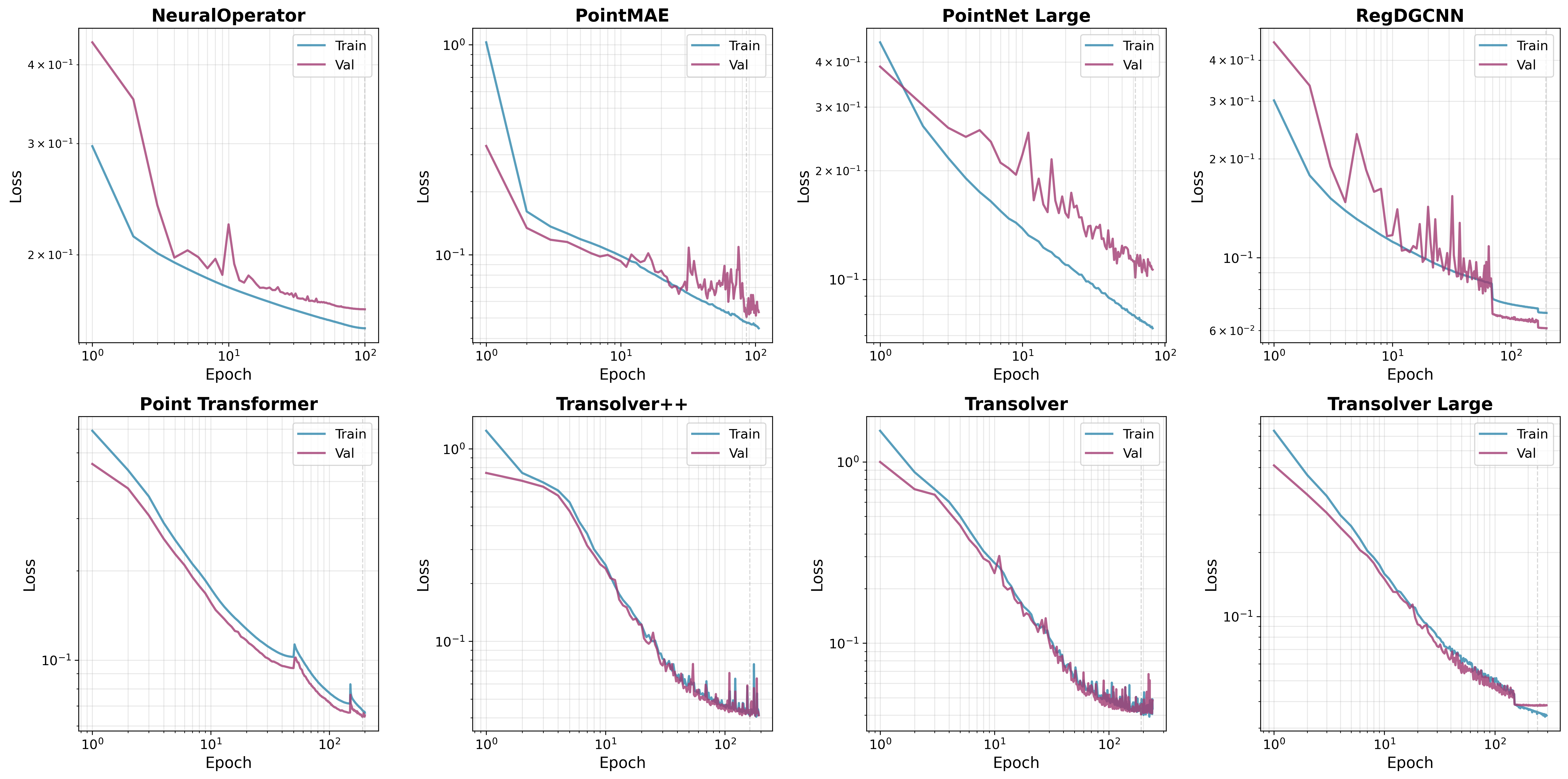}
    \caption{Training and validation loss curves for eight
    models evaluated on the DrivAerNet++ dataset. Each plot shows model
    convergence behavior over training epochs, with blue indicating training
    loss and red indicating validation loss. These trends highlight differences
    in optimization stability and representational efficiency across network
    architectures for aerodynamic learning.}
    \label{fig:all_models_grid_comparison}
\end{figure}

Figure~\ref{fig:all_models_grid_comparison} shows the training and validation
loss trajectories for the representative models in our benchmark. Most
architectures exhibit smooth and stable convergence under the unified training setup, mixed-precision optimization, and early stopping. PointNetLarge is a notable exception, as its persistent gap
between training and validation loss indicates signs of overfitting.
Lightweight encoder--decoder models such as NeuralOperator and PointMAE
converge rapidly within the first 50--80 epochs, though they settle at higher
final losses compared to more expressive architectures.

Attention- and graph-based models, including PointTransformer and RegDGCNN,
require longer training horizons but achieve lower and more stable validation
losses. Among high-capacity models, Transolver, Transolver++, and
TransolverLarge display the smoothest and most consistent optimization
behavior, with Transolver++ reaching strong performance particularly quickly.
Across these architectures, convergence speed, generalization behavior, and final accuracy vary systematically with model capacity. The curves were produced with model-specific loss functions and optimization settings (Table~\ref{tab:ml-config}), so absolute loss magnitudes are not directly comparable across panels; each panel should be read for its convergence behaviour.

\section*{Discussion}
CarBench establishes a physically consistent framework for evaluating machine learning models in aerodynamic prediction. Pressure errors are reported in kinematic units (m$^2$/s$^2$), ensuring consistency with CFD conventions and allowing direct comparison to physically meaningful scales; relative and coefficient-of-determination metrics are dimensionless. 
By combining predictive accuracy, computational efficiency, and scalability within a unified assessment framework, the benchmark characterizes how different modeling paradigms balance precision, speed, and resource requirements. 

A key insight from this work is the critical role of large-scale and diverse training datasets in enabling robust generalization to unseen designs. 
Cross-category experiments show that models trained on larger and more geometrically varied data exhibit stronger zero-shot performance, although the evaluated splits cannot separate the contributions of dataset size and geometric diversity. 
Equally important is the choice of evaluation resolution: unlike prior studies that report results only on subsampled point clouds or decimated meshes, CarBench systematically evaluates both subsampled predictions and full-mesh resolution, ensuring that model performance reflects the fidelity of the original high-resolution CFD data. 
This approach provides a fair and realistic measure of physical accuracy and generalization beyond the limitations of reduced data representations. In addition to numerical evaluation, the benchmark incorporates qualitative visualization and error analysis, revealing the flow regions where models tend to struggle, such as stagnation regions, high-curvature surfaces, and areas with strong pressure gradients. 
These visual analyses help identify specific geometric or flow regimes that require denser sampling or model refinement. 
Finally, the qualitative component-level comparisons, such as the rotating-wheel region, illustrate the benchmark's capacity to expose model behaviour under detailed boundary conditions and localized flow effects; we report these as visual comparisons rather than as component-specific error metrics.

Beyond ranking models, the benchmark permits some analysis of \emph{why} the ranking arises, and two factors recur. The first is how much geometric information reaches the network. A controlled experiment that holds the Transolver architecture and training schedule fixed and varies only the inputs (Supplementary Table~5) is, however, inconclusive on this point: neither supplying a signed distance field nor adopting the best architecture found by an extensive sweep over depth, width and attention slices changes accuracy by more than the difference between two runs of the same configuration. We therefore report the association below as an observation about the ranking rather than as a demonstrated cause. The models that rank highest in Table~\ref{tab:quant-compare} are those whose inductive bias grants the widest access to geometric context: AB-UPT\textsuperscript{*} through supernode pooling and anchored attention over the whole surface, TripNet through a volumetric triplane encoding, and the Transolver family through global slice attention. Purely local architectures such as PointNet, whose per-point features are aggregated only through a global max-pool, sit at the opposite end of both the representation and the accuracy ordering.

The second factor is how a model is queried at evaluation resolution, which affects measured accuracy as much as the architecture itself. Models with a query-based or implicit decoder evaluate any surface point directly, whereas models that consume a fixed-size token set must either interpolate from a subsample or be applied patchwise. Quantifying this (Table~\ref{tab:surface_fullmesh_comparison}) shows that most of the degradation previously attributed to resolution is interpolation error, and that it largely disappears once every node is predicted directly. Evaluation protocol and architecture are thus entangled, and a benchmark that fixes one protocol for all models risks measuring the protocol rather than the model.

These observations suggest concrete directions for improvement. For point- and graph-based baselines, supplying volumetric context such as a signed distance field is an obvious candidate, though our controlled experiment (Supplementary Table~5) does not detect a benefit from it at the scale tested, and establishing whether one exists would require multi-seed training with an uncontaminated validation criterion. For the spectral baseline, replacing the fixed voxelization with a geometry-aware operator such as Geo-FNO~\cite{li2023fourier} or GINO~\cite{li2023geometry} would test whether its deficit stems from the spectral inductive bias itself or merely from the crude geometry-to-grid mapping. For transformer-based solvers, decoupling the number of supervised points from the number of evaluated points, through query-based decoding of the kind AB-UPT employs, would remove the resolution penalty entirely. Progress on high-fidelity aerodynamic surrogates is therefore likely to come as much from how geometry is encoded and queried as from further scaling of the architectures themselves. 

Beyond cross-model comparison, we also analyzed the effect of model capacity on performance. We tested two architectures in both their baseline and enlarged configurations: PointNet and Transolver. For PointNet, increasing model size from 1.67M to 32.58M parameters reduced the relative L2 error from 0.3803 to 0.2436, a substantial improvement that highlights the benefit of higher representational capacity for complex aerodynamic surfaces. In contrast, scaling Transolver from 2.47M to 7.58M parameters led to only a modest improvement in relative L2 error---from 0.1503 to 0.1457---an absolute reduction of $0.0046$, or about $3\%$ in relative terms. This suggests diminishing returns at higher capacity levels for already well-performing transformer architectures, and highlights that architectural design and training strategy may be more influential than raw parameter count beyond a certain threshold. These results underscore the importance of balancing model capacity with data efficiency to maximize both physical accuracy and practical utility.

We further investigated different geometric representations. 
While most models evaluated in this benchmark are point-cloud-based, we also included architectures that leverage alternative spatial encodings. 
TripNet employs a triplane implicit neural representation that maps the 3D car surface into orthogonal feature planes, enabling dense reconstruction of aerodynamic quantities, whereas RegDGCNN constructs dynamic graphs over sampled point clouds to capture local neighborhood relations and surface connectivity. 
Our experiments show that for the task of predicting surface pressure distributions, point cloud-based representations, when densely sampled, are highly reliable and faithfully capture the underlying aerodynamic fields on complex car geometries. This finding suggests that high-resolution surface sampling provides sufficient spatial information for accurate flow reconstruction without requiring volumetric representations for steady-state external aerodynamics. Together, these elements establish CarBench as a rigorous, interpretable, and scalable foundation for advancing data-driven research in computational aerodynamics. 
By promoting standardized evaluation in physically meaningful units, this benchmark lays the groundwork for reproducible progress in neural CFD modeling and for the next generation of generalizable, data-driven aerodynamic solvers.

We summarize the main findings of our evaluation across architectures, datasets, and geometric representations:
\begin{itemize}
    \item \textbf{Dataset scale and diversity.} Larger and more varied training sets are associated with stronger cross-category performance in the evaluated splits. Models trained on large and geometrically varied datasets outperform those trained on narrow distributions, underscoring a core strength of the DrivAerNet++ dataset.
    \item \textbf{Transformer-based models} (e.g., TransolverLarge, Transolver, and AB-UPT\textsuperscript{*}) consistently achieve state-of-the-art results, demonstrating strong generalization and spatial coherence in pressure field prediction.
    \item \textbf{Evaluation at full mesh resolution} is critical for realism, and the protocol used to reach it matters more than the resolution itself: most of the degradation previously attributed to full-resolution evaluation is interpolation error, and it largely disappears when every node is predicted directly (Table~\ref{tab:surface_fullmesh_comparison}); our framework emphasizes fidelity by comparing against high-resolution CFD ground truth across the entire surface mesh.
    \item \textbf{Representation matters.} While implicit triplane and graph-based networks provide complementary strengths, high-resolution point-cloud representations remain highly effective for surface pressure prediction.
\end{itemize}

\section*{Conclusions}

This work introduces a unified, model-agnostic framework for training and evaluating neural surrogates for automotive aerodynamics. 
Spanning eight car archetypes, thousands of high-fidelity steady RANS simulations, and eleven SOTA deep learning models, CarBench enables reproducible comparison across diverse model configurations, with pressure errors reported in accordance with CFD conventions. 
Our standardized pipeline, covering data preprocessing, mixed-precision optimization, early stopping, and a common evaluator, helps ensure that, for models retrained in our framework, measured gains largely reflect architectural differences rather than experimental confounders. Across extensive experiments, among the models evaluated, transformer-based neural solvers and operators achieve the strongest overall combination of accuracy and scalability, with AB-UPT\textsuperscript{*} and TransolverLarge achieving the highest overall accuracy and Transolver++ offering an especially favorable accuracy--efficiency trade-off. 
Analyses on full mesh evaluations, rotating wheel aerodynamics, and cross-category generalization highlight three central findings: (i) evaluating at CFD resolution is essential for realistic error accounting, (ii) component-level comparisons qualitatively expose localized error concentrations around high curvature and separation zones, and (iii) the experiments show a strong association between training-set composition, training-set size, and zero-shot performance on unseen archetypes.

While machine learning applications in industrial aerodynamic design are still in their early stages, it is important to acknowledge that CFD itself provides an approximation of experimental reality. 
For the DrivAerNet++ dataset, steady-state RANS baselines show deviations below 3\% for the Fastback archetype and below 5\% for the Estateback and Notchback configurations compared to wind tunnel measurements~\cite{elrefaie2024drivaernet++}. 
Generating the DrivAerNet++ dataset required an immense computational investment of more than three million CPU hours distributed across eight archetypes and more than eight thousand simulations, highlighting the prohibitive cost of large-scale CFD data generation. 
In contrast, training the benchmarked surrogate models requires less than a few hundred GPU hours per architecture, enabling rapid experimentation and model iteration at a fraction of the cost. 
Moreover, the cross-category generalization experiments, where models trained on one car archetype were evaluated on unseen categories, demonstrate that scaling such datasets across multiple DrivAerNet++ variants could otherwise require tens or even hundreds of thousands of additional simulations, corresponding to tens of millions of additional core hours. 
This further emphasizes the necessity of robust and generalizable machine learning surrogates.

The best-performing models, AB-UPT\textsuperscript{*} and TransolverLarge, achieve relative L2 errors of approximately 13--15\%. 
These discrepancies should therefore be interpreted with care. The benchmark metrics compare predictions against the CFD labels alone and therefore measure surrogate error only; any deviation of the simulations themselves from physical reality is a separate source of error that these numbers do not capture. The underlying simulations carry their own approximations arising from CFD modelling and the surrogate modeling process. 
Nevertheless, large-scale datasets based on RANS remain the only practical path toward training machine learning models at the tens or hundreds of thousands of simulation scale required for generalization and foundation modeling. 
Accordingly, the goal of this benchmark is not merely to replicate CFD but to catalyze the development of more accurate, robust, and physically grounded surrogates that can ultimately bridge the gap between numerical simulation and real-world aerodynamics. 
We believe that CarBench provides a critical foundation for advancing three-dimensional machine learning for large-scale numerical simulations, promoting fair comparison, reproducibility, and continued progress in data-driven aerodynamic design. While this work focuses on car aerodynamics, the benchmark framework is generalizable to other domains where the objective is to learn flow fields conditioned on 3D geometry. Potential applications include external aerodynamics of aircraft, turbomachinery design, and other geometry-dependent fluid dynamics systems.

In summary, CarBench establishes a foundation for reproducible benchmarking of neural CFD surrogates, clarifies the strengths of modern transformer/operator architectures, and charts a practical path toward generalizable, data-driven aerodynamic solvers for real-world design applications. For generative and inverse design, where a surrogate replaces the solver inside the optimization loop, the error bars reported here quantify how precisely the average accuracy of competing surrogates can be compared.

\section*{Methods}
\subsection*{DrivAerNet++ Dataset}

\begin{figure}[h!]
    \centering
    \includegraphics[width=\linewidth]{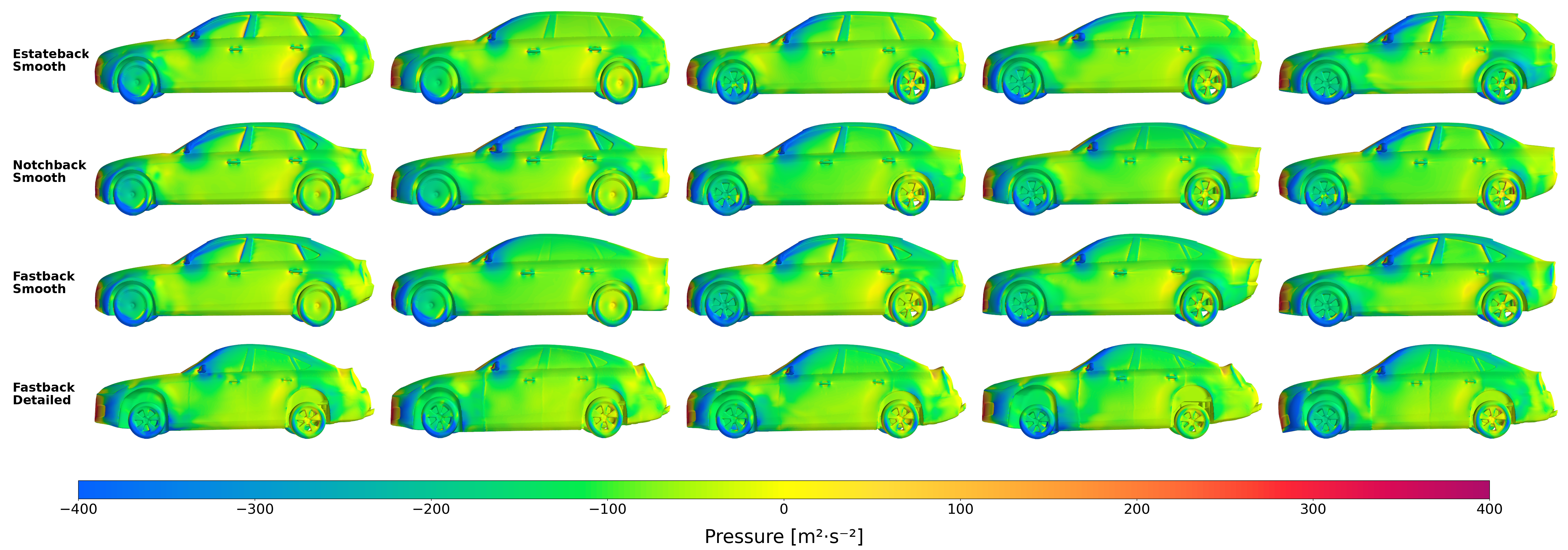}
    \caption{Surface pressure distributions for representative designs across different car categories in DrivAerNet++. 
    Rows correspond to Estateback, Notchback, and Fastback archetypes under smooth and detailed configurations. 
    The color scale represents kinematic pressure (m$^2$/s$^2$), highlighting regions of low pressure over the windshield and roof and high pressure near the front bumper and rear stagnation zones. 
    These examples illustrate the geometric and aerodynamic diversity captured in the dataset.}        \label{fig:surface_pressure_examples}
\end{figure}

DrivAerNet++ represents a significant advancement in automotive aerodynamics datasets, expanding upon the original DrivAerNet with enhanced geometric diversity and higher-fidelity simulations. The dataset comprises 8,150 parametrically generated car configurations, each representing a physically realistic automotive design within the constraints of modern passenger cars. The scale, fidelity, and multimodal structure of DrivAerNet++ have enabled a range of downstream research applications. These include \textit{TripOptimizer~\cite{vatani2025tripoptimizer},} a fully differentiable framework for three-dimensional shape optimization and inverse design; a multi-agent \textit{Design Agents} system~\cite{elrefaie2025ai} that integrates aesthetic and aerodynamic objectives; and surrogate models for accurate prediction of flow fields and aerodynamic coefficients. Collectively, these applications demonstrate the dataset's role as a foundation for developing and benchmarking advanced methods in data-driven design and scientific machine learning. Representative surface pressure distributions across multiple car archetypes and configurations in the DrivAerNet++ dataset are shown in Figure~\ref{fig:surface_pressure_examples}. 
Distinct aerodynamic behaviors are observed between the Estateback, Notchback, and Fastback designs, as well as between smooth and detailed configurations that differentiate electric and internal combustion vehicles. 

\begin{figure}[h!]
    \centering
    \includegraphics[width=\linewidth]{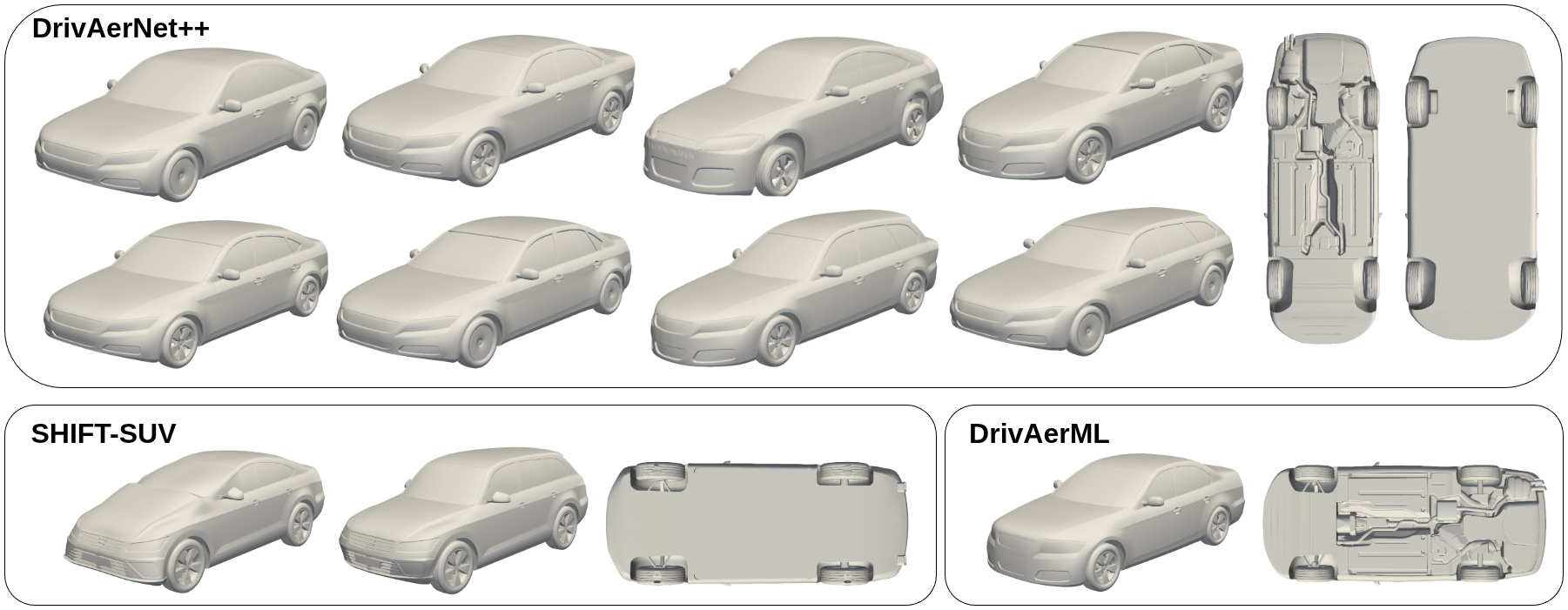}
\caption{Comparison of large-scale, high-fidelity, and industry-standard open-source car datasets. 
DrivAerNet++ (8,150 designs, based on \textbf{eight parametric models}) provides the broadest coverage, modeling both detailed underbodies representative of internal combustion engine (ICE) cars and smooth underbodies for electric vehicles (EVs). 
It is also the only dataset to incorporate wheel and tire shape variations, including open and closed wheel designs. 
DrivAerML ($\sim$484 designs) is derived from a single DrivAer parametric model, while SHIFT-SUV ($\sim$2,000 designs) is based on two AeroSUV baseline models.}
    \label{fig:CarsDatasets}
\end{figure}

\begin{figure}[h!]
    \centering
    \includegraphics[width=\linewidth]{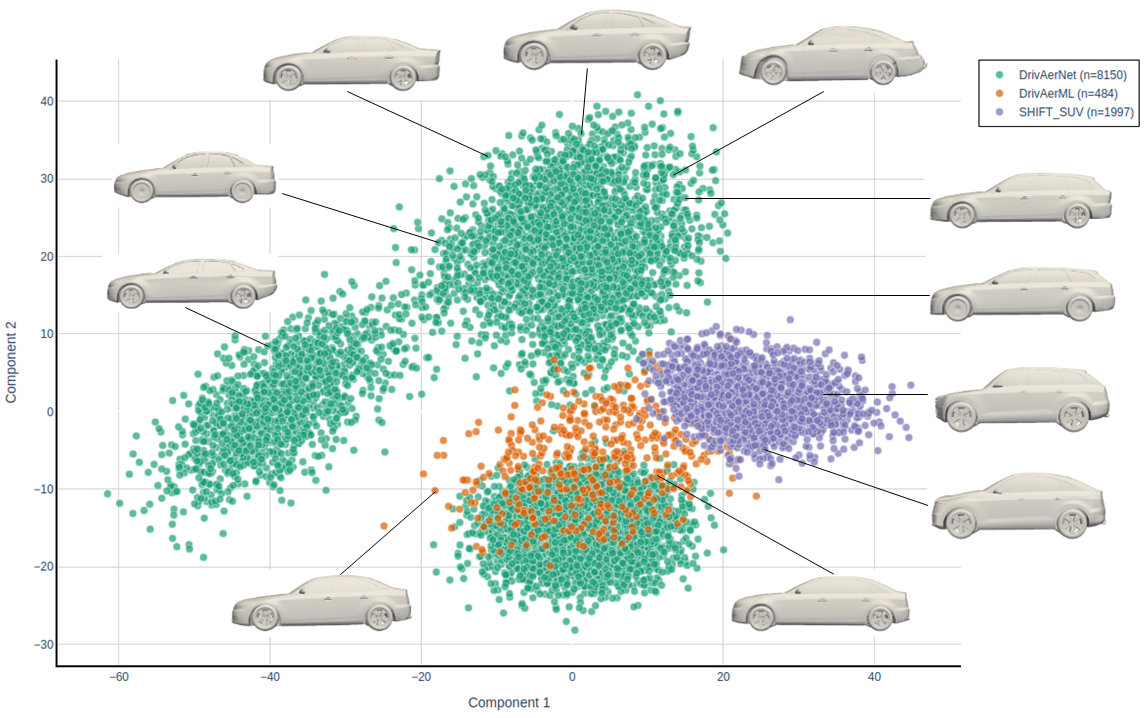}
\caption{PCA projection of dataset distributions across DrivAerNet++, DrivAerML, and SHIFT-SUV. 
Each point represents a car geometry, with DrivAerNet++ spanning a wider and more continuous design space, 
while DrivAerML and SHIFT-SUV are confined to narrower regions. 
Notably, DrivAerML lies largely within the coverage of DrivAerNet++, as it is derived from the DrivAer Notchback baseline, one of the eight parametric archetypes used to construct DrivAerNet++. 
This highlights the geometric diversity and design-space coverage of DrivAerNet++.}
    \label{fig:pca_datasets}
\end{figure}

A comparison of major open-source automotive aerodynamics datasets in terms of scale, geometric diversity, and simulation fidelity is presented in Figure~\ref{fig:CarsDatasets}. 
DrivAerNet++, comprising 8,150 car designs generated from eight parametric base archetypes, exhibits the broadest geometric coverage among existing datasets. 
It includes both detailed underbodies representative of internal combustion vehicles and smooth configurations typical of electric vehicles, alongside diverse wheel and tire variants (open, closed, detailed, and smooth) that are critical for accurately modeling wake dynamics. 
All simulations were performed using steady Reynolds-Averaged Navier--Stokes (RANS) computations with the $k$--$\omega$ SST turbulence model~\cite{menter2003ten}, ensuring consistent aerodynamic fidelity across thousands of designs. In contrast, datasets such as SHIFT-SUV~\cite{luminarycloud_shift_suv_2025}, derived from only two parametric AeroSUV  archetypes~\cite{ecara_drivaer_reference} and employing transient Detached Eddy Simulation (DES), and DrivAerML~\cite{ashton2024drivaerml}, based on a single DrivAer geometry~\cite{heft2012introduction,tum_drivaer_geometry} simulated using Hybrid RANS--LES (HRLES), offer limited geometric and categorical diversity. 
DrivAerNet++ provides a complementary balance, roughly sixteen times larger than DrivAerML and four times larger than SHIFT-SUV, while maintaining standardized steady-state RANS fidelity suitable for large-scale benchmarking and model comparison. To further examine geometric variability, Figure~\ref{fig:pca_datasets} visualizes dataset distributions using Principal Component Analysis (PCA). 
Each point represents a distinct car geometry embedded in a reduced shape feature space, revealing clear differences in design-space coverage across datasets. 
DrivAerNet++ spans a broad and continuous region of the design manifold, capturing smooth morphological transitions between archetypes. 
By comparison, DrivAerML occupies a compact subregion almost entirely contained within the DrivAerNet++ distribution, consistent with its origin from the DrivAer Notchback baseline. 
SHIFT-SUV forms a distinct, narrow cluster reflecting production-grade SUV geometries with limited topological variation.

Overall, DrivAerNet++ achieves the most extensive geometric coverage among publicly available automotive CFD datasets, encompassing both continuous intra-category variations and distinct inter-category separations across Fastback, Estateback, and Notchback configurations. 
This diversity ensures that trained models encounter a wide spectrum of aerodynamic phenomena, from roof curvature and underbody shaping to wheel enclosure design, enabling more robust generalization to unseen geometries.

Because a benchmark is ultimately characterized by its \emph{test} split rather than by the dataset as a whole, we repeat this embedding on held-out geometries alone in the Supplementary Information (Supplementary Figure~1). The structure is unchanged, and the accompanying analysis relates test-set size to the statistical precision with which generalization can be measured.

\subsection*{Data Processing and Quality Assurance}

\paragraph{Train/Validation/Test Splits.}
For consistency and reproducibility, we adopt the official train/validation/test split provided in the DrivAerNet++ repository at \url{https://github.com/Mohamedelrefaie/DrivAerNet/tree/main/train_val_test_splits}. Of the $8{,}150$ designs in DrivAerNet++, $8{,}121$ have complete surface-pressure solutions and constitute the benchmark splits used throughout. These predefined splits ensure that evaluation is performed strictly on unseen geometries, preventing any overlap between training and test samples across the full parametric design space.

\paragraph{Pressure Statistics Computation.}
To prevent data leakage, pressure normalization statistics are computed exclusively from the DrivAerNet++ training set containing 5,819 samples. The computation pipeline processes all VTK files corresponding to training design IDs, loading each file using PyVista~\cite{sullivan2019pyvista} and extracting pressure field data from the computational mesh points. All pressure values from the training files are aggregated into a single array, from which population statistics are computed, including the mean $\mu_p$ and standard deviation $\sigma_p$. Additional distribution characteristics, such as minimum, maximum, median, and quartiles, are also calculated. These training-only statistics provide the normalization parameters $(\mu_p, \sigma_p)$ used throughout the neural network pipeline. Importantly, all evaluation metrics, including relative L1 and L2 errors, are computed in the denormalized (original physical) space to ensure meaningful comparisons across different models and datasets, while maintaining strict separation between training and validation/test statistics.

\begin{figure}[h!]
  \centering
\includegraphics[width=\linewidth]{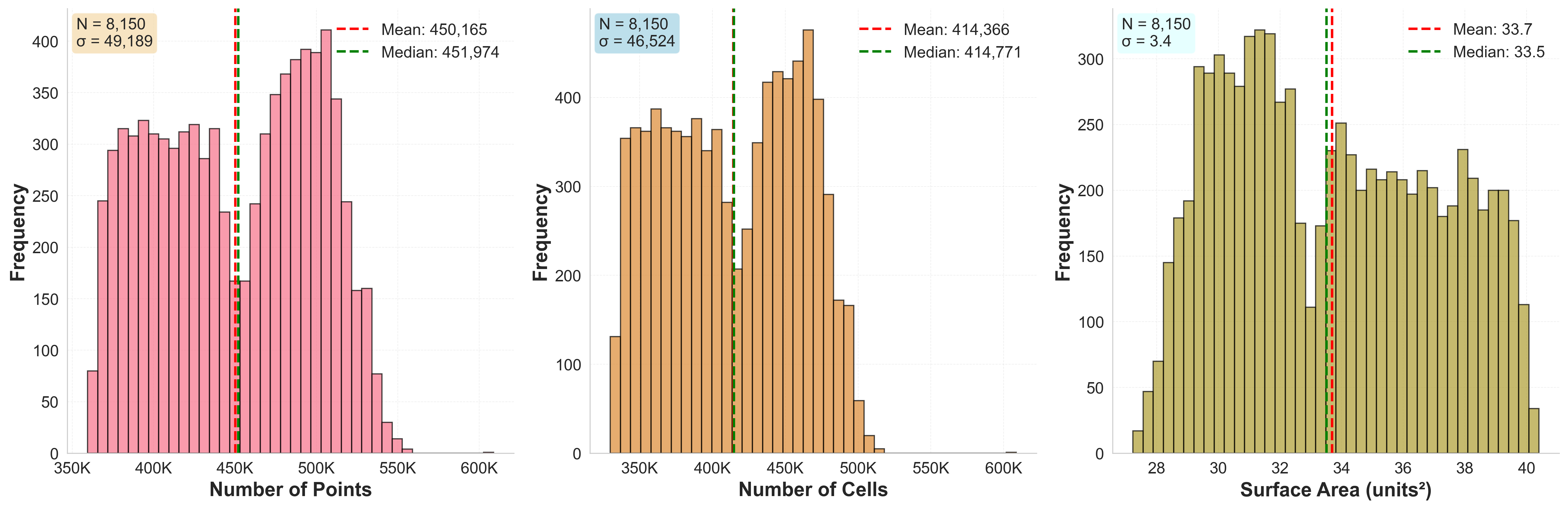}
\caption{
DrivAerNet++ dataset geometric statistics. 
Left: Distribution of point counts across all meshes, showing a consistent resolution with mean and median values around 450K. 
Middle: Distribution of cell counts, exhibiting a similar bimodal trend and central tendency to the point distribution. 
Right: Distribution of total surface area, indicating moderate variation across designs with a mean of 33.7~units\textsuperscript{2}. 
Each subplot reports dataset size ($N = 8{,}150$) and standard deviation, providing an overview of mesh density and geometric uniformity across the corpus.
}
\label{fig:vtk-geom-analysis}
\end{figure}

Figure~\ref{fig:vtk-geom-analysis} highlights the substantial geometric complexity captured in the DrivAerNet++ surface meshes. The consistently high point and cell counts, often reaching several hundred thousand per design, reflect a high-fidelity surface discretization that preserves subtle geometric features across all vehicle classes. This level of detail creates a challenging learning setting because models must generalize across a wide spectrum of shapes with significant variability in proportions, curvature, and surface topology. This geometric diversity closely mirrors the exploration space encountered in early conceptual design stages, where broad shape variations must be evaluated efficiently. By combining high-resolution fidelity with large inter-design variability, DrivAerNet++ provides a rigorous testbed for data-driven aerodynamic prediction and design space learning.

\section*{Automated Training and Evaluation Framework}

A central goal of this work is to establish a standardized, transparent, and easily extensible framework for evaluating machine learning models in large-scale CFD predictions. 
Despite rapid progress in this domain, inconsistent data preprocessing, metric definitions, and training protocols have often hindered fair comparison across studies. 
To address this, we designed an automated benchmark pipeline that ensures full reproducibility and uniform evaluation across architectures, ranging from classical geometric deep learning models to neural operators and implicit field networks. 
The framework is configuration-driven, modular, and open-source, enabling both state-of-the-art replication and straightforward integration of new models. We developed an automated and model-agnostic framework to train and evaluate all machine learning baselines in a unified and reproducible manner. 
Each model is encapsulated within a modular wrapper that standardizes data preprocessing, normalization, and forward-pass interfaces, ensuring consistent input-output behavior across architectures. 
The framework is built around a \textit{UnifiedTrainer} class that manages key aspects of training, including mixed-precision computation for efficiency, gradient clipping for stability, adaptive learning rate scheduling, and early stopping based on validation metrics. 
All runs automatically log losses, learning rates, and generalization gaps throughout training to facilitate transparent comparison.

\begin{table}[h]
\centering
\caption{Training and model configurations for all ML baselines, as executed in the runs reported in Table~\ref{tab:quant-compare}. Epoch entries give the executed count, with the configured budget in parentheses where early stopping intervened; learning-rate entries give the initial and, where a schedule was active, final values. All models are trained on 10{,}000 points per surface sample with mean squared error loss except where noted.}
\label{tab:ml-config}
\small
\setlength{\tabcolsep}{3pt}
\resizebox{\linewidth}{!}{%
\begin{tabular}{l|ccccccc|cccccc}
\hline
\multicolumn{1}{c|}{\textbf{Model}} &
\multicolumn{7}{c|}{\textbf{Training Configuration}} &
\multicolumn{6}{c}{\textbf{Model Architecture}} \\
\cline{2-14}
& Loss & Epochs & LR & Opt & Batch & WD & Scheduler &
Layers & Heads & Embed & Dropout & $k$-NN & Extra \\
\hline
PointNet            & MSE & 58 (150) & $10^{-3}\!\to\!10^{-5}$ & Adam   & 12  & $10^{-4}$ & ReduceLR & -- & -- & 1024 & 0.3 & -- & -- \\
NeuralOperator      & MSE & 100 & $2{\times}10^{-3}\!\to\!5{\times}10^{-7}$ & AdamW & 16  & $10^{-4}$ & Cosine & 2 & -- & 16  & -- & -- & Grid=$32^3$, $M$=8 \\
PointMAE            & SmoothL1 & 106 (500) & $10^{-3}$ & AdamW & 128 & $10^{-2}$ & Cosine$^{\ddagger}$ & 4 & -- & 256 & 0.3/0.2 & -- & Efficient \\
PointNetLarge       & MSE & 82 (150) & $4{\times}10^{-4}$ & AdamW  & 8   & $10^{-4}$ & OneCycle$^{\ddagger}$ & -- & -- & 4096 & 0.3/0.5 & -- & Residual \\
RegDGCNN            & MSE & 200 & $10^{-3}\!\to\!10^{-5}$ & Adam   & 12   & $10^{-4}$ & ReduceLR & 3  & -- & 1024 & 0.4 & 40 & EdgeConv \\
PointTransformer    & MSE & 200 & $10^{-4}\!\to\!8.6{\times}10^{-5}$ & AdamW  & 16   & $10^{-2}$ & CosineWU & 12 & -- & 128 & 0.1 & 16 & Local attn \\
TripNet             & MSE & 200 & $10^{-3}$ & Adam  & 8   & 0     & --       & 4  & -- & 256 & -- & -- & Triplane $32{\times}128^2$ \\
Transolver++        & MSE & 192 (200) & $10^{-3}$ & Adam  & 128   & 0     & --$^{\ddagger}$   & 5  & 8  & 256 & 0.0 & -- & 32 slices, UPE \\
Transolver          & MSE & 241 (1000) & $10^{-3}$ & Adam  & 128  & 0     & --$^{\ddagger}$   & 5  & 8  & 256 & 0.0 & -- & 32 slices \\
TransolverLarge     & MSE & 294 (1000) & $10^{-4}\!\to\!10^{-5}$ & Adam  & 64   & 0     & StepLR   & 7  & 12 & 384 & 0.1 & -- & 48 slices \\
AB-UPT              & MSE & 30  & $5{\times}10^{-5}$ & Lion & 1 & 0.05 & CosineWU & 11 & 3 & 192 & -- & -- & EMA, 1024 supernodes \\
\hline
\end{tabular}%
}

\vspace{2pt}
\begin{minipage}{\linewidth}
{\footnotesize \textbf{Notes.}
(1) \textbf{WD}: weight decay; \textbf{Opt}: optimizer; \textbf{UPE}: unified position encoding.
(2) \textbf{Schedulers}: ReduceLR = ReduceLROnPlateau (patience 10, factor 0.1); Cosine ($T_{\text{max}}$ = configured budget); CosineWU = cosine with warmup; StepLR (step 150, $\gamma$ = 0.1).
(3) Supplementary Table~3 reports the same models under a single shared optimization budget.
(4) The models trained use gradient clipping (norm $1.0$) and early stopping (patience $20$ for point-based models, $30$ for transformers, $50$ for TransolverLarge); AB-UPT clips at $0.25$, and TripNet uses neither. Mixed precision is enabled for all models except NeuralOperator and TransolverLarge.
(5) TripNet is trained in two stages: the triplane occupancy field is fitted first ($500$ epochs jointly over $500$ cars, then $70$ epochs per remaining car; Adam, $10^{-3}$), after which the pressure predictor listed here is trained on the cached triplanes with $200{,}000$ query points per car.
(6) The AB-UPT row lists the settings of the from-scratch control in Supplementary Table~6, which match the architecture and training budget of the released checkpoint reported in Table~\ref{tab:quant-compare}: Lion with $5\%$ linear warmup and cosine decay to $10^{-6}$, gradient clipping at $0.25$, and an exponential moving average (decay $0.9999$) whose weights are used at evaluation. Each sample comprises $16{,}384$ geometry points, $1{,}024$ supernodes, and $512$ surface anchors; the $10{,}000$-point sampling stated in the caption does not apply to AB-UPT or TripNet.
(7) Architectural details for each baseline are given in the Baseline Models section of the Supplementary Information.
(8) $^{\ddagger}$The scheduler entry reflects the executed run; the released configuration file for this model names a different schedule or base learning rate.}
\end{minipage}
\end{table}

For evaluation, a dedicated \textit{UnifiedEvaluator} computes a consistent suite of quantitative metrics, including Mean Absolute Error (MAE), Root Mean Square Error (RMSE), Relative L2 error, and inference latency. 
These metrics are tracked across models and datasets under identical sampling conditions to ensure a fair comparison of predictive performance and computational efficiency. 
New architectures can be integrated by registering additional models through standardized configuration files, without modifying core training scripts. 
This automation supports reproducible experiments and allows future models to be evaluated under the same protocol. Table~\ref{tab:ml-config} summarizes the architectural and training configurations of all machine learning baselines used in the benchmark. 
All models share the same train/validation/test split, the same target normalization and the same evaluator. Optimization settings follow each model's published training configuration and are reported in Table~\ref{tab:ml-config}; they are not identical across models, and the effect of imposing a single shared budget is examined separately in Supplementary Table~3. 
For the subset of baselines that we retrain within our pipeline (Table~\ref{tab:ml-config}), this standardization substantially reduces confounding from training setup and makes performance differences easier to attribute to architectural choices. AB-UPT\textsuperscript{*} is evaluated using the publicly released checkpoint from its original authors, so its result should be interpreted as best-effort but not strictly normalized to the same training budget as the in-house models; the architecture is additionally retrained here under our own pipeline (Supplementary Table~6). TripNet is reported solely as retrained in this work, since no public TripNet implementation or checkpoint is available.

\paragraph{Design Principles.}
The benchmark was built around three guiding principles: (1) \textit{standardization:} all models are trained and evaluated on the same data splits with the same normalization and metric definitions; (2) \textit{reproducibility:} every experiment is defined via declarative configuration files with fixed random seeds and logged metadata; and (3) \textit{accessibility:} new architectures can be integrated with minimal code overhead.

\paragraph{Unified Experimental Pipeline.}
Experiments are defined in YAML configuration files specifying dataset splits, sampling strategies, optimizer settings, learning rate schedulers, and early-stopping criteria. 
All models are scored on the same $10{,}000$ uniformly sampled surface points per geometry. The two models with query-based decoders additionally receive their own geometry encoding as input, as recorded in Table~\ref{tab:ml-config}.

\paragraph{Standardized Evaluation Metrics.}
The benchmark computes a unified set of metrics for all models, covering absolute errors (MAE, MSE, RMSE), relative errors (Relative L2), and statistical measures (median relative error and P50/P90/P95/P99 percentiles of absolute error). 
All quantities are expressed in physically meaningful units (m$^2$/s$^2$ for kinematic pressure). 
Model performance is further analyzed in terms of $R^2_{\text{test}}$, computational throughput (samples/s), memory footprint (GB), and latency (ms), under matched batch size and GPU configurations.

\paragraph{Automation and Reporting.}
Training, validation, and evaluation are executed via a single command-line interface, automatically generating structured logs, plots, and result tables in CSV and JSON formats. 

\paragraph{Extensibility and Community Contribution.}
The benchmark is intentionally designed for extensibility. 
Researchers can contribute new architectures, loss functions, or evaluation tasks by adding configuration files and registering modules without altering the core codebase. 
All newly added metrics are automatically incorporated into the final evaluation pipeline. 
This modular design encourages collaborative benchmarking and continuous expansion of the aerodynamic learning ecosystem, fostering an open, evolving standard for machine learning in CFD.

\paragraph{Preprocessed Data Release.} To facilitate accessibility and accelerate experimentation, we also release the preprocessed version of the DrivAerNet++ dataset. 
The data are organized in a standardized, serialized file format, containing normalized surface pressure and geometric point-cloud representations for all samples. 
This version significantly reduces the need for manual preprocessing or CFD file parsing, enabling researchers to directly load the dataset into training pipelines with minimal overhead. 
By providing ready-to-use data splits and consistent normalization statistics, the preprocessed release supports faster training, reproducibility, and integration with new models.

\section*{Limitations and Future Work}

While CarBench represents a significant step toward standardized and high-fidelity aerodynamic surrogate evaluation, several limitations must be acknowledged. The current evaluation protocol focuses primarily on pointwise error metrics such as Rel L2 and MAE, along with computational efficiency measures. Although these metrics are essential for benchmarking, they do not directly quantify discrepancies in global aerodynamic quantities. Future extensions of the benchmark will incorporate physically meaningful integrals, including lift and drag coefficients, integrated force vectors, and descriptors of flow topology, to better assess the practical relevance of surrogate predictions in design tasks. 

A further limitation concerns how generalization is quantified. All simulations in DrivAerNet++ are performed under a single set of operating conditions, so the dataset varies vehicle \emph{geometry} while the flow regime, inlet velocity and boundary conditions are held fixed. Consequently the cross-category experiments reported here isolate geometric extrapolation: they measure whether a model trained on one family of rear-end shapes transfers to another, but they do not probe generalization across Reynolds numbers, yaw angles, or unsteady flow regimes. Disentangling these two axes would require explicit metrics for geometric diversity, for instance shape-descriptor or latent-space coverage of the training set, alongside metrics for physical diversity such as the spread of flow regimes and wake topologies represented. Because no such metrics are established for automotive aerodynamics, we report generalization qualitatively by archetype rather than as a calibrated distance, and we regard the development of quantitative geometry- and physics-diversity measures, together with multi-condition simulations that vary the flow regime, as an important direction for future benchmarks.

A further limitation relates to uncertainty quantification in model training. Due to the significant computational cost associated with training eleven high-capacity models on more than five thousand samples, each model in this study was trained only once. This means that run-to-run stochastic variability, including differences caused by random initialization or optimizer noise, is not captured. In many benchmarking settings, multiple independent training runs are performed to quantify this variance and ensure that conclusions are not influenced by unusually favorable or unfavorable single runs. We therefore report the single-run protocol as a limitation of the present study, and future benchmark iterations may incorporate multi-run analyses as computational resources allow.

CarBench targets one class of problem, incompressible external aerodynamics of passenger-car geometries at a fixed operating point, and additional benchmarks are needed for other regimes and target fields, including wall-shear stress and skin-friction prediction, compressible flows with shocks, slender lifting surfaces, and other complex flow structures. CarBench should therefore be viewed as a complementary benchmark for one important class of realistic three-dimensional CFD problems rather than a complete benchmark for aerodynamic surrogate modeling.

Overall, we envision CarBench as a living framework that will continue to expand in scope and fidelity. Future developments may include additional physical tasks, broader flow regimes, diverse design domains, and deeper integration with design optimization and real-time simulation pipelines. We invite the community to extend, refine, and build upon this resource to advance robust, generalizable, and physically grounded machine learning for fluid dynamics.

\section*{Community Contributions and Future Extensions}

We encourage contributions from the research community to extend this benchmark with additional models and learning tasks. 
In particular, future releases will integrate recently proposed architectures that have demonstrated promising performance, such as \textit{LRQ-Solver}~\cite{zeng2025lrq}, \textit{FigConvNet}~\cite{choy2025figconvnet}, \textit{GAOT}~\cite{wen2025geometry}, and \textit{AeroGTO}~\cite{liu2025aerogto}, as well as expand beyond surface pressure prediction to include full volumetric flow field estimation. 
Through continuous collaboration and open benchmarking, we aim to establish a long-term, evolving standard for data-driven aerodynamic learning.

\section*{Acknowledgments}

We thank Johannes Brandstetter and Richard Kurle for sharing the pretrained AB-UPT model. We are especially grateful to Richard Kurle for his thoughtful feedback and valuable insights on benchmarking methodologies and the evaluation of machine learning models for aerodynamic prediction. 
We also thank Qian Chen for assisting with benchmarking the TripNet model. 
Finally, we thank Luminary Cloud and Michael Emory for providing access to the SHIFT-SUV dataset.

\section*{Funding}

\noindent
This work was supported by the Toyota Research Institute (TRI).

\section*{Author Contributions Statement}
M.E. conceived the study, developed the benchmark framework, implemented all experiments, performed the analyses, and prepared the manuscript. 
D.S. and M.K. contributed to discussions related to the project scope and provided feedback during the development of the benchmark. 
F.A. supervised the research, provided guidance on experimental design and interpretation, and contributed to manuscript review and refinement. 
All authors reviewed and approved the final version of the manuscript.

\section*{Data availability}

\noindent
The DrivAerNet++ dataset analysed in this study is publicly available from the Harvard Dataverse at
\url{https://dataverse.harvard.edu/dataverse/DrivAerNet} and is described in ref.~\cite{elrefaie2024drivaernet++}.
It comprises 8{,}150 parametrically generated car geometries together with the corresponding steady-state RANS
solutions; this work uses the surface-pressure fields and the official train/validation/test split of
5{,}819, 1{,}148 and 1{,}154 designs distributed with the dataset at
\url{https://github.com/Mohamedelrefaie/DrivAerNet/tree/main/train_val_test_splits}.
The data are stored permanently in the Harvard Dataverse repository and are released under the Creative Commons
Attribution--NonCommercial 4.0 International License (CC BY--NC 4.0); no restrictions apply beyond the terms of
that license. The DrivAerML and SHIFT-SUV datasets used for the design-space comparison are available from their
original providers under the CC BY--SA 4.0 and CC BY--NC 4.0 licenses, respectively. Requests for further
information should be addressed to the corresponding author, M.E.

\section*{Code availability}

\noindent
The complete CarBench framework, comprising the preprocessing scripts, the training and evaluation pipelines,
the model configurations, the bootstrap uncertainty-estimation routines and the trained weights of every
evaluated model, will be made publicly available upon publication at
\url{https://github.com/Mohamedelrefaie/CarBench}. The code is written in Python and built on PyTorch and
PyVista; the per-model hyperparameters, random seeds and software versions needed to reproduce each reported
result are given in Table~\ref{tab:ml-config} and in the repository documentation. The public CarBench
leaderboard is hosted at \url{https://web.mit.edu/decode/www/carbench/}. The pretrained AB-UPT checkpoint
denoted AB-UPT\textsuperscript{*} was provided by the authors of that method and is identified as such
throughout the manuscript.

\section*{Additional information}

\noindent
\textbf{Competing interests.} The authors declare no competing financial or non-financial interests.

\noindent
\textbf{Correspondence.} Correspondence should be addressed to Mohamed Elrefaie: \url{mohamed.elrefaie@mit.edu}

\bibliography{sample}

@article{elrefaie2025drivaernet,
  title={Drivaernet: A parametric car dataset for data-driven aerodynamic design and prediction},
  author={Elrefaie, Mohamed and Dai, Angela and Ahmed, Faez},
  journal={Journal of Mechanical Design},
  volume={147},
  number={4},
  pages={041712},
  year={2025},
  publisher={American Society of Mechanical Engineers}
}

@inproceedings{liu2025aerogto,
  title={Aerogto: An efficient graph-transformer operator for learning large-scale aerodynamics of 3d vehicle geometries},
  author={Liu, Pengwei and Wang, Pengkai and Ren, Xingyu and Yuan, Hangjie and Hao, Zhongkai and Xu, Chao and Cai, Shengze and Ni, Dong},
  booktitle={Proceedings of the AAAI Conference on Artificial Intelligence},
  volume={39},
  number={18},
  pages={18924--18932},
  year={2025}
}

@article{alkin2025UPT,
  title={AB-UPT: Scaling Neural CFD Surrogates for High-Fidelity Automotive Aerodynamics Simulations via Anchored-Branched Universal Physics Transformers},
  author={Alkin, Benedikt and Bleeker, Maurits and Kurle, Richard and Kronlachner, Tobias and Sonnleitner, Reinhard and Dorfer, Matthias and Brandstetter, Johannes},
  journal={arXiv preprint arXiv:2502.09692},
  year={2025}
}

@article{luo2025transolverPlus,
  title={Transolver++: An Accurate Neural Solver for PDEs on Million-Scale Geometries},
  author={Luo, Huakun and Wu, Haixu and Zhou, Hang and Xing, Lanxiang and Di, Yichen and Wang, Jianmin and Long, Mingsheng},
  journal={arXiv preprint arXiv:2502.02414},
  year={2025}
}

@article{choy2025figconvnet,
  title={Factorized implicit global convolution for automotive computational fluid dynamics prediction},
  author={Choy, Chris and Kamenev, Alexey and Kossaifi, Jean and Rietmann, Max and Kautz, Jan and Azizzadenesheli, Kamyar},
  journal={arXiv preprint arXiv:2502.04317},
  year={2025}
}

@article{wu2024transolver,
  title={Transolver: A fast transformer solver for pdes on general geometries},
  author={Wu, Haixu and Luo, Huakun and Wang, Haowen and Wang, Jianmin and Long, Mingsheng},
  journal={arXiv preprint arXiv:2402.02366},
  year={2024}
}

@article{chen2025tripnet,
  title={TripNet: Learning Large-scale High-fidelity 3D Car Aerodynamics with Triplane Networks},
  author={Chen, Qian and Elrefaie, Mohamed and Dai, Angela and Ahmed, Faez},
  journal={arXiv preprint arXiv:2503.17400},
  year={2025}
}

@article{ashton2024drivaerml,
  title={DrivAerML: High-fidelity computational fluid dynamics dataset for road-car external aerodynamics},
  author={Ashton, Neil and Mockett, Charles and Fuchs, Marian and Fliessbach, Louis and Hetmann, Hendrik and Knacke, Thilo and Schonwald, Norbert and Skaperdas, Vangelis and Fotiadis, Grigoris and Walle, Astrid and others},
  journal={arXiv preprint arXiv:2408.11969},
  year={2024}
}

@article{ashton2024ahmedml,
  title={AhmedML: High-fidelity computational fluid dynamics dataset for incompressible, low-speed bluff body aerodynamics},
  author={Ashton, Neil and Maddix, Danielle C and Gundry, Samuel and Shabestari, Parisa M},
  journal={arXiv preprint arXiv:2407.20801},
  year={2024}
}

@article{jacob2025benchmarking,
  title={Benchmarking Convolutional Neural Network and Graph Neural Network based Surrogate Models on a Real-World Car External Aerodynamics Dataset},
  author={Jacob, Sam Jacob and Mrosek, Markus and Othmer, Carsten and K{\"o}stler, Harald},
  journal={arXiv preprint arXiv:2504.06699},
  year={2025}
}

@article{pang2023masked,
  title={Masked autoencoders for 3d point cloud self-supervised learning},
  author={Pang, Yatian and Tay, Eng Hock Francis and Yuan, Li and Chen, Zhenghua},
  journal={World Scientific Annual Review of Artificial Intelligence},
  volume={1},
  pages={2440001},
  year={2023},
  publisher={World Scientific}
}

@article{li2020fourier,
  title={Fourier neural operator for parametric partial differential equations},
  author={Li, Zongyi and Kovachki, Nikola and Azizzadenesheli, Kamyar and Liu, Burigede and Bhattacharya, Kaushik and Stuart, Andrew and Anandkumar, Anima},
  journal={arXiv preprint arXiv:2010.08895},
  year={2020}
}

@article{li2023geometry,
  title={Geometry-informed neural operator for large-scale 3d pdes},
  author={Li, Zongyi and Kovachki, Nikola and Choy, Chris and Li, Boyi and Kossaifi, Jean and Otta, Shourya and Nabian, Mohammad Amin and Stadler, Maximilian and Hundt, Christian and Azizzadenesheli, Kamyar and others},
  journal={Advances in Neural Information Processing Systems},
  volume={36},
  pages={35836--35854},
  year={2023}
}

@inproceedings{elrefaie2024drivaernet,
  title={Drivaernet: A parametric car dataset for data-driven aerodynamic design and graph-based drag prediction},
  author={Elrefaie, Mohamed and Dai, Angela and Ahmed, Faez},
  booktitle={International Design Engineering Technical Conferences and Computers and Information in Engineering Conference},
  volume={88360},
  pages={V03AT03A019},
  year={2024},
  organization={American Society of Mechanical Engineers}
}

@article{elrefaie2024drivaernet++,
  title={Drivaernet++: A large-scale multimodal car dataset with computational fluid dynamics simulations and deep learning benchmarks},
  author={Elrefaie, Mohamed and Morar, Florin and Dai, Angela and Ahmed, Faez},
  journal={Advances in Neural Information Processing Systems},
  volume={37},
  pages={499--536},
  year={2024}
}

@article{elrefaie2025ai,
  title={AI agents in engineering design: a multi-agent framework for aesthetic and aerodynamic car design},
  author={Elrefaie, Mohamed and Qian, Janet and Wu, Raina and Chen, Qian and Dai, Angela and Ahmed, Faez},
  journal={arXiv preprint arXiv:2503.23315},
  year={2025}
}

@article{takamoto2022pdebench,
  title={Pdebench: An extensive benchmark for scientific machine learning},
  author={Takamoto, Makoto and Praditia, Timothy and Leiteritz, Raphael and MacKinlay, Daniel and Alesiani, Francesco and Pfl{\"u}ger, Dirk and Niepert, Mathias},
  journal={Advances in Neural Information Processing Systems},
  volume={35},
  pages={1596--1611},
  year={2022}
}

@article{vatani2025tripoptimizer,
  title={TripOptimizer: Generative 3D Shape Optimization and Drag Prediction using Triplane VAE Networks},
  author={Vatani, Parsa and Elrefaie, Mohamed and Nazarpour, Farhad and Ahmed, Faez},
  journal={arXiv preprint arXiv:2509.12224},
  year={2025}
}

@article{wang2018glue,
  title={GLUE: A multi-task benchmark and analysis platform for natural language understanding},
  author={Wang, Alex and Singh, Amanpreet and Michael, Julian and Hill, Felix and Levy, Omer and Bowman, Samuel R},
  journal={arXiv preprint arXiv:1804.07461},
  year={2018}
}

@inproceedings{deng2009imagenet,
  title={Imagenet: A large-scale hierarchical image database},
  author={Deng, Jia and Dong, Wei and Socher, Richard and Li, Li-Jia and Li, Kai and Fei-Fei, Li},
  booktitle={2009 IEEE conference on computer vision and pattern recognition},
  pages={248--255},
  year={2009},
  organization={Ieee}
}

@article{wang2019superglue,
  title={Superglue: A stickier benchmark for general-purpose language understanding systems},
  author={Wang, Alex and Pruksachatkun, Yada and Nangia, Nikita and Singh, Amanpreet and Michael, Julian and Hill, Felix and Levy, Omer and Bowman, Samuel},
  journal={Advances in neural information processing systems},
  volume={32},
  year={2019}
}

@article{hendrycks2020measuring,
  title={Measuring massive multitask language understanding},
  author={Hendrycks, Dan and Burns, Collin and Basart, Steven and Zou, Andy and Mazeika, Mantas and Song, Dawn and Steinhardt, Jacob},
  journal={arXiv preprint arXiv:2009.03300},
  year={2020}
}

@article{jacob2021deep,
  title={Deep learning for real-time aerodynamic evaluations of arbitrary vehicle shapes},
  author={Jacob, Sam Jacob and Mrosek, Markus and Othmer, Carsten and K{\"o}stler, Harald},
  journal={arXiv preprint arXiv:2108.05798},
  year={2021}
}

@article{wen2025geometry,
  title={Geometry aware operator transformer as an efficient and accurate neural surrogate for pdes on arbitrary domains},
  author={Wen, Shizheng and Kumbhat, Arsh and Lingsch, Levi and Mousavi, Sepehr and Zhao, Yizhou and Chandrashekar, Praveen and Mishra, Siddhartha},
  journal={arXiv preprint arXiv:2505.18781},
  year={2025}
}

@inproceedings{zhao2021point,
  title={Point transformer},
  author={Zhao, Hengshuang and Jiang, Li and Jia, Jiaya and Torr, Philip HS and Koltun, Vladlen},
  booktitle={Proceedings of the IEEE/CVF international conference on computer vision},
  pages={16259--16268},
  year={2021}
}

@inproceedings{qi2017pointnet,
  title={Pointnet: Deep learning on point sets for 3d classification and segmentation},
  author={Qi, Charles R and Su, Hao and Mo, Kaichun and Guibas, Leonidas J},
  booktitle={Proceedings of the IEEE conference on computer vision and pattern recognition},
  pages={652--660},
  year={2017}
}

@article{zeng2025lrq,
  title={LRQ-Solver: A Transformer-Based Neural Operator for Fast and Accurate Solving of Large-scale 3D PDEs},
  author={Zeng, Peijian and Wang, Guan and Gu, Haohao and Hu, Xiaoguang and Wang, Zhuowei and Yang, Aimin and Song, Xiaoyu and others},
  journal={arXiv preprint arXiv:2510.11636},
  year={2025}
}

@article{sung2025blendednet,
  title={BlendedNet: A Blended Wing Body Aircraft Dataset and Surrogate Model for Aerodynamic Predictions},
  author={Sung, Nicholas and Spreizer, Steven and Elrefaie, Mohamed and Samuel, Kaira and Jones, Matthew C and Ahmed, Faez},
  journal={arXiv preprint arXiv:2509.07209},
  year={2025}
}

@inproceedings{elrefaie2024surrogate,
  title={Surrogate modeling of the aerodynamic performance for airfoils in transonic regime},
  author={Elrefaie, Mohamed and Ayman, Tarek and Elrefaie, Mayar and Sayed, Eman and Ayyad, Mahmoud and AbdelRahman, Mohamed M},
  booktitle={AIAA SCITECH 2024 Forum},
  pages={2220},
  year={2024}
}

@misc{luminarycloud_shift_suv_2025,
  author       = {Luminary-Cloud},
  title        = {SHIFT-SUV: High-Fidelity Computational Fluid Dynamics Dataset for SUV External Aerodynamics},
  year         = {2025},
  howpublished = {\url{https://huggingface.co/datasets/luminary-shift/SUV}},
  note         = {Accessed: 2025-10-19}
}

@misc{ecara_drivaer_reference,
  author       = {{European Car Aerodynamic Research Association (ECARA)}},
  title        = {AeroSUV / DrivAer Reference Models --- Data Exchange},
  year         = {2019},
  howpublished = {\url{https://www.ecara.org/drivaer-1}},
  note         = {Accessed: 2025-10-20}
}

@misc{tum_drivaer_geometry,
  author       = {{Chair of Aerodynamics and Fluid Mechanics, Technical University of Munich}},
  title        = {DrivAer Model Geometry},
  howpublished = {\url{https://www.epc.ed.tum.de/en/aer/research-groups/automotive/drivaer/geometry/}},
  note         = {Accessed: 2025-10-20}
}

@article{wang2019dynamic,
  title={Dynamic graph cnn for learning on point clouds},
  author={Wang, Yue and Sun, Yongbin and Liu, Ziwei and Sarma, Sanjay E and Bronstein, Michael M and Solomon, Justin M},
  journal={ACM Transactions on Graphics (tog)},
  volume={38},
  number={5},
  pages={1--12},
  year={2019},
  publisher={Acm New York, NY, USA}
}

@article{sullivan2019pyvista,
  doi = {10.21105/joss.01450},
  url = {https://doi.org/10.21105/joss.01450},
  year = {2019},
  month = {May},
  publisher = {The Open Journal},
  volume = {4},
  number = {37},
  pages = {1450},
  author = {Bane Sullivan and Alexander Kaszynski},
  title = {{PyVista}: {3D} plotting and mesh analysis through a streamlined interface for the {Visualization Toolkit} ({VTK})},
  journal = {Journal of Open Source Software}
}

@article{sudin2014review,
  title={Review of research on vehicles aerodynamic drag reduction methods},
  author={Sudin, Mohd Nizam and Abdullah, Mohd Azman and Shamsuddin, Shamsul Anuar and Ramli, Faiz Redza and Tahir, Musthafah Mohd},
  journal={International Journal of Mechanical and Mechatronics Engineering},
  volume={14},
  number={2},
  pages={35--47},
  year={2014}
}

@book{schuetz2015aerodynamics,
  title={Aerodynamics of road vehicles},
  author={Schuetz, Thomas Christian},
  year={2015},
  publisher={Sae International}
}

@article{lighthill1952sound,
  title={On sound generated aerodynamically I. General theory},
  author={Lighthill, Michael James},
  journal={Proceedings of the Royal Society of London. Series A. Mathematical and Physical Sciences},
  volume={211},
  number={1107},
  pages={564--587},
  year={1952},
  publisher={The Royal Society London}
}

@article{oettle2017automotive,
  title={Automotive aeroacoustics: An overview},
  author={Oettle, Nicholas and Sims-Williams, David},
  journal={Proceedings of the Institution of Mechanical Engineers, Part D: Journal of Automobile Engineering},
  volume={231},
  number={9},
  pages={1177--1189},
  year={2017},
  publisher={SAGE Publications Sage UK: London, England}
}

@techreport{heft2012introduction,
  title={Introduction of a new realistic generic car model for aerodynamic investigations},
  author={Heft, Angelina I and Indinger, Thomas and Adams, Nikolaus A},
  year={2012},
  institution={SAE Technical Paper}
}

@misc{greenshields2023openfoam,
  title={OpenFOAM v11 User Guide. The OpenFOAM Foundation, London},
  author={Greenshields, C},
  year={2023},
  publisher={UK},
url={https://doc.cfd.direct/openfoam/user-guide-v11/bookindex}
}

@article{brandt2019effects,
  title={The effects of wheel design on the aerodynamic drag of passenger vehicles},
  author={Brandt, Adam and Berg, Henrik and Bolzon, Michael and Josefsson, Linda},
  journal={SAE International Journal of Advances and Current Practices in Mobility},
  volume={1},
  number={2019-01-0662},
  pages={1279--1299},
  year={2019}
}

@inproceedings{pfadenhauer1996influence,
  title={On the influence of wheels and tyres on the aerodynamic drag of vehicles},
  author={Pfadenhauer, M and Wickern, G and Zwicker, K},
  booktitle={MIRA international conference on vehicle aerodynamics},
  year={1996}
}

@article{menter2003ten,
  title={Ten years of industrial experience with the SST turbulence model},
  author={Menter, Florian R and Kuntz, Martin and Langtry, Robin and others},
  journal={Turbulence, heat and mass transfer},
  volume={4},
  number={1},
  pages={625--632},
  year={2003}
}

@article{yagoubi2024neurips,
  title={Neurips 2024 ml4cfd competition: Harnessing machine learning for computational fluid dynamics in airfoil design},
  author={Yagoubi, Mouadh and Danan, David and Leyli-Abadi, Milad and Brunet, Jean-Patrick and Mazari, Jocelyn Ahmed and Bonnet, Florent and Farjallah, Asma and Cinnella, Paola and Gallinari, Patrick and Schoenauer, Marc and others},
  journal={arXiv preprint arXiv:2407.01641},
  year={2024}
}

@book{efron1994introduction,
  title={An introduction to the bootstrap},
  author={Efron, Bradley and Tibshirani, Robert J},
  year={1994},
  publisher={Chapman and Hall/CRC}
}

@incollection{efron1992bootstrap,
  title={Bootstrap methods: another look at the jackknife},
  author={Efron, Bradley},
  booktitle={Breakthroughs in statistics: Methodology and distribution},
  pages={569--593},
  year={1992},
  publisher={Springer}
}

@article{rabeh2025benchmarking,
  title={Benchmarking scientific machine-learning approaches for flow prediction around complex geometries},
  author={Rabeh, Ali and Herron, Ethan and Balu, Aditya and Sarkar, Soumik and Hegde, Chinmay and Krishnamurthy, Adarsh and Ganapathysubramanian, Baskar},
  journal={Communications Engineering},
  volume={4},
  number={1},
  pages={182},
  year={2025},
  publisher={Nature Publishing Group UK London}
}

@inproceedings{bekemeyer2025introduction,
  title={Introduction of Applied Aerodynamics Surrogate Modeling Benchmark Cases},
  author={Bekemeyer, Philipp and Hariharan, Nathan and Wissink, Andrew M and Cornelius, Jason},
  booktitle={AIAA SCITECH 2025 Forum},
  pages={0036},
  year={2025}
}

@article{krizhevsky2012imagenet,
  title={Imagenet classification with deep convolutional neural networks},
  author={Krizhevsky, Alex and Sutskever, Ilya and Hinton, Geoffrey E},
  journal={Advances in neural information processing systems},
  volume={25},
  year={2012}
}

@inproceedings{devlin2019bert,
  title={Bert: Pre-training of deep bidirectional transformers for language understanding},
  author={Devlin, Jacob and Chang, Ming-Wei and Lee, Kenton and Toutanova, Kristina},
  booktitle={Proceedings of the 2019 conference of the North American chapter of the association for computational linguistics: human language technologies, volume 1 (long and short papers)},
  pages={4171--4186},
  year={2019}
}

@article{brown2020language,
  title={Language models are few-shot learners},
  author={Brown, Tom and Mann, Benjamin and Ryder, Nick and Subbiah, Melanie and Kaplan, Jared D and Dhariwal, Prafulla and Neelakantan, Arvind and Shyam, Pranav and Sastry, Girish and Askell, Amanda and others},
  journal={Advances in neural information processing systems},
  volume={33},
  pages={1877--1901},
  year={2020}
}

@article{tangsali2025benchmarking,
  title={A Benchmarking Framework for AI models in Automotive Aerodynamics},
  author={Tangsali, Kaustubh and Ranade, Rishikesh and Nabian, Mohammad Amin and Kamenev, Alexey and Sharpe, Peter and Ashton, Neil and Cherukuri, Ram and Choudhry, Sanjay},
  journal={arXiv preprint arXiv:2507.10747},
  year={2025}
}

@book{hucho2013aerodynamik,
  title={Aerodynamik des Automobils: eine Br{\"u}cke von der Str{\"o}mungsmechanik zur Fahrzeugtechnik},
  author={Hucho, Wolf-Heinrich},
  year={2013},
  publisher={Springer-Verlag}
}

@article{umetani2018learning,
  title={Learning three-dimensional flow for interactive aerodynamic design},
  author={Umetani, Nobuyuki and Bickel, Bernd},
  journal={ACM Transactions on Graphics (TOG)},
  volume={37},
  number={4},
  pages={1--10},
  year={2018},
  publisher={ACM New York, NY, USA}
}

@article{li2023fourier,
  title={Fourier neural operator with learned deformations for {PDE}s on general geometries},
  author={Li, Zongyi and Huang, Daniel Zhengyu and Liu, Burigede and Anandkumar, Anima},
  journal={Journal of Machine Learning Research},
  volume={24},
  number={388},
  pages={1--26},
  year={2023}
}

@article{casenave2025physics,
  title={Physics-Learning AI Datamodel (PLAID) datasets: a collection of physics simulations for machine learning},
  author={Casenave, Fabien and Roynard, Xavier and Staber, Brian and Piat, William and Bucci, Michele Alessandro and Akkari, Nissrine and Kabalan, Abbas and Nguyen, Xuan Minh Vuong and Saverio, Luca and Perez, Rapha{\"e}l Carpintero and others},
  journal={arXiv preprint arXiv:2505.02974},
  year={2025}
}

\clearpage
\setcounter{table}{0}
\setcounter{figure}{0}
\setcounter{algorithm}{0}
\captionsetup[table]{name=Supplementary Table}
\captionsetup[figure]{name=Supplementary Figure}
\floatname{algorithm}{Supplementary Algorithm}
\renewcommand{\thetable}{\arabic{table}}
\renewcommand{\thefigure}{\arabic{figure}}

\section*{Supplementary Information}

\noindent
This section contains the supplementary notes, tables and figure supporting the
main text. References to Tables and Figures given without the word
``Supplementary'' refer to the main text above.

\section*{Existing Benchmarks Comparison}

Recent initiatives have introduced standardized benchmarks for learning-based
scientific simulation, providing unified datasets, evaluation metrics, and
experimental pipelines across various physics domains. 
PLAID~\cite{casenave2025physics} introduces a unified and extensible data layer for
heterogeneous physics simulations, together with six openly released
structural-mechanics and computational-fluid-dynamics datasets, reproducible
evaluation protocols, and Hugging Face integration for community-driven
benchmarking across variable geometries, meshes, and topologies.
PDEBench~\cite{takamoto2022pdebench} offers a broad suite of time-dependent
PDE problems, including advection, Burgers', compressible Navier--Stokes, and
Darcy flow, together with baseline neural solvers to support reproducible
scientific machine learning.
The NeurIPS 2024 ML4CFD Competition~\cite{yagoubi2024neurips} advanced
benchmarking in external aerodynamics by evaluating surrogate models on the
AirfRANS dataset via a multi-criteria framework that emphasizes not only
predictive accuracy but also inference efficiency and out-of-distribution
robustness.
Complementing these efforts, Rabeh~et~al.~\cite{rabeh2025benchmarking}
systematically compared scientific machine-learning architectures for
steady-state flow prediction around complex shapes, assessing global error,
boundary-layer fidelity, and physical consistency through a unified scoring
protocol.
Bekemeyer et al.~\cite{bekemeyer2025introduction} expanded benchmarking for
applied aerodynamics by introducing four well-documented test cases for
surrogate modeling of airfoils, missiles, and three-dimensional aerodynamic
surfaces. Their work highlights the need for standardized formats, rigorous
documentation, and reproducible pipelines in surrogate-based aerodynamic
modeling.

\begin{table*}[h]
\centering
\scriptsize
\caption{
Comparison of existing CFD-based automotive aerodynamics datasets and benchmarks. 
Each entry lists the corresponding study and dataset, characterized by simulation scale, test set size, number of design categories, number of evaluated machine learning models, flow regime, and benchmark availability. 
DrivAerNet++ is the largest and most diverse open dataset, supporting full-field aerodynamic supervision, standardized benchmarking, and cross-category generalization across distinct car archetypes. 
RANS stands for Reynolds-Averaged Navier--Stokes Simulation, DES stands for Detached Eddy Simulation, and LES for Large Eddy Simulation. 
Bold values indicate the best entries, while underlined values indicate the second best. *Note that the VW Real-World benchmark evaluates only drag coefficients and does not provide full-field flow supervision.}
\label{tab:dataset_comparison}
\setlength{\tabcolsep}{2.5pt}
\begin{tabular}{llcccccccc}
\toprule
\textbf{Study} & 
\textbf{Dataset} & 
\shortstack{\textbf{Simulations}} & 
\textbf{Test Size} & 
\shortstack{\textbf{Design}\\\textbf{Categories}} & 
\shortstack{\textbf{\# Models}\\\textbf{Evaluated}} & 
\textbf{Flow Regime} & 
\shortstack{\textbf{Benchmark}\\\textbf{Availability}} & 
\shortstack{\textbf{Cross-Category}\\\textbf{Generalization}} &
\shortstack{\textbf{Uncertainty}\\\textbf{Quantification}} \\
\midrule
CarBench (Ours) & DrivAerNet++ & \textbf{8{,}150} & \textbf{1{,}154} & \underline{8} & \underline{11} & Steady RANS (OpenFOAM) & \checkmark & \checkmark & \checkmark \\
Jacob et al. ~\cite{jacob2025benchmarking} & VW Real-World* & 343 & 69 & \textbf{32} & 2 & DES (OpenFOAM) & \xmark & \xmark & \xmark \\
Li et al.~\cite{li2023geometry} & ShapeNet-Car & 889 & 100 & 1 & 5 & Steady RANS (OpenFOAM) & \xmark & \xmark & \xmark \\
Tangsali et al. ~\cite{tangsali2025benchmarking} & DrivAerML & 484 & 48 & 1  & 3 & Hybrid RANS-LES (OpenFOAM) & \checkmark & \xmark & \xmark \\
Li et al.~\cite{li2023geometry}  & Ahmed body & 551 & 51 & 1 & 5 & Steady RANS (OpenFOAM) & \xmark & \xmark & \xmark \\
Luo et al.~\cite{luo2025transolverPlus} & DrivAerNet++ & 200 & 10 & 3 & \textbf{13} & Steady RANS (OpenFOAM) & \xmark & \xmark & \xmark \\
Jacob et al.~\cite{jacob2021deep} & VW Internal & \underline{1000} & \underline{150} & 2 & 3 & Lattice Boltzmann (ultraFluidX) & \xmark & \xmark & \xmark \\
\bottomrule
\end{tabular}
\end{table*}

While these initiatives represent important progress, they primarily focus on lower-dimensional PDE settings or simplified aerodynamic configurations. Only a limited number of datasets address high-resolution three-dimensional shapes, and fewer still offer sufficient diversity to support meaningful generalization studies. 
To contextualize this landscape, Supplementary Table~\ref{tab:dataset_comparison} compares existing 3D CFD-based datasets for automotive aerodynamics in terms of scale, category diversity, availability of full-field flow data, and benchmark completeness.

Despite their contributions, current benchmarks leave several open gaps. 
Transolver++ by Luo et al.~\cite{luo2025transolverPlus} provides one of the earliest attempts at benchmarking transformer-based PDE solvers across multiple CFD tasks, but its evaluation on DrivAerNet++ relies on only 200 simulations and ten test samples across three car categories. 
Although useful for proof-of-concept validation, such limited sampling cannot capture the geometric and aerodynamic diversity needed to assess model robustness or cross-category transfer. Similarly, the Volkswagen Real-World dataset~\cite{jacob2025benchmarking} represents a valuable step toward connecting academic and industrial CFD workflows, containing 343 simulations across 32 production car designs. However, its exclusive focus on drag coefficient prediction, coupled with the absence of full-field aerodynamic supervision, constrains its suitability for training general-purpose surrogate models. Moreover, the VW Real-World~\cite{jacob2025benchmarking} and VW Internal~\cite{jacob2021deep} datasets represent valuable contributions from industry that help bridge academic and real-world CFD workflows. However, both datasets are proprietary, which limits reproducibility and broader accessibility for the open research community. Other open datasets, such as ShapeNet-Car~\cite{umetani2018learning} and the Ahmed body geometries~\cite{li2023geometry} provide 889 and 551 RANS simulations, respectively, but contain only a single vehicle class, making them insufficient for cross-category generalization or broad aerodynamic design tasks.  
More recently, the PhysicsNeMo benchmark~\cite{tangsali2025benchmarking} introduced an evaluation framework based on the open-source PhysicsNeMo-CFD toolkit. However, its scale remains modest (484 simulations with 48 test cases), and its benchmark lacks uncertainty quantification, category diversity, and the breadth of aerodynamic supervision required for benchmarking modern deep learning models.

In contrast, \textbf{CarBench} introduces the first open, large-scale, and geometrically diverse benchmark dedicated specifically to high-resolution three-dimensional external aerodynamics. Built on 8{,}150 steady-state RANS simulations across eight parametric base models from DrivAerNet++, grouped into three body-style categories (Fastback, Estateback and Notchback),, CarBench supports full-field aerodynamic supervision, standardized train--validation--test splits (including 1{,}154 held-out test samples), model-agnostic evaluation metrics, and rigorous uncertainty quantification. 
Its consistent numerical settings and explicit support for cross-category generalization together provide the scale and diversity required to evaluate modern geometric deep learning, neural operator, and transformer-based solvers.  By combining statistical representativeness and reproducible pipelines, CarBench addresses a gap in physics-based ML benchmarking and enables fair, large-scale comparison of learning-based aerodynamic models.

\newpage

\section*{Evaluation Metrics}

To rigorously evaluate the performance of the models, we employ a set of well-established metrics that quantify the accuracy, error, and generalization capabilities of the predictions. These metrics are defined as follows:

\begin{itemize}
    \item \textbf{Mean Absolute Error (MAE):}  
    The MAE measures the average magnitude of absolute errors between the predicted values ($\hat{y}_i$) and the ground truth values ($y_i$). It is defined as:
    \[
    \text{MAE} = \frac{1}{N} \sum_{i=1}^{N} \left| \hat{y}_i - y_i \right|
    \]
    where $N$ is the total number of data points. This metric provides an intuitive measure of the average error magnitude, making it easy to interpret. Lower MAE values indicate better model performance.

    \item \textbf{Mean Squared Error (MSE):}  
    The MSE quantifies the average of the squared differences between predicted and true values. It is expressed as:
    \[
    \text{MSE} = \frac{1}{N} \sum_{i=1}^{N} \left( \hat{y}_i - y_i \right)^2
    \]
    This metric penalizes larger deviations more severely, emphasizing outliers and high-error regions. It serves as the most common loss function during training for regression-based models and provides a direct measure of model accuracy in squared physical units (e.g., m$^4$/s$^4$ for kinematic pressure).

    \item \textbf{Root Mean Square Error (RMSE):}  
    The RMSE computes the square root of the mean squared differences between predictions and ground truth. It is given by:
    \[
    \text{RMSE} = \sqrt{\frac{1}{N} \sum_{i=1}^{N} \left( \hat{y}_i - y_i \right)^2}
    \]
    RMSE penalizes larger errors more heavily than MAE due to the squaring operation, making it particularly sensitive to outliers. It is commonly used when large deviations are of greater concern.

    \item \textbf{Coefficient of Determination ($R^2$):}  
    The $R^2$ metric, also known as the coefficient of determination, quantifies the proportion of variance in the ground truth that is explained by the model's predictions. It is defined as:
    \[
    R^2 = 1 - \frac{\sum_{i=1}^{N} \left( y_i - \hat{y}_i \right)^2}{\sum_{i=1}^{N} \left( y_i - \bar{y} \right)^2}
    \]
    where $\bar{y}$ is the mean of the ground truth values. An $R^2$ value close to 1 indicates that the model explains most of the variance, while a value close to 0 suggests poor predictive performance.

    \item \textbf{Relative L2 Error:}  
    The Relative L2 error evaluates the normalized L2 error between predictions and ground truth. It is defined as:
    \[
    \text{Relative L2} = \frac{\| \hat{y} - y \|_2}{\| y \|_2}
    \]
    where $\| \cdot \|_2$ denotes the Euclidean norm. This metric provides a scale-invariant measure of error, making it particularly useful for comparing models across datasets with different scales.

    \item \textbf{Relative L1 Error:}  
    The Relative L1 error computes the normalized L1 error, which is the sum of absolute differences between predictions and ground truth, normalized by the sum of absolute ground truth values:
    \[
    \text{Relative L1} = \frac{\sum_{i=1}^{N} \left| \hat{y}_i - y_i \right|}{\sum_{i=1}^{N} \left| y_i \right|}
    \]
    This metric is also scale-invariant and provides an alternative to Relative L2 for datasets where absolute differences are more meaningful.

    \item \textbf{Maximum Error:}  
    The Maximum Error captures the largest absolute error in the predictions:
    \[
    \text{Max Error} = \max_{i} \left| \hat{y}_i - y_i \right|
    \]
    This metric highlights the worst-case performance of the model, which is critical in applications where large errors can have significant consequences.

    \item \textbf{Median Relative Error:}
    To characterize the \emph{typical} pointwise accuracy, we define the
    elementwise relative error at each surface point $i$ as
    \[
    e_i = \frac{\left| \hat{y}_i - y_i \right|}{\left| y_i \right|},
    \]
    and report its median over all $N$ surface points pooled across every
    geometry in the test set:
    \[
    \text{Median Relative Error} \;=\;
    \operatorname*{median}_{1 \le i \le N} e_i
    \;=\;
    \operatorname*{median}_{1 \le i \le N}
    \frac{\left| \hat{y}_i - y_i \right|}{\left| y_i \right|}.
    \]
    This quantity is dimensionless. We use the median rather than the mean
    because the pointwise relative error is heavy-tailed: the surface pressure
    $y_i$ passes through zero in stagnation and transition regions, so $e_i$ can
    become arbitrarily large at isolated points. The median is robust to these
    singularities and therefore reflects the typical relative accuracy of a
    model. Lower values indicate better performance.
    \item \textbf{Percentile Errors (P50, P90, P95, P99):}  
    Let $\varepsilon_i = \left| \hat{y}_i - y_i \right|$ denote the absolute
    error at surface point $i$, pooled across all $N$ surface points of all test
    geometries, and let $F(\varepsilon) = \frac{1}{N}\sum_{i=1}^{N}
    \mathbf{1}\!\left[\varepsilon_i \le \varepsilon\right]$ be its empirical
    cumulative distribution. The $k$-th percentile error is
    \[
    P_k \;=\; \inf\left\{ \varepsilon \ge 0 \;:\; F(\varepsilon) \ge k/100 \right\},
    \]
    computed as the empirical $k$-th percentile of $\{\varepsilon_i\}_{i=1}^{N}$
    (linear interpolation between order statistics). By construction $P_{50}$ is
    the median absolute error, while $P_{90}$, $P_{95}$, and $P_{99}$ quantify the
    upper tail of the error distribution, i.e.\ the worst-case deviations at the
    $10\%$, $5\%$, and $1\%$ most-erroneous surface points, respectively. These
    are reported in physical pressure units (m$^2$/s$^2$). Because they summarize
    the full cumulative distribution of pointwise errors rather than a single
    average, the percentile errors expose spatially localized failure modes that
    aggregate metrics such as MAE or RMSE can mask. Lower values indicate better
    performance.
\end{itemize}

Together, these metrics capture absolute errors, relative errors, and explained variance, characterizing predictive performance from complementary perspectives.

\subsection*{Pressure Units}

\noindent
In this work, all simulations were performed using the \textsc{simpleFoam} solver in \textsc{OpenFOAM}~\cite{greenshields2023openfoam}, which operates with \textit{kinematic pressure} (\(p/\rho\)) rather than absolute pressure. 
Kinematic pressure has SI dimensions of \(\mathrm{m^2\,s^{-2}}\), as opposed to absolute pressure with units of \(\mathrm{Pa = kg\,m^{-1}\,s^{-2}}\). 
This formulation simplifies the pressure--velocity coupling by removing the explicit dependence on density, making it particularly suitable for incompressible steady-state flows. 
Consequently, all reported pressure values and visualizations in this paper are expressed in terms of the kinematic pressure. 
For physical interpretation or validation, these quantities can be converted to Pascals by multiplying by the reference fluid density, ensuring consistent comparison across different flow conditions. All machine learning models were trained using normalized pressure values to ensure numerical stability and balanced gradient magnitudes during optimization. 
During evaluation, however, results are reported in kinematic pressure units (\(\mathrm{m^2/s^2}\)) following the conventions of \textsc{OpenFOAM}. 
This approach combines the advantages of normalized learning, including stable convergence and scale consistency, with physically interpretable outputs at inference time. Reporting evaluation metrics such as MAE, MSE, and RMSE in physical units provides a meaningful link between model performance and aerodynamic quantities, enabling direct comparison with CFD reference data and quantifying predictive deviations in real-world terms.

\newpage

\section*{Bootstrap Confidence Intervals for Performance Metrics}

To quantify the statistical reliability and uncertainty of model performance, we employed the bootstrap resampling method~\cite{efron1992bootstrap}. This nonparametric approach estimates the sampling distribution of a statistic by repeatedly resampling with replacement from the observed test data. Unlike parametric methods that assume Gaussian or other specific error distributions, bootstrap resampling does not rely on such assumptions, making it particularly suitable for evaluating complex performance metrics such as $R^2$ or relative L2 error.

\subsection*{Bootstrap Methodology}

Given a test set $\mathcal{D}_{\text{test}} = \{(\mathbf{x}_i, y_i)\}_{i=1}^{N}$, where $\mathbf{x}_i$ denotes the input geometry and $y_i$ the ground-truth pressure field, we generate $B$ bootstrap replicates $\{\mathcal{D}_{\text{test}}^{(b)}\}_{b=1}^{B}$ by randomly sampling $N$ pairs with replacement. For each replicate $\mathcal{D}_{\text{test}}^{(b)}$, we compute the performance metric $\theta^{(b)} = f(\mathcal{D}_{\text{test}}^{(b)})$ (e.g., MSE, $R^2$, or MAE).

The bootstrap estimates of the mean and standard deviation are given by:
\begin{equation}
\bar{\theta}_{\text{boot}} = \frac{1}{B}\sum_{b=1}^B \theta^{(b)}, \qquad
\sigma_{\text{boot}} = \sqrt{\frac{1}{B-1}\sum_{b=1}^B(\theta^{(b)} - \bar{\theta}_{\text{boot}})^2}.
\end{equation}

Confidence intervals are obtained using the \emph{percentile method}:
\begin{equation}
\text{CI}_{1-\alpha} = [\theta_{\alpha/2}^*,\, \theta_{1-\alpha/2}^*],
\end{equation}
where $\theta_{p}^*$ denotes the $p$-th percentile of the bootstrap distribution.

\subsection*{Implementation Details}

We used the following settings in our experiments:
\begin{itemize}
    \item \textbf{Number of bootstrap iterations:} $B = 2000$, ensuring stable confidence interval estimates while remaining computationally tractable~\cite{efron1994introduction}.
    \item \textbf{Confidence level:} $95\%$ ($\alpha = 0.05$), corresponding to the 2.5th and 97.5th percentiles.
    \item \textbf{Metrics evaluated:} $R^2$, MSE, MAE, RMSE, relative L2 error, and maximum error.
\end{itemize}

The complete algorithmic procedure is summarized in Supplementary Algorithm~\ref{alg:bootstrap-ci}.

\begin{algorithm}[H]
\caption{Bootstrap Confidence Interval Estimation}
\label{alg:bootstrap-ci}
\begin{algorithmic}[1]
\Require Test set $\mathcal{D}_{\text{test}}$, predictions $\{\hat{y}_i\}_{i=1}^N$, number of iterations $B$
\Ensure Mean $\bar{\theta}$, confidence interval $[\theta_{\text{lower}}, \theta_{\text{upper}}]$
\State Initialize list $\Theta = []$
\For{$b = 1$ to $B$}
    \State Sample $N$ indices with replacement: $\mathcal{I}^{(b)} \sim \{1,\dots,N\}$
    \State Compute metric on resampled pairs: $\theta^{(b)} = f(\{(y_i,\hat{y}_i): i \in \mathcal{I}^{(b)}\})$
    \State Append $\theta^{(b)}$ to $\Theta$
\EndFor
\State Sort $\Theta$ in ascending order
\State Compute mean: $\bar{\theta} = \frac{1}{B}\sum_{b=1}^B \theta^{(b)}$
\State Extract 95\% CI: $\theta_{\text{lower}} = \Theta[0.025B]$, $\theta_{\text{upper}} = \Theta[0.975B]$
\State \Return $\bar{\theta}$, $[\theta_{\text{lower}}, \theta_{\text{upper}}]$
\end{algorithmic}
\end{algorithm}

\subsection*{Interpretation and Relevance}

Bootstrap confidence intervals provide several important advantages for model evaluation:

\begin{enumerate}
    \item \textbf{Uncertainty quantification:} They provide an empirical measure of variability in model performance, distinguishing genuine improvements from random fluctuations due to finite test samples.
    \item \textbf{Distribution-free robustness:} The method does not rely on normality or homoscedasticity assumptions, ensuring reliability for skewed or heavy-tailed metric distributions.
    \item \textbf{Statistical comparability:} Overlap or separation of marginal confidence intervals is not, by itself, a formal test of the difference between two models. Because models are evaluated on the same geometries, pairwise comparisons should use paired resampling of model differences.
    \item \textbf{Stability assessment:} Narrow intervals indicate consistent generalization, while wider intervals suggest sensitivity to test-set composition.
\end{enumerate}

All bootstrap distributions, sample statistics, and confidence bounds were stored to facilitate reproducibility and downstream statistical analyses. This uncertainty quantification provides the statistical basis for the model comparisons reported in this work.

\newpage

\section*{Baseline Models}

The following baseline models span a diverse set of architectural families widely used in geometric deep learning, including point-based, graph-based, and attention-based paradigms. Each model is adapted for dense surface-pressure regression on 3D car geometries and trained using a standardized protocol for fair comparison. Below, we describe the design choices, training setups, and architectural adaptations used for each method.

\subsection*{RegDGCNN}
The RegDGCNN model, introduced by Elrefaie et al.~\cite{elrefaie2025drivaernet}, is a modified version of Dynamic Graph Convolutional Neural Networks (DGCNN)~\cite{wang2019dynamic}. It constructs dynamic $k$-nearest neighbors graphs ($k=40$) and first applies a learnable spatial transformation network that predicts a $3\times3$ alignment matrix to normalize input coordinates. Three edge convolution stages then compute edge features $[\mathbf{x}_j-\mathbf{x}_i,\ \mathbf{x}_i]$; each stage processes these with pointwise ($1\times1$) convolutions, Batch Normalization, and a Leaky Rectified Linear Unit (negative slope $0.2$), followed by max aggregation over neighbors, yielding three local feature maps with $64$ feature channels each. The concatenated local representation is projected to a global embedding of dimension $1024$ and globally max-pooled across points; this global vector is broadcast back to all points, concatenated with the three local maps (for a total of $1{,}216$ feature channels), and processed by pointwise convolutions $256\rightarrow256\rightarrow128\rightarrow1$ with dropout probability $0.4$ to regress one pressure value per point. Training uses $10{,}000$ points per surface, batch size $12$, and $200$ epochs, with the adaptive moment estimation optimizer (learning rate $10^{-3}$, weight decay $10^{-4}$), a reduce-on-plateau learning rate scheduler (patience $10$, reduction factor $0.1$, minimum learning rate $10^{-7}$), mean squared error loss, automatic mixed precision, and gradient clipping with norm $1.0$.

\subsection*{PointNet}
PointNet~\cite{qi2017pointnet} operates on raw point coordinates $\mathbf{X}\in\mathbb{R}^{3\times N}$. A learnable spatial transformation network predicts a $3\times3$ alignment matrix to normalize inputs. The feature extractor applies shared pointwise convolutions with Batch Normalization and ReLU activations, mapping $3\!\rightarrow\!64\!\rightarrow\!128\!\rightarrow\!1024$, followed by global max pooling to obtain a $1024$-dimensional global descriptor. For per-point regression, this global descriptor is broadcast to all points and concatenated with the intermediate $64$-dimensional point features to form a $1088$-dimensional per-point representation, which is processed by a shared regression head $1088\!\rightarrow\!512\!\rightarrow\!256\!\rightarrow\!128\!\rightarrow\!1$ with Batch Normalization, Rectified Linear Unit activations, and dropout probability $0.3$, yielding one pressure value per point.  
We modified the original PointNet architecture~\cite{qi2017pointnet} to accommodate the high-dimensional per-point regression task, replacing the classification layers with a fully convolutional regression head optimized for continuous field prediction.  
Training uses $10{,}000$ points per surface, batch size $12$, and $150$ epochs, with the adaptive moment estimation optimizer (learning rate $10^{-3}$, weight decay $10^{-4}$), a reduce-on-plateau learning rate scheduler (patience $10$, reduction factor $0.1$, minimum learning rate $10^{-7}$), mean squared error loss, and gradient clipping with norm $1.0$.

\subsection*{PointNetLarge}
In this work, we test a larger variant of PointNet~\cite{qi2017pointnet} to evaluate the effect of increased model capacity on predictive performance. The large PointNet variant operates on raw point coordinates $\mathbf{X}\in\mathbb{R}^{3\times N}$ and begins with a learnable spatial transformation network that predicts a $3\times 3$ alignment matrix to normalize input geometry. The feature extractor applies shared pointwise one--dimensional convolutions across points with channel widths $64\rightarrow128\rightarrow256\rightarrow512\rightarrow1024\rightarrow2048$, followed by a global $1\times1$ convolution to $4096$ dimensions and global max pooling to form a global descriptor. For per-point regression, the global descriptor is broadcast to all points and concatenated with multi-scale local features from the intermediate layers; this concatenated representation is processed by a deep per-point regression head with successive pointwise convolutions $2048\rightarrow1024\rightarrow512\rightarrow256\rightarrow128\rightarrow64$, Batch Normalization, and Rectified Linear Unit activations. Two residual connections are used via $1\times1$ skip projections (from $1024$ to $512$ and from $256$ to $128$) to improve information flow, followed by final pointwise layers $32\rightarrow16\rightarrow1$ to predict one pressure value per point. Regularization includes dropout with probability $0.3$ in early stages and a heavier dropout with probability $0.5$ in deeper layers. Training uses $10{,}000$ points per surface, batch size $8$, mean squared error loss, gradient clipping with norm $1.0$, automatic mixed precision, and early stopping; the optimizer was configured as the adaptive moment estimation optimizer with decoupled weight decay (base learning rate $10^{-4}$, weight decay $10^{-4}$) under a One-Cycle policy peaking at $10^{-2}$ over a $150$-epoch budget.

\subsection*{PointMAE}
We adapted an efficient encoder-decoder architecture for surface pressure field prediction, inspired by the PointMAE framework~\cite{pang2023masked}. In this benchmark, PointMAE denotes a lightweight PointMAE-inspired encoder-decoder adapted for dense surface-pressure regression, rather than a reproduction of the original masked-autoencoder pretraining architecture; the short label \emph{PointMAE} is used for this variant throughout. The model processes $N$ input points with three-dimensional coordinates. The encoder applies shared one-dimensional convolutions with Batch Normalization and Rectified Linear Unit activations to map $3\!\rightarrow\!64\!\rightarrow\!128\!\rightarrow\!256\!\rightarrow\!512$, followed by a global projection $512\!\rightarrow\!1024$ and global max pooling across points to obtain a scene-level descriptor. This global descriptor is broadcast back to all points and concatenated with the local encoder features (512-dimensional) and the raw coordinates (3-dimensional), forming a $1024{+}512{+}3$-dimensional per-point representation that is decoded by pointwise convolutions $1539\!\rightarrow\!512\!\rightarrow\!256\!\rightarrow\!128\!\rightarrow\!64\!\rightarrow\!1$ with Batch Normalization, Rectified Linear Unit activations, and dropout probabilities $0.3$ and $0.2$ in the early decoder stages, yielding one pressure value per point. Training uses $10{,}000$ points per surface with the smooth L1 loss (Huber parameter $1.0$), gradient clipping with norm $1.0$, early stopping with patience $20$ epochs, and validation every $5$ epochs. The optimizer, batch size, learning rate and schedule as \emph{executed} are reported in Table~5; the run stopped at $106$ of its $500$ configured epochs.

\subsection*{NeuralOperator}
The NeuralOperator model maps the input point cloud to a voxel grid of resolution $32^3$, using an occupancy scalar field and augmenting it with normalized grid coordinates; these four channels form the input to a three--dimensional Fourier Neural Operator~\cite{li2020fourier}. The operator lifts inputs to a width of $16$ through a linear layer, applies two spectral convolution blocks that learn complex Fourier coefficients up to $8$ modes along each spatial dimension, and combines them with parallel $1{\times}1{\times}1$ convolutions via residual connections; a Gaussian Error Linear Unit activation follows the first residual, and a final linear projection produces a single pressure channel on the grid. The grid prediction is interpolated back to the original points by trilinear sampling, then refined with a point--wise network (one--dimensional convolutions $1\rightarrow32\rightarrow16\rightarrow1$ with Batch Normalization and Rectified Linear Unit activations) to yield per--point pressure. Training uses $10{,}000$ points per surface, batch size $16$, and $100$ epochs with the adaptive moment estimation optimizer with decoupled weight decay (learning rate $2\times10^{-3}$, weight decay $10^{-4}$), cosine learning--rate schedule with period $100$ epochs, mean squared error loss, gradient clipping with norm $1.0$, and mixed precision disabled due to fast Fourier transforms.

The geometry-to-grid step of this baseline is a \emph{fixed} voxelization followed by trilinear resampling, not a learned coordinate map. The procedure therefore does not correspond to Geo-FNO~\cite{li2023fourier}, which learns a diffeomorphic deformation from the irregular physical domain onto a uniform latent mesh on which the fast Fourier transform is applied, nor to GINO~\cite{li2023geometry}, which employs graph-neural-operator layers to encode and decode between the point cloud and a latent regular grid. This baseline pairs the canonical Fourier Neural Operator with the simplest established geometry-to-grid mapping, isolating the spectral (neural-operator) inductive bias rather than the additional contribution of a geometry-specialised encoder or decoder. For this reason we refer to the baseline as \emph{FNO (voxelized)} where the distinction matters. Geometry-aware neural operators such as Geo-FNO and GINO are left as future extensions of this benchmark.

\subsection*{PointTransformer}
We employ a modified version of the PointTransformer~\cite{zhao2021point} architecture that processes $N$ input points with three--dimensional coordinates through a local self-attention mechanism. The model begins with an input projection that maps the coordinates via $3\rightarrow128\rightarrow128$ with Layer Normalization and Rectified Linear Unit activations. The core architecture consists of $12$ transformer blocks, each comprising a PointTransformer layer with $16$ nearest neighbors and a feed-forward network. The PointTransformer layer computes queries, keys, and values through a shared linear transformation $128\rightarrow384$ (split into three $128$-dimensional vectors), and applies position encoding on the relative positions between each point and its $k$-nearest neighbors via $3\rightarrow128\rightarrow128$ to inject geometric information. Attention weights are computed using a learned transformation on the difference $\mathbf{q}_i - \mathbf{k}_j + \text{pos}_{ij}$ followed by softmax normalization, and the output is aggregated as a weighted sum of value vectors plus position encodings. Each block includes a feed-forward network with expansion ratio $4$ (i.e., $128\rightarrow512\rightarrow128$) using Gaussian Error Linear Unit activations, with residual connections and Layer Normalization after both the attention and feed-forward stages. After the transformer blocks, global context is extracted via mean pooling across points, followed by $128\rightarrow128$ projection with Rectified Linear Unit activation, then broadcast and concatenated with local features to form a $256$-dimensional per-point representation. The output head applies $256\rightarrow128\rightarrow64\rightarrow1$ with Layer Normalization, Rectified Linear Unit activations, and dropout probability $0.1$, yielding one pressure value per point. Training uses $10{,}000$ points per surface with mean squared error loss, gradient clipping with norm $1.0$, and validation every epoch. The optimizer, batch size, learning rate and schedule as \emph{executed} are reported in Table~5.

\subsection*{Transolver}
The Transolver model~\cite{wu2024transolver} ingests per-point geometric features composed of three-dimensional coordinates and surface normals, forming a six-dimensional feature vector $[\mathbf{x}, \mathbf{y}, \mathbf{z}, \mathbf{n}_x, \mathbf{n}_y, \mathbf{n}_z]$. Unified position encoding is disabled, so features are processed directly without additional positional augmentation, yielding a six-dimensional input that is lifted by a multilayer perceptron to a hidden dimension of 256. The network stacks five transformer blocks with eight attention heads each. In every block, a physics-aware attention module for irregular meshes first assigns points to a fixed number of slices (32 per head) using learned soft assignment weights, pools features within each slice to form slice tokens, performs self-attention among slice tokens, and then projects the attended slice information back to points ("deslicing"); this is wrapped with Layer Normalization and a residual connection. A position-wise feed-forward network with Gaussian Error Linear Unit activations and a width expansion ratio of 2 follows, also with Layer Normalization and a residual connection; in the final block, a linear projection produces a single pressure value per point. A learnable global offset vector is added to the hidden representation before the transformer stack, and dropout is set to 0.0. Training uses $10{,}000$ points per surface sample, mean squared error loss, gradient clipping with norm $1.0$, early stopping with patience $50$ epochs, validation every $5$ epochs, and mixed precision disabled. The optimizer, batch size, learning rate and schedule as \emph{executed} are reported in Table~5; the run stopped at $241$ of its $1{,}000$ configured epochs, and the configured step schedule was not applied.

\subsection*{TransolverLarge}
We evaluate a larger variant of the Transolver model~\cite{wu2024transolver} with increased capacity. This enhanced architecture processes point-wise geometric representations consisting of three-dimensional spatial coordinates and surface normal vectors, forming a six-dimensional input feature space $[\mathbf{x}, \mathbf{y}, \mathbf{z}, \mathbf{n}_x, \mathbf{n}_y, \mathbf{n}_z]$ that is lifted by a multi-layer perceptron to a 384-dimensional hidden representation. The framework employs seven sequential transformer layers, each utilizing twelve parallel attention mechanisms. Within each layer, a physics-aware attention mechanism for irregular geometric meshes initially distributes points across 48 discrete slices per attention head through learned soft assignment weights computed via Gumbel-Softmax, aggregates information within each slice to form slice-level token representations, executes self-attention computations across these slice tokens, and subsequently redistributes the processed information back to individual points through an inverse slicing operation; this process incorporates Layer Normalization and residual connections. A position-wise feed-forward network follows with Gaussian Error Linear Unit nonlinearities and an expansion factor of 2, also employing Layer Normalization and residual pathways; the terminal layer applies Layer Normalization followed by a linear transformation to generate scalar pressure predictions per point. The hidden representations are augmented with a learnable global placeholder vector prior to transformer processing, while dropout regularization is configured at 0.1. Training uses $10{,}000$ surface points per sample, mean squared error loss, gradient norm clipping at $1.0$, early termination with $50$-epoch patience, and validation assessment every $5$ epochs. The optimizer, batch size, learning rate and schedule as \emph{executed} are reported in Table~5; the run stopped at $294$ of its $1{,}000$ configured epochs.

\subsection*{Transolver++}
The Transolver++ model~\cite{luo2025transolverPlus} processes point-wise geometric representations consisting of three-dimensional spatial coordinates and surface normal vectors, forming a six-dimensional input feature space $[\mathbf{x}, \mathbf{y}, \mathbf{z}, \mathbf{n}_x, \mathbf{n}_y, \mathbf{n}_z]$. The framework employs five sequential transformer layers, each utilizing eight parallel attention mechanisms. Within each layer, a specialized physics-aware attention mechanism for irregular geometric meshes initially distributes points across 32 discrete slices per attention head through learned soft assignment weights computed via Gumbel-Softmax, aggregates information within each slice to form slice-level token representations, executes self-attention computations across these slice tokens, and subsequently redistributes the processed information back to individual points through an inverse slicing operation; this process incorporates Layer Normalization and residual connections. A position-wise feed-forward network follows with Gaussian Error Linear Unit nonlinearities and an expansion factor of 1, also employing Layer Normalization and residual pathways; the terminal layer applies Layer Normalization followed by a linear transformation to generate scalar pressure predictions per point. The hidden representations are augmented with a trainable global placeholder vector prior to transformer processing, while dropout regularization is set to 0.0. Training uses $10{,}000$ surface points per sample with mean squared error loss, gradient norm clipping at $1.0$, and validation assessment every $5$ epochs. As executed and reported in Table~5, the model was trained with batch size $128$ for $192$ of a $200$-epoch budget using the adaptive moment estimation optimizer (learning rate $10^{-3}$, no weight decay) with early termination at $30$-epoch patience.

\subsection*{TripNet}
The model encodes 3D car geometries into a triplane representation that combines global volumetric context with local spatial resolution~\cite{chen2025tripnet}. 
Specifically, the 3D input shape is projected onto three orthogonal feature planes ($xy$, $yz$, $xz$), each represented as a $32$-channel latent feature map of size $128\times128$. 
These planes are processed independently using a shared U-Net encoder--decoder with residual convolutional blocks, skip connections, and Group Normalization layers, producing refined triplane features that capture multi-scale geometric information. 
To reconstruct local aerodynamic quantities, query points in 3D space $(x,y,z)$ are used to bilinearly sample corresponding features from each plane, which are then concatenated. 
This concatenated feature vector is passed through a four-layer MLP ($96\rightarrow256\rightarrow256\rightarrow256\rightarrow1$) with SiLU activations, predicting a scalar aerodynamic field value such as surface pressure. 
Training proceeds in two stages. The triplane occupancy field is fitted first, jointly over $500$ geometries for $500$ epochs and then per geometry for $70$ epochs (Adam, learning rate $10^{-3}$, $200{,}000$ points per step). The pressure predictor is then trained on the cached triplanes for $200$ epochs with Adam (learning rate $10^{-3}$, no weight decay and no learning-rate schedule), batch size $8$, and $200{,}000$ query points per geometry, as summarized in Table~5. 
By leveraging implicit neural representations and triplane embeddings, TripNet achieves high-fidelity, spatially continuous predictions while maintaining computational efficiency, effectively bridging volumetric field learning and geometric conditioning for aerodynamic analysis.

\subsection*{AB-UPT}
\noindent
Anchored-Branched Universal Physics Transformers (AB\mbox{-}UPT)~\cite{alkin2025UPT} are transformer-based neural surrogates. The architecture (i) decouples geometry encoding from field prediction using \emph{multi-branch operators}, improving specialization for complex surface--volume interactions; (ii) achieves scalability by performing neural simulation in a low-dimensional latent space and recovering high-resolution outputs via \emph{anchored neural field decoders}, enabling predictions on meshes with $3.3\times 10^4$ to $1.5\times 10^8$ cells; and (iii) enforces physics consistency through a \emph{divergence-free} formulation for vector fields (e.g., vorticity). The anchored decoder further permits hard constraint enforcement without degrading performance, and the overall design supports neural simulation directly from CAD geometry, obviating costly CFD meshing at inference time. The authors of AB-UPT~\cite{alkin2025UPT} kindly provided details about their training setup, reporting a total training duration of 6~hours, 6~minutes, and 51~seconds (including evaluation). 
The evaluation phase accounted for approximately 25~minutes, resulting in an effective training time of around 5~hours and 40~minutes on a single NVIDIA~H100~GPU. 
To ensure correctness and fair comparison across models, we independently re-ran the evaluation using the same configuration and verified the reported performance on our benchmark pipeline.


\subsection*{Comparative Architectural Analysis of Transformer Models}
Supplementary Table~\ref{tab:transformer_simplified} highlights the key distinctions between the transformer architectures used in the CarBench benchmark. 
PointTransformer relies on local $k$-NN attention with explicit relative position encoding, giving it strong geometric inductive bias but high memory usage due to storing neighborhood graphs. 
The Transolver family introduces slice-based global attention, which groups tokens into a small number of learnable slices, reducing the quadratic cost of full attention to $O(NG + G^2 d)$. 
Transolver++ improves stability and efficiency via temperature-scaled slicing and compact MLPs, achieving the smallest parameter count (1.81M). 
TransolverLarge expands the capacity (384 channels, 48 slices), yielding higher accuracy at higher cost. 
AB-UPT employs a hierarchical design with anchor tokens that perform full self-attention and query tokens that interact with anchors only via cross-attention. This structure yields excellent global receptive fields and extremely low memory usage (0.27 GB) while achieving the best predictive accuracy among all models. 
Together, these models illustrate the progression from local geometric attention (PointTransformer) to global, efficient, and multi-scale mechanisms (Transolver-family) and finally to hierarchical compression (AB-UPT).

\begin{table*}[h]
\centering
\caption{
Comparison of transformer-based architectures evaluated on DrivAerNet++ for 3D surface-field prediction. 
The table summarizes the core attention mechanisms, positional encoding strategies, model scales, and computational complexity of each architecture. 
Complexity notation: $N$ = number of surface points; $k$ = number of nearest neighbors for local attention; $G$ = number of learnable slices/groups in slice-based attention; 
$d$ = hidden feature dimension; 
$N_a$ = number of anchor tokens in AB-UPT encoder self-attention; 
$N_q$ = number of query tokens in decoder cross-attention. 
These quantities determine the computational cost of attention operations across different architectural paradigms.
}
\label{tab:transformer_simplified}
\small
\resizebox{\textwidth}{!}{
\begin{tabular}{lcccccc}
\toprule
\textbf{Model} &
\textbf{Attention Type} &
\textbf{Position Encoding} &
\textbf{Dim} &
\textbf{Layers} &
\textbf{Params (M)} &
\textbf{Complexity} \\
\midrule
\textbf{PointTransformer} &
Local $k$-NN self-attn &
Relative offsets $\mathbf{p}_i - \mathbf{p}_j$ (MLP) &
128 &
12 &
3.05 &
$O(Nk d)$ \\
\textbf{Transolver} &
Slice-based physics attention &
Features only (6D) &
256 &
5 &
2.47 &
$O(NG + G^2 d)$ \\
\textbf{Transolver++} &
Slice-based physics attn (Gumbel) &
Coord concat (9D) &
256 &
5 &
1.81 &
$O(NG + G^2 d)$ \\
\textbf{TransolverLarge} &
Slice-based attention (scaled-up) &
Features only (6D) &
384 &
7 &
7.58 &
$O(NG + G^2 d)$ \\
\textbf{AB-UPT}\textsuperscript{*} &
Anchor self-attn + cross-attn &
Rotary Positional Embedding (RoPE) &
192 &
11 &
6.01 &
$O(N_a^2 + N_q N_a)$ \\
\bottomrule
\end{tabular}
}
\end{table*}

\subsection*{Strictly Unified Training Configuration}
\label{sec:unified_training}

To complement the model-specific best-effort configurations used in the main
benchmark, we introduce an additional reference baseline in which all models
are trained using an identical optimization configuration. Specifically, every model
is trained with Adam using a learning rate of $10^{-3}$ and no weight decay,
MSE loss, a cosine learning-rate schedule, $100$ epochs without early
stopping, batch size $4$, $10^4$ sampled surface points, and random seed $42$.
Only the training hyperparameters are unified; each model retains its original
architecture.
\begin{table*}[h!] \centering \small \caption{ Performance of all models trained using the same optimization configuration: Adam with $\mathrm{lr}=10^{-3}$ and no weight decay, MSE loss, a cosine learning-rate schedule, $100$ epochs without early stopping, batch size $4$, $10^4$ sampled surface points, and random seed $42$. Results are reported on the held-out test split and sorted by $R^2_{\text{test}}$. MAE and RMSE are reported in kinematic pressure units (m$^2$/s$^2$). } \label{tab:strict_unified_training} \setlength{\tabcolsep}{8pt} \begin{tabular}{lccccc} \toprule \textbf{Model} & \textbf{Parameters} & \textbf{MAE (m$^2$/s$^2$)} & \textbf{RMSE (m$^2$/s$^2$)} & \textbf{Rel L2} & $\mathbf{R^2_{\text{test}}}$ \\ \midrule TransolverLarge & 7.58M & 11.61 & 23.61 & 0.1557 & 0.9610 \\ Transolver & 2.47M & 11.83 & 23.63 & 0.1558 & 0.9609 \\ Transolver++ & 1.81M & 11.78 & 23.72 & 0.1564 & 0.9606 \\ RegDGCNN & 1.44M & 17.08 & 30.04 & 0.1985 & 0.9364 \\ PointMAE & 1.67M & 17.15 & 30.63 & 0.2024 & 0.9339 \\ PointNetLarge & 32.58M & 22.07 & 34.03 & 0.2248 & 0.9184 \\ NeuralOperator & 2.10M & 24.92 & 48.45 & 0.3181 & 0.8378 \\ PointNet & 1.67M & 30.12 & 57.40 & 0.3792 & 0.7679 \\ PointTransformer & 3.05M & \multicolumn{4}{c}{Diverged under the unified configuration} \\ \bottomrule \end{tabular} \end{table*}

Supplementary Table~\ref{tab:strict_unified_training} reports the results on the held-out
test split. The overall ranking remains consistent with the main benchmark,
with the Transolver family achieving the highest accuracy
($R^2 \approx 0.96$). This indicates that the main conclusions are not an
artifact of model-specific hyperparameter selection. PointTransformer was the only model that failed to converge under the unified
configuration, with the training loss becoming NaN from the first epoch. Its
model-specific configuration uses a lower learning rate of $10^{-4}$ together
with linear learning-rate warmup, whereas the unified configuration uses
$10^{-3}$ without warmup. Gradient clipping was enabled but did not prevent
divergence. This result highlights that a single optimization configuration may not
be equally stable or appropriate for every architecture and motivates the
model-specific best-effort configurations used in the primary comparison.

\subsection*{Model Input Representations} \label{sec:model_input_representations} 
The evaluated architectures operate on different geometric representations. Supplementary Table~\ref{tab:model_inputs} summarizes the input representation, geometric features, and tensor dimensions used by each model.

\begin{table*}[h!] \centering \small \caption{ Input representation, feature format, and tensor shape used by each evaluated model, with the batch dimension omitted. Point-based models sample $10{,}000$ surface points from the full mesh. All models predict per-point surface pressure. } \label{tab:model_inputs} \setlength{\tabcolsep}{5pt} \renewcommand{\arraystretch}{1.1} \begin{tabular}{llll} \toprule \textbf{Model} & \textbf{Input representation} & \textbf{Features} & \textbf{Tensor shape} \\ \midrule PointNet & Surface point cloud & $[x,y,z]$ & $(10000,\,3)$ \\ PointNetLarge & Surface point cloud & $[x,y,z]$ & $(10000,\,3)$ \\ PointTransformer & Point cloud + local kNN ($k{=}16$) & $[x,y,z]$ & $(10000,\,3)$ \\ RegDGCNN & Point cloud + dynamic kNN ($k{=}40$) & $[x,y,z]$ & $(10000,\,3)$ \\ PointMAE & Surface point cloud & $[x,y,z]$ & $(10000,\,3)$ \\ NeuralOperator & Voxel grid from $10$k points & Occupancy ($1$ channel) & $(1,\,32,32,32)$ \\ Transolver & Point cloud + normals & $[x,y,z,n_x,n_y,n_z]$ & $(10000,\,6)$ \\ TransolverLarge & Point cloud + normals & $[x,y,z,n_x,n_y,n_z]$ & $(10000,\,6)$ \\ Transolver++ & Point cloud + normals & $[x,y,z,n_x,n_y,n_z]$ & $(10000,\,6)$ \\ TripNet & Frozen triplane + query points & $3{\times}32$ feature planes & $(3,32,128,128)+(N_q,3)$ \\ AB-UPT & Supernodes + anchors + query points & $[x,y,z]$ & $(16384,3)\!\rightarrow\!1024;\ (512,3);\ (N_q,3)$ \\ \bottomrule \end{tabular} \end{table*}

\subsection*{Exploratory Analysis of Geometry-Representation Variants} \label{sec:representation_ablation}

The comparison in Table~1 confounds two factors that vary simultaneously across models: the neural architecture and the geometric information that architecture is given as input (Supplementary Table~\ref{tab:model_inputs}). To separate them we hold the architecture, optimizer, schedule, training designs and training budget fixed and vary only the input representation of a single model, Transolver, training each variant for $200$ epochs. The benchmark configuration supplies coordinates augmented with surface normals, $[x,y,z,n_x,n_y,n_z]$; the enriched configuration adds a signed distance field, $[x,y,z,\mathrm{sdf},n_x,n_y,n_z]$, which supplies volumetric context that surface points cannot convey. To test whether architectural capacity can substitute for geometric information, we additionally train the best architecture identified in a separate architectural sweep (depth $7$, MLP ratio $2$, $64$ slices) with and without the signed distance field.

Two properties of this experiment affect the interpretation of its results. First, all variants are scored on a single common set of $5{,}000$ measured surface points per design, $5{,}770{,}000$ points in total. Because the squared error is dominated by a small number of extreme points, scoring the same model on $10{,}000$ rather than $5{,}000$ points of the same geometry shifts its MSE by roughly $8\%$, which is larger than any effect reported below. Second, one variant configured to remove surface normals trained on inputs identical to the benchmark configuration, because the corresponding option was not implemented in the data pipeline. This variant is therefore treated as a repeat run; it provides one observed run-to-run difference, although two runs are insufficient to estimate the distribution of training variability.

\begin{table*}[h!]
\centering
\small
\caption{Effect of the input geometry representation on surface-pressure accuracy, with architecture and training held fixed. All variants use the Transolver backbone trained for $200$ epochs on the same training designs. Every variant is scored on the same $5{,}000$ measured surface points per design ($5{,}770{,}000$ points over the $1{,}154$ test geometries), and errors are in kinematic-pressure units. The final column reports the change in MSE relative to the benchmark configuration with a $95\%$ confidence interval from a paired bootstrap over test geometries ($B=10{,}000$). $^{\dagger}$This variant was configured to remove surface normals, but the option had no effect and the model trained on inputs identical to the benchmark configuration; the row therefore constitutes a repeat run of that configuration, and its deviation provides one observation of run-to-run variability. No representational or architectural change is statistically distinguishable from that repeat-run difference.}
\label{tab:representation_ablation}
\setlength{\tabcolsep}{5pt}
\renewcommand{\arraystretch}{1.1}
\begin{tabular}{llccccc}
\toprule
\textbf{Variant} & \textbf{Input representation} & \textbf{MSE} & \textbf{MAE} & \textbf{RMSE} & $\mathbf{R^2_{\mathrm{test}}}$ & $\Delta$\textbf{MSE vs.\ benchmark} \\
\midrule
Repeat of benchmark config$^{\dagger}$ & $[x,y,z,n_x,n_y,n_z]$ & 575.7 & 13.09 & 23.99 & 0.9595 & $+2.14\%$ $[+1.31, +2.86]$ \\
Benchmark configuration & $[x,y,z,n_x,n_y,n_z]$ & 563.6 & 12.87 & 23.74 & 0.9604 & --- \\
{\hskip 0.5em}$+$ signed distance field & $[x,y,z,\mathrm{sdf},n_x,n_y,n_z]$ & 578.5 & 12.94 & 24.05 & 0.9593 & $+2.64\%$ $[+1.52, +3.87]$ \\
\midrule
Best architecture, no SDF   & $[x,y,z,n_x,n_y,n_z]$                  & 566.3 & 12.87 & 23.80 & 0.9602 & $+0.47\%$ $[-0.47, +1.31]$ \\
Best architecture $+$ SDF   & $[x,y,z,\mathrm{sdf},n_x,n_y,n_z]$     & 565.4 & 12.76 & 23.78 & 0.9602 & $+0.32\%$ $[-0.51, +1.14]$ \\
\bottomrule
\end{tabular}
\end{table*}

Within these particular runs, the observed changes are comparable in magnitude to the difference between the two repeated benchmark runs. Repeating the benchmark configuration changes MSE by $+2.14\%$ ($95\%$ CI $[+1.31, +2.86]$), a variant identical to the benchmark configuration in architecture and inputs, differing only in training stochasticity. Neither intervention is statistically separable from this level. The best architecture found by an extensive sweep over depth, width, attention slices and MLP ratio changes MSE by $+0.47\%$ ($[-0.47, +1.31]$), an interval containing zero. Supplying the signed distance field changes it by $+2.64\%$ ($[+1.52, +3.87]$) in the comparison that isolates it, and by $-0.14\%$ ($[-0.85, +0.59]$) when the same field is added on top of the swept architecture: the two comparisons disagree in sign, and neither is separable from the variability observed between the repeated runs. The same holds on the other metrics, which do not move coherently: adding the field to the benchmark configuration raises both MSE and mean absolute error, whereas adding it to the swept architecture lowers both slightly and leaves $R^2$ unchanged at the printed precision.

We therefore do not claim that geometric representation dominates architecture, nor the converse. Within these runs, neither lever produces a change in surface-pressure accuracy statistically separable from the difference observed between two runs of the same configuration.

Three caveats bound this analysis. It isolates representation within a single architecture family rather than across all eleven models, since the input pipeline of each baseline is tied to its own inductive bias. Each configuration was trained once, so the confidence intervals above are computed by resampling test geometries and capture only test-set sampling; they do not capture variability across training runs, which the repeated run indicates is the larger of the two. We therefore treat the outcome as an absence of measurable benefit for this pipeline as implemented rather than as evidence that volumetric conditioning is unhelpful in principle; establishing either would require several training runs per configuration.

Two observations are independent of the null result and are relevant to benchmark design. The aggregate $R^2$ is nearly constant across every variant ($0.9593$--$0.9604$) because it is dominated by the large-scale pressure variation that all variants already capture, so a single aggregate score can mask precisely the differences a study is designed to detect. And the squared error is concentrated in a small number of points: discarding the worst $0.01\%$ of surface points lowers the benchmark configuration's MSE from $563.6$ to $475.1$, so MSE comparisons at the percent level are sensitive to which points are scored, and a common scoring basis is a precondition for comparing configurations at all.

\subsection*{Design-Space Coverage of the Test Splits} \label{sec:testset_embedding}
\begin{figure}[h!]
    \centering
    \includegraphics[width=0.85\linewidth]{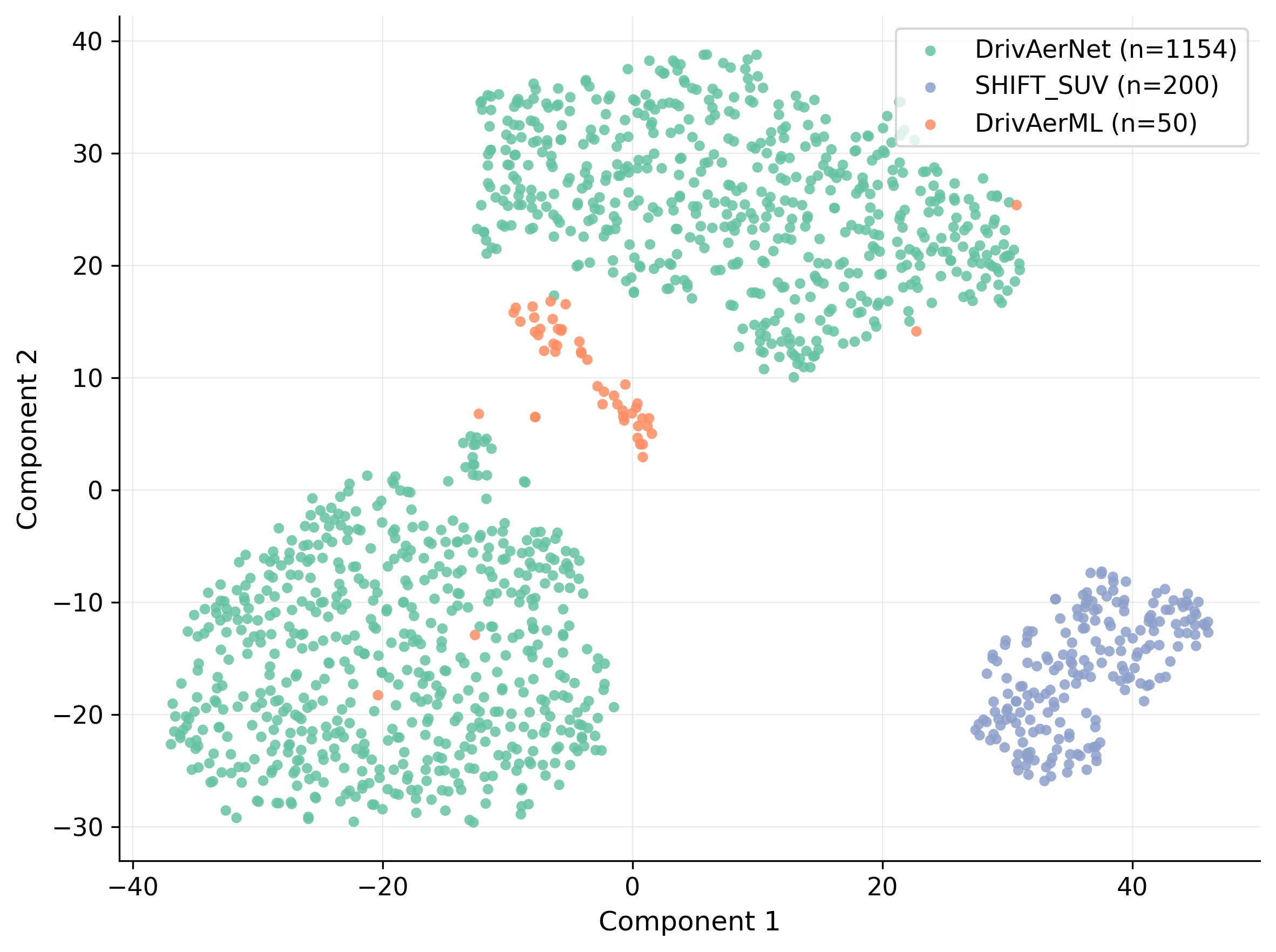}
\caption{t-SNE projection of the \emph{test} geometries only: the official DrivAerNet++ test split ($1{,}154$ designs), a $10\%$ sample of SHIFT-SUV ($200$ designs) and $50$ DrivAerML cases. Point clouds of $1{,}024$ points per geometry are embedded with the same feature extractor used for Fig.~8, which is projected linearly by PCA; here the same features are projected non-linearly by t-SNE (perplexity $30$, PCA initialization, random seed $42$). Distances and apparent containment in a t-SNE layout are qualitative and should not be read as quantitative measures of design-space coverage. Restricting the embedding to held-out geometries preserves the structure seen over the full datasets, with SHIFT-SUV forming a separate cluster and DrivAerML confined to a compact region inside the DrivAerNet++ distribution. The DrivAerNet++ test split alone is roughly $23$ times larger than a comparable DrivAerML test set and nearly six times larger than a $10\%$ SHIFT-SUV split, which is what allows the confidence intervals in Table~1 to separate models whose errors differ by less than $0.01$ Relative L2.}
    \label{fig:tsne_datasets_testset}
\end{figure}

The comparison in Fig.~8 concerns the datasets in their entirety, whereas what a benchmark actually measures is determined by its \emph{test} split. Supplementary Figure~\ref{fig:tsne_datasets_testset} therefore repeats the embedding using held-out geometries only: the official DrivAerNet++ test split ($1{,}154$ designs), a $10\%$ sample of SHIFT-SUV ($200$ designs), and $50$ DrivAerML cases, a comparable fraction of that dataset. The structure is unchanged. SHIFT-SUV remains a separate cluster and DrivAerML remains a compact group embedded within the DrivAerNet++ manifold, confirming that the differences in design-space coverage are a property of the datasets themselves.

The disparity in test-set size has a direct consequence for how confidently generalization can be assessed. A held-out set of $50$ geometries supports only a coarse estimate of expected error. Because the width of a bootstrap confidence interval scales approximately as $n^{-1/2}$, the intervals reported here over $1{,}154$ test geometries, between $\pm 0.0013$ and $\pm 0.0025$ Relative L2 (Table~1), would widen by a factor of roughly $4.8$ on a $50$-geometry test set, to approximately $\pm 0.012$. Intervals of that width, together with the much greater sensitivity of a small test set to its own composition, make reliable separation of closely ranked models far harder to achieve. Small test sets are also more sensitive to their own composition: when the held-out geometries derive from a single archetype, a model can appear to generalize well simply because the evaluation is narrow. A large and geometrically diverse test split is therefore a precondition for statistically meaningful claims about generalization, and providing one is a central design goal of CarBench.

\subsection*{From-Scratch Evaluation of AB-UPT and TripNet} \label{sec:from_scratch_evaluation}  The primary benchmark evaluates AB-UPT from the pretrained checkpoint released by its original authors, denoted AB-UPT\textsuperscript{*}. To assess whether AB-UPT's reported performance depends on the released checkpoint, we introduce an additional control in which it is trained from random initialization using only the official DrivAerNet++ training split, and we describe the TripNet reimplementation alongside it. The resulting models are evaluated on the same held-out test split and with the same surface-pressure metrics as the primary benchmark. For AB-UPT, we match the released architecture and reproduce the documented training configuration as closely as possible: Lion with a learning rate of $5\times10^{-5}$, weight decay of $0.05$, gradient clipping at $0.25$, a warmup followed by a cosine learning-rate schedule, and an exponential moving average of $0.9999$. The architecture contains approximately $6.0$M parameters and is trained for $30$ epochs with batch size $1$. Because no public TripNet implementation is available, TripNet had to be reimplemented from its published description. The implementation consists of a two-stage triplane occupancy auto-decoder, a shared U-Net, and a four-layer multilayer perceptron. The model is trained from random initialization for $200$ epochs using only the official DrivAerNet++ training split. As an implementation check, reevaluating the released AB-UPT checkpoint with our evaluation pipeline produces a pooled $R^2$ of $0.964$ under an independent resampling of the test geometries, in line with the $0.9607$ reported in the primary benchmark. All $R^2$ values in this work are pooled over test points rather than averaged over geometries; the two differ appreciably for this model, the mean-per-geometry value being $0.968$.   Supplementary Table~\ref{tab:from_scratch_comparison} compares the released AB-UPT checkpoint with the same architecture trained here. The from-scratch model retains most of the predictive accuracy of the released one, achieving $R^2_{\mathrm{test}}=0.9576$ against $0.9607$ for AB-UPT\textsuperscript{*}, which demonstrates that its strong performance can be reproduced using only the official DrivAerNet++ training split and does not depend entirely on the released checkpoint. The pretrained and from-scratch results are reported separately to make the training source explicit, and we mark the author-provided checkpoint with an asterisk (AB-UPT\textsuperscript{*}) throughout the manuscript, including Table~1 and Fig.~2, so that the headline benchmark numbers can always be traced to their training source. TripNet carries no asterisk anywhere, since every TripNet number in this work comes from our own reimplementation; our reimplementation reaches $R^2_{\mathrm{test}}=0.9537$ on the primary benchmark (Table~1), somewhat below the accuracy reported by its authors, so TripNet's results here reflect this reimplementation and may understate the method's best-reported performance.  \begin{table*}[h!] \centering \small \caption{ Comparison between the pretrained AB-UPT checkpoint released by its original authors and the same architecture trained from random initialization in this work. The from-scratch model uses only the official DrivAerNet++ training split and both are evaluated on the same held-out surface-pressure test split containing $1{,}154$ geometries. MSE is reported in squared kinematic-pressure units (m$^4$/s$^4$), while MAE and RMSE are reported in kinematic-pressure units (m$^2$/s$^2$). Throughout the manuscript, AB-UPT\textsuperscript{*} denotes the checkpoint obtained from the original authors, which is the variant reported in Table~1 and Fig.~2, whereas AB-UPT without the asterisk denotes the model trained from random initialization in this work.} \label{tab:from_scratch_comparison} \setlength{\tabcolsep}{6pt} \renewcommand{\arraystretch}{1.1} \begin{tabular}{llccccc} \toprule \textbf{Model} & \textbf{Training source} & \textbf{MSE} & \textbf{MAE} & \textbf{RMSE} & \textbf{Rel L2} & $\mathbf{R^2_{\mathrm{test}}}$ \\ \midrule AB-UPT\textsuperscript{*} & Original authors & 559 & 10.81 & 23.6 & 0.1358 & 0.9607 \\ AB-UPT & This work & 605 & 12.57 & 24.6 & 0.1539 & 0.9576 \\  \bottomrule \end{tabular} \end{table*}

\subsection*{Training Scalability at Full-Mesh Resolution}
\label{sec:full_resolution_scalability}

To examine whether the computational-efficiency rankings observed at the
$10{,}000$-point training resolution persist at higher resolutions, we profile
the peak training memory of each model as the number of input surface points
increases toward the full DrivAerNet++ mesh resolution. Peak memory is measured
during one forward-and-backward pass with batch size $1$ on a single NVIDIA
A100 GPU with 80\,GB of memory. The input resolution is increased from
$10^4$ to $5\times10^5$ surface points, approximately corresponding to the
full CFD surface mesh.

Supplementary Table~\ref{tab:full_resolution_memory} shows that scalability depends strongly
on the model architecture. RegDGCNN and PointTransformer exhibit
superlinear memory growth because their neighborhood operations construct
$k$-nearest-neighbor feature tensors at multiple layers. Although their memory
requirements remain manageable at $10^4$ points, both models require
approximately 41\,GB at $5\times10^4$ points and run out of memory at
$10^5$ points. Consequently, training these models directly on the complete
surface mesh is infeasible on a single 80\,GB accelerator.

\begin{table*}[h!]
\centering
\caption{
Peak training memory in GB for one forward-and-backward pass with batch size
$1$ as a function of the number of input surface points. Measurements were
performed on a single NVIDIA A100 GPU with 80\,GB of memory. OOM denotes an
out-of-memory failure, and ``--'' indicates configurations that were not
evaluated after an earlier out-of-memory failure.
}
\label{tab:full_resolution_memory}
\small
\setlength{\tabcolsep}{8pt}
\begin{tabular}{lrrrrrr}
\toprule
Model
& 10k
& 25k
& 50k
& 100k
& 200k
& 500k \\
\midrule
Transolver
& 0.8
& 1.9
& 3.9
& 7.6
& 15.0
& 37.1 \\

Transolver++
& 0.7
& 1.7
& 3.3
& 6.4
& 12.8
& 31.9 \\

TransolverLarge
& 1.8
& 4.3
& 8.5
& 16.6
& 33.1
& \textbf{OOM} \\

PointNet
& 0.3
& 0.8
& 1.5
& 3.0
& 6.0
& 15.0 \\

PointNetLarge
& 1.9
& 4.1
& 7.7
& 15.0
& 36.3
& \textbf{OOM} \\

PointMAE
& 0.4
& 1.0
& 1.9
& 3.8
& 7.6
& 19.0 \\

NeuralOperator
& 0.1
& 0.1
& 0.1
& 0.1
& 0.2
& 0.4 \\

RegDGCNN
& 3.4
& 13.0
& 41.0
& \textbf{OOM}
& --
& -- \\

PointTransformer
& 6.7
& 18.2
& 41.2
& \textbf{OOM}
& --
& -- \\
\bottomrule
\end{tabular}
\end{table*}

In contrast, Transolver and Transolver++ exhibit approximately linear memory
scaling with the number of input points. Their Physics-Attention mechanism
maps the input points to a fixed number of learned physical slices, resulting
in a computational complexity of approximately $\mathcal{O}(NM)$ for $N$
input points and $M$ slices. Both models remain trainable at
$5\times10^5$ points, requiring 37.1\,GB and 31.9\,GB, respectively.
PointNet and PointMAE also scale approximately linearly, whereas the NeuralOperator
operates on a fixed voxel representation and therefore exhibits only a weak
dependence on the number of original surface points.

These results demonstrate that computational-efficiency rankings are
resolution-dependent. Neighborhood-based models that appear competitive at
the subsampled resolution become impractical as the input approaches the full
mesh, whereas the approximately linear scaling of Transolver and Transolver++
allows their efficiency advantage to persist at substantially higher
resolutions. The reported out-of-memory thresholds are conservative because a
small amount of GPU memory was occupied by resident processes during
profiling.

\clearpage

\section*{Navier--Stokes Equations}

In this benchmark, the objective is to learn the steady-state pressure distribution \( p(\mathbf{x}) \) on the surface of a car, as computed by high-fidelity computational fluid dynamics (CFD) simulations. The underlying physical behavior is governed by the incompressible Navier--Stokes equations:
\begin{align}
\nabla \cdot \mathbf{u} &= 0, \label{eq:continuity} \\
\rho \left( \frac{\partial \mathbf{u}}{\partial t} + (\mathbf{u} \cdot \nabla)\mathbf{u} \right) &= -\nabla p + \mu \nabla^2 \mathbf{u} + \mathbf{f}, \label{eq:momentum}
\end{align}
where \( \mathbf{u} \) is the velocity vector field, \( p \) the pressure field, \( \rho \) the fluid density, \( \mu \) the dynamic viscosity, and \( \mathbf{f} \) represents external body forces (e.g., gravity). Equation~\eqref{eq:continuity} enforces mass conservation, while Equation~\eqref{eq:momentum} represents momentum conservation.

For turbulent flow the equations above are solved in Reynolds-averaged form: each field is decomposed into a mean and a fluctuation, and averaging Equation~\eqref{eq:momentum} introduces the Reynolds-stress term $-\partial(\rho\overline{u_i'u_j'})/\partial x_j$, which is modelled through an eddy viscosity $\mu_t$ so that the mean momentum balance carries an effective viscosity $\mu_{\mathrm{eff}}=\mu+\mu_t$. Equations~\eqref{eq:continuity}--\eqref{eq:momentum} and the pressure Poisson equation below are therefore to be read for the mean fields. To close the system for turbulent flows at high Reynolds numbers typical of automotive applications, we use the Reynolds-Averaged Navier--Stokes (RANS) approach with the \( k\text{-}\omega \) Shear Stress Transport (SST) turbulence model~\cite{menter2003ten}. This model solves two additional transport equations for the turbulent kinetic energy \( k \) and the specific dissipation rate \( \omega \):
\begin{align}
\frac{\partial (\rho k)}{\partial t} + \nabla \cdot (\rho k \mathbf{u}) &= P_k - \beta^* \rho k \omega 
+ \nabla \cdot \left[ (\mu + \sigma_k \mu_t) \nabla k \right], \\
\frac{\partial (\rho \omega)}{\partial t} + \nabla \cdot (\rho \omega \mathbf{u}) &= \alpha \frac{\omega}{k} P_k - \beta \rho \omega^2 
+ \nabla \cdot \left[ (\mu + \sigma_\omega \mu_t) \nabla \omega \right] \nonumber \\
&\quad + 2(1-F_1) \rho \sigma_{\omega 2} \frac{1}{\omega} \nabla k \cdot \nabla \omega.
\end{align}
Here \( P_k \) denotes the production of turbulent kinetic energy and \( \mu_t \) is the eddy viscosity, which in the SST formulation is $\mu_t = \rho a_1 k / \max(a_1\omega,\, S F_2)$ and reduces to $\rho k/\omega$ away from regions where the shear-stress limiter is active. The SST model blends the standard \( k\text{-}\omega \) model near the wall with the \( k\text{-}\varepsilon \) model in the freestream, improving accuracy for adverse pressure gradients and separation.

By taking the divergence of the momentum equation~\eqref{eq:momentum} and using the continuity constraint~\eqref{eq:continuity}, we obtain the pressure Poisson equation:
\begin{equation}
\nabla^2 p = -\rho \, \nabla \cdot \left[ (\mathbf{u} \cdot \nabla) \mathbf{u} \right] + \nabla \cdot \mathbf{f}.
\end{equation}
This equation reveals that the pressure field arises from the spatial acceleration of the flow and any applied forces. In our learning task, the surrogate models aim to approximate this surface pressure field directly from the car geometry, bypassing the need for solving the full Navier--Stokes system during inference. The pressure field on the car surface emerges from the solution to the momentum equation and is influenced by velocity gradients and turbulence-induced stresses. In regions with complex flow separation, curvature, or underbody vortices, the accurate prediction of surface pressure becomes a challenging task. Our deep learning models aim to learn the nonlinear mapping from surface geometry to the pressure distribution without explicitly solving the above PDE system.

\subsection*{Pressure Drag and Its Role in Automotive Aerodynamics}

The total aerodynamic drag \( D \) experienced by a car moving through air can be decomposed into two primary components:
\begin{equation}
D = D_p + D_f,
\end{equation}
where
\begin{itemize}
    \item \( D_p \) is the pressure drag (form drag),
    \item \( D_f \) is the friction drag (skin-friction drag due to shear stresses).
\end{itemize}

Pressure drag arises from the imbalance between the high pressure at the front of the car and the low pressure in the wake region at the rear. It is computed by integrating the surface pressure over the body in the streamwise direction. Using the outward surface normal \( \mathbf{n} \) and freestream direction \( \hat{\mathbf{x}} \), the pressure drag can be expressed as
\begin{equation}
D_p = - \int_{A} \bigl(p(\mathbf{x}) - p_\infty\bigr) \, \mathbf{n} \cdot \hat{\mathbf{x}} \, dA,
\end{equation}
where
\begin{itemize}
    \item \( A \) is the wetted surface of the vehicle, over which the integral is taken (distinct from the projected frontal reference area $A_{\text{ref}}$ used in the drag coefficient below),
    \item \( \mathbf{n} \) is the outward surface normal vector,
    \item \( \hat{\mathbf{x}} \) is the unit vector in the freestream (drag) direction,
    \item \( p_\infty \) is the freestream static pressure.
\end{itemize}

In contrast, friction drag is computed from the wall shear stress vector \( \boldsymbol{\tau}_w \):
\begin{equation}
D_f = \int_{A} \boldsymbol{\tau}_w \cdot \hat{\mathbf{x}} \, dA,
\qquad
\boldsymbol{\tau}_w = \mu \left. \frac{\partial \mathbf{u}_t}{\partial n} \right|_{w},
\end{equation}
where \( \mathbf{u}_t \) is the tangential velocity at the wall and \( n \) is the wall-normal direction.

The dimensionless drag coefficient \( C_D \) is defined as
\begin{equation}
C_D = \frac{D}{\tfrac{1}{2} \rho U_\infty^2 A_{\text{ref}}},
\end{equation}
where \( U_\infty \) is the freestream velocity and \( A_{\text{ref}} \) is the frontal area of the car.

For streamlined cars, friction drag can be significant; however, in typical passenger cars with bluff geometries and separated wakes, pressure drag is often the dominant contributor, accounting for up to 80--90\%~\cite{schuetz2015aerodynamics, sudin2014review} of the total drag. Therefore, accurately learning and predicting the surface pressure distribution is critical for estimating and optimizing aerodynamic performance.

\subsection*{Relevance of Pressure Distribution to Aeroacoustics}

In addition to its critical role in aerodynamic drag, the pressure field around a car also contributes significantly to aerodynamic noise generation, particularly at high speeds. This phenomenon, known as aeroacoustic noise, originates from unsteady pressure fluctuations associated with turbulent flows, separations, wakes, and vortices.

One of the key frameworks in aeroacoustics is Lighthill's acoustic analogy~\cite{lighthill1952sound}, which describes the generation of sound due to turbulence:
\begin{equation}
\frac{\partial^2 \rho'}{\partial t^2} - c_0^2 \nabla^2 \rho' = \frac{\partial^2 T_{ij}}{\partial x_i \partial x_j},
\end{equation}
where
\begin{itemize}
    \item \( \rho' \) is the acoustic density fluctuation,
    \item \( c_0 \) is the speed of sound,
    \item \( T_{ij} \) is Lighthill's stress tensor:
    \[
    T_{ij} = \rho u_i u_j + \bigl[(p - c_0^2 \rho)\delta_{ij} - \tau_{ij}\bigr],
    \]
    \item \( \tau_{ij} \) is the viscous stress tensor.
\end{itemize}

The dominant aerodynamic noise sources in cars arise from turbulent boundary layers forming over mirrors and A-pillars, flow separation at the rear, and vortex shedding in the wheel arches and underbody regions~\cite{oettle2017automotive}. These localized and rapid pressure fluctuations, particularly in the near-wake, generate broadband acoustic emissions that contribute significantly to overall car noise. Consequently, accurately learning and predicting the surface pressure distribution is essential for estimating drag and for locating the regions in which such sources arise. We note that the steady RANS simulations used here provide the \emph{mean} surface pressure only; quantifying acoustic source strength or radiated noise would require the unsteady pressure spectrum, which this dataset does not contain.

\end{document}